\documentclass[letterpaper, 10 pt, conference]{ieeeconf}  

\IEEEoverridecommandlockouts                              

\usepackage{graphics} 
\usepackage{graphicx} 
\usepackage{wrapfig}
\usepackage{hyperref}
\usepackage{amsmath} 
\usepackage{amssymb}  
\usepackage{color}
\usepackage{lipsum}
\usepackage{soul}
\usepackage{subcaption}
\usepackage[ruled,vlined,linesnumbered,noend]{algorithm2e}
\usepackage{comment}
\usepackage{float}
\usepackage{adjustbox}
\usepackage[noadjust,compress]{cite}
\usepackage{ifthen}

\DeclareMathOperator*{\argmin}{arg\,min}
\newcommand{\rulesep}{\unskip\ \vrule\ }

\newenvironment{myitem}
{
    \begin{list}{$\circ$ }{}
        \setlength{\topsep}{0pt}
        \setlength{\parskip}{0pt}
        \setlength{\partopsep}{0pt}
        \setlength{\parsep}{0pt}         
        \setlength{\itemsep}{0pt} 
	\setlength{\leftskip}{-12pt}
}
{
    \end{list} 
}

 \def\length{long}

\title{\LARGE \bf
Roadmaps with Gaps over Controllers:\\ Achieving Efficiency in Planning under Dynamics
}

\author{Aravind Sivaramakrishnan, Sumanth Tangirala, Edgar Granados, Noah R. Carver,  and Kostas E. Bekris
\thanks{The authors are with the Dept. of Computer Science, Rutgers University, NJ, USA. E-mail: {\tt \{as2578, kb572\}@rutgers.edu}. This work is partly supported by NSF NRT-FW-HTS award 2021628.
}%
}

\begin{document}

\maketitle
\thispagestyle{empty}
\pagestyle{empty}

\begin{abstract}

This paper aims to improve the computational efficiency of motion planning for mobile robots with non-trivial dynamics through the use of learned controllers. Offline, a system-specific controller is first trained in an empty environment. Then, for the target environment, the approach constructs a data structure, a ``Roadmap with Gaps,'' to approximately learn how to solve planning queries using the learned controller. The roadmap nodes correspond to local regions. Edges correspond to applications of the learned controller that approximately connect these regions. Gaps arise as the controller does not perfectly connect pairs of individual states along edges. Online, given a query, a tree sampling-based motion planner uses the roadmap so that the tree's expansion is informed towards the goal region. The tree expansion selects local subgoals given a wavefront on the roadmap that guides towards the goal. When the controller cannot reach a subgoal region, the planner resorts to random exploration to maintain probabilistic completeness and asymptotic optimality. The accompanying experimental evaluation shows that the approach significantly improves the computational efficiency of motion planning on various benchmarks, including physics-based vehicular models on uneven and varying friction terrains as well as a quadrotor under air pressure effects. 
Website: \url{https://prx-kinodynamic.github.io/projects/rogue}
\end{abstract}


\ifthenelse{\equal{\length}{long}}{
    \section{Introduction}
\label{sec:introduction}

Kinodynamic motion planning allows mobile robots with non-trivial dynamics to negotiate environments with obstacles, such as a warehouse, or physical features, such as uneven terrain and ice. The problem is challenging when there is no local steering function available that connects two states. Tree sampling-based planners \cite{lavalle2001randomized} do not need a steering function as they only forward propagate the system's dynamics.  Some variants provide Asymptotic Optimality (AO) by propagating randomly sampled controls \cite{li2016asymptotically, hauser2016asymptotically,  LB-DIRT, kleinbort2020refined}. While random controls provide theoretical properties, they result in slow convergence to high-quality solutions in practice. 

This work aims to improve the efficiency of such AI kinodynamic planners \cite{Faust2018PRMRLLR, chiang2019rl} by integrating controllers trained via Reinforcement Learning (RL) \cite{ROB-063}. Supervision has also be used to train controllers that are then integrated with planners \cite{Li2018NeuralNA, kontoudis2019kinodynamic, johnson2020mpnet, li2021mpc}. This is often done, however, under the assumption that a steering function is available for the robot, which is not assumed here. 
While one may attempt to solve planning problems directly with RL, this often suffers from large training data requirements and requires careful tuning of rewards \cite{xu2022benchmarking}. Most RL solutions lack long-horizon reasoning, which a planner provides. 

Several recent works have leveraged RL to build an abstract representation of the planning problem and search for an optimal path between the start and the goal. Search on the Replay Buffer ({\tt SoRB}) \cite{eysenbach2019search} and Sparse Graphical Memory ({\tt SGM}) \cite{emmons2020sparse} build a graph where the nodes correspond to states visited by the RL agent during training. Similar to {\tt SoRB} and {\tt SGM}, Deep Skill Graphs \cite{bagaria2019option, bagaria2021robustly, bagaria2021skill} builds a graph representation, where nodes correspond to subgoals, and edges correspond to control policies between them. A common feature of these methods is that the control policy is trained in the planning environment of interest. This requires the learning algorithm to reason jointly about the system's dynamics and the obstacles present in the environment, which is challenging for second-order systems \cite{strudel2020learning}. 

\begin{figure}
    \centering
    \includegraphics[width=.49\columnwidth]{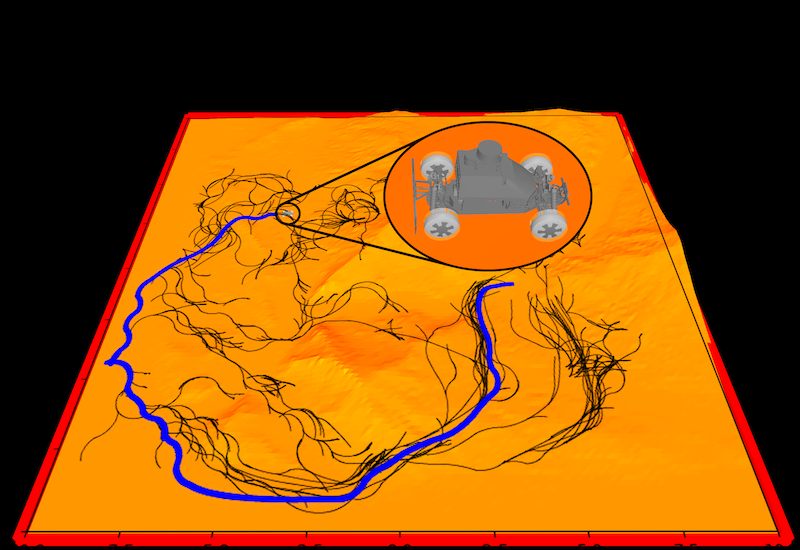}
    \includegraphics[width=.49\columnwidth]{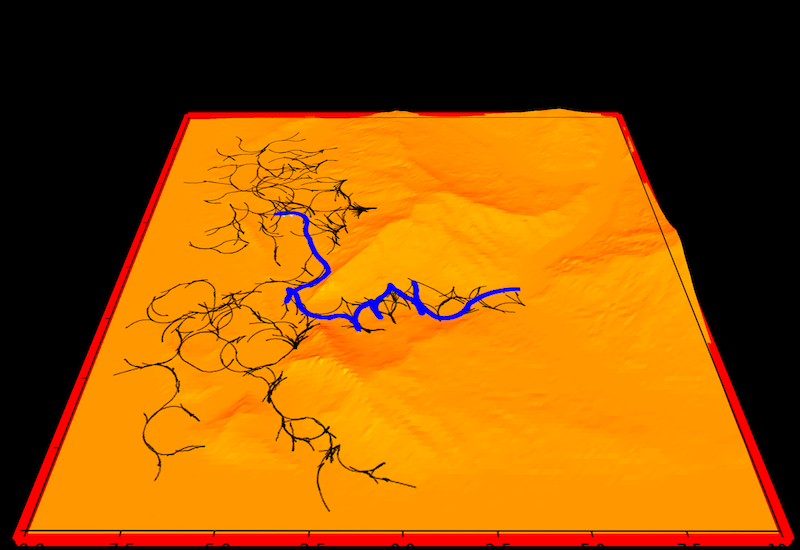}
    \put(-103pt,72pt){\textcolor{white}{\smaller $\mathtt{RLG}$ solution cost: 79.01s}}
    \put(-241pt,72pt){\textcolor{white}{\smaller $\mathtt{RoGuE}$ (Ours) solution cost: 47.05s}}
    \vspace{-.05in}
    \caption{\small Solution trajectories (thick lines) and planning trees (thin lines) for a MuSHR vehicle over an uneven terrain in MuJoCo. The cost function is trajectory duration, which is impacted by the uneven terrain. The proposed expansion method for kinodynamic planning (left) leverages a ``Roadmap with Gaps" to avoid difficult terrain resulting in shorter duration solutions. The alternative expansion (right), which samples random local goals, navigates the rough hills less effectively. It results in a shorter distance path but a much slower trajectory. Planning time was 60s for both methods.}
    \label{fig:overview}
    \vspace{-0.25in}
\end{figure}

Prior work by the authors integrates RL in kinodynamic motion planning via a decoupled strategy \cite{sivaramakrishnan2021improving, troy2022terrains}, where a controller is first trained offline in an empty environment and the learning process only deals with the system's dynamics. The absence of obstacles allows learning the controller with significantly fewer data and only a sparse reward. During online planning, these prior efforts engineer local goals that bias expansion toward the global goal via manual engineering. Designing, however, an informed local goal procedure requires manual effort, which is undesirable and may not work well across environments. Furthermore, the local goals these procedures generate are unaware of the trained controller's reachability properties. 


This work proposes a data structure, a \textbf{Roadmap with Gaps},  that learns the approximate reachability of local regions in a given environment given a controller. The approach constructs a directed graph in the free configuration space of the robot. An edge exists from a source node/region to a target when the source node's nominal state can reach the target node's termination set without collisions by applying the controller. Gaps arise since the learned controller does not connect any two exact states exactly. Given a new planning query, a wavefront expressing the cost-to-goal on this roadmap provides every node with the direction of motion towards the goal.  This roadmap guidance is integrated with an AO tree sampling-based planner. During the tree planner's expansion, nodes that are closer to the global goal given the roadmap guidance are prioritized.

The proposed approach provides a tradeoff. It incurs an offline computational cost for pre-processing a known environment and the robot's dynamics to provide online computational benefits for multiple planning queries in the same workspace. At the same time it retains AO properties and is an informed, efficient solution by understanding the reachability properties of the learned controller and the environment's connectivity. Prior approaches that guide tree sampling-based planners using graphical abstractions \cite{le2014guiding, westbrook2020anytime} do not provide desirable properties (e.g., AO) or do not reason about the system's dynamics. A related kinodynamic planning approach introduced the \textit{Bundle of Edges (BoE)} notion \cite{shome2021asymptotically} but relies on random control propagation, which reduces efficiency. 

In summary, this paper proposes the \textbf{Ro}admap-\textbf{Gu}ided \textbf{E}xpansion (\textbf{\tt RoGuE})-Tree method, which:
\begin{myitem}
\item Introduces the ``Roadmap with Gaps,'' a data structure that approximately guides how a robot can navigate a target environment given a controller, and
\item Adapts an Asymptotically Optimal (AO) framework for kinodynamic planning to utilize the ``Roadmap with Gaps''.
\end{myitem}

The accompanying evaluation shows that \textbf{\tt RoGuE} results in lower-cost solutions faster than random propagation given an AO sampling-based kinodynamic planner both on analytically and physically simulated models in various environments. This includes physics-based vehicular models on uneven and varying friction terrains as well as a quadrotor under air pressure effects.

    \section{Preliminaries}
\label{sec:prelims}

Consider a system with state space $\mathbb{X}$ and control space $\mathbb{U}$. $\mathbb{X}$ is divided into  collision-free ($\mathbb{X}_f$) and obstacle ($\mathbb{X}_{\mathrm{o}}$) subsets. The dynamics $\dot{x} = f(x,u)$ (where $x \in \mathbb{X}_f, u \in \mathbb{U}$) govern the robot's motions.  The process $f$ can be an analytical ordinary differential equation (ODE) or modeled via a physics engine, e.g., MuJoCo \cite{todorov2012mujoco}.  A function $\mathbb{M}: \mathbb{X} \rightarrow \mathbb{Q}$ maps a state $x \in \mathbb{X}$ to its corresponding \textit{configuration space} point $q \in \mathbb{Q}$ ($q = \mathbb{M}(x)$). The inverse mapping $\mathbb{M}^{-1}(q_i)$ returns a state $x_i \in \mathbb{X}$ by setting the dynamics term to some nominal value (e.g., by setting all velocity terms to 0). A distance function $d(\cdot,\cdot)$ is defined over $\mathbb{Q}$, which corresponds to a weighted Euclidean distance metric in SE(2).  


A \textit{plan} $p_T$ is a sequence of piecewise-constant controls of duration $T$ that induce a trajectory $\tau \in \mathcal{T}$, where $\tau: [0,T] \mapsto \mathbb{X}_f$. Given a start state $x_0 \in \mathbb{X}_f$ and a goal set $X_G \subset \mathbb{X}_f$, a feasible motion planning problem admits a solution trajectory of the form $\tau(0) = x_0, \tau(T) \in X_G$.   The goal set is defined as $X_G = \{x \in \mathbb{X}_f \ \vert \ d(\mathbb{M}(x), q_G) < \epsilon \}$, or equivalently, $\mathcal{B}(q_G, \epsilon)$ where $\epsilon$ is a tolerance parameter.
A heuristic $h: \mathbb{X} \rightarrow \mathbb{R}^+$ estimates the \textit{cost-to-go} of an input state $x$ to the goal region $X_G$.
Each solution trajectory has a cost given by $\texttt{cost}: \mathcal{T} \rightarrow \mathbb{R}^+$. An optimal motion planning problem aims to minimize the cost of the solution trajectory.


{\bf Sampling-Based Planning Framework:} The framework uses the forward propagation model $f$ to explore $\mathbb{X}_f$ by incrementally constructing a tree via sampling in $\mathbb{U}$. Algo.~\ref{alg:tree-sbmp} outlines the high-level operation of a sampling-based motion planner (Tree-SBMP) that builds a tree of states rooted at the initial state $x_0$ until it reaches $X_G$. 

\vspace{-.15in}
\begin{algorithm}
        \SetAlgoLined
        $\mathrm{T} \leftarrow \{x_0\}$; \\
        \While{termination condition is not met}
        {
        $x_\text{sel} \leftarrow \texttt{SELECT-NODE}(\mathrm{T})$; \\ 
        $u \leftarrow \texttt{EXPAND}(x_\text{sel})$; \\
        $x_\text{new} \leftarrow \texttt{PROPAGATE}(x_\text{sel},u)$; \\
        \If{$(x_\text{sel} \rightarrow x_\text{new}) \in \mathbb{X}_f$}
        {
        \texttt{EXTEND-TREE}($\mathrm{T}, x_\text{sel} \rightarrow x_\text{new}$);
        }
        }
        \caption{\small Tree-SBMP($\mathbb{X},\mathbb{U},x_0,X_G$)}
        \label{alg:tree-sbmp}
\end{algorithm}
\vspace{-.2in}


Each iteration of Tree-SBMP selects an existing tree node/state $x_\text{sel}$ to expand (Line 3). Then, it generates a control sequence $u$ and propagates it from $x_\text{sel}$ to obtain a new state $x_\text{new}$ (Lines 4-5). The resulting edge is added to the tree if not in collision (Lines 6-7). Upon termination (e.g., a time threshold), if the tree has states in $X_G$, the best-found solution according to $cost$ is returned.  By varying how the key operations in Algo. ~\ref{alg:tree-sbmp} are implemented, different variations can be obtained. The specific variant this work adopts is the informed and AO ``Dominance-Informed Region Tree'' (DIRT) \cite{LB-DIRT}. It uses an admissible state heuristic function $h$ to select nodes in an informed manner. If $x_\text{new}$ improves upon $x_\text{sel}$ given $h$, then it is selected as the next $x_\text{sel}$. DIRT also propagates a ``blossom'' of $k$ controls at every iteration and prioritizes the propagation of the edge that brings the robot closer to the goal given the heuristic.


    \section{Proposed Method}
\label{sec:proposed}

\begin{figure}[t]
    \centering
    \includegraphics[width=\linewidth,trim={0 5.5cm 0 0},clip]{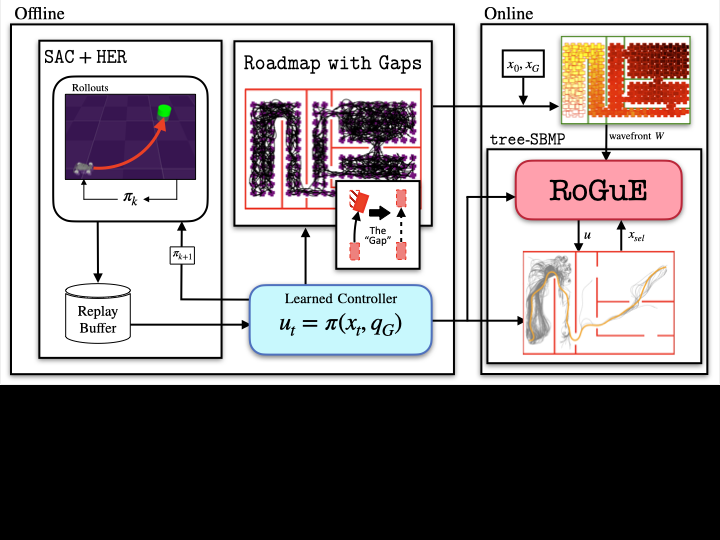}
\vspace{-.2in}
    \caption{\small Stages of the proposed pipeline. \textit{Offline: } A controller is trained in an empty environment. The roadmap with gaps is built over the robot's free C-space in a target environment. \textit{Online:} Given a new query $(x_0,x_G)$, a wavefront is computed over the roadmap. At every iteration of the tree sampling-based planner, a local goal is computed given the wavefront information, and a candidate control is propagated towards the local goal via the trained controller.}
    \label{fig:Pipeline}
    \vspace{-.15in}
\end{figure}

The focus of this work is on line 4 of Algo. ~\ref{alg:tree-sbmp} above, i.e., how to generate the control that is expanded out of $x_\text{sel}$. In particular, the method computes first a local goal $q_\text{lg}$ and propagates a control $u = \pi(x_\text{sel},q_\text{lg})$ that progresses towards that local goal given a controller $\pi$. The controller is first applied at the selected node $x_\text{sel} \in \mathbb{X}$ and generates a control sequence towards reaching the local goal $q_\text{lg}$ without considering obstacles. 

Overall, the proposed method (Fig~\ref{fig:Pipeline}) has 2 offline stages: (i) training a controller $\pi(x,q)$ in an obstacle-free environment, and (ii) building a ``Roadmap with Gaps" $(\mathcal{V,E})$ in the target environment given the available controller. Finally, it has an online phase given a motion planning query $(x_0,x_G)$ in the target environment. Online, a sampling-based kinodynamic planner expands a tree of feasible trajectories guided by a wavefront function computed over the ``Roadmap with Gaps".

\subsection{Training a Controller offline}

A goal-conditioned controller $u = \pi(x,q)$ is first trained via reinforcement learning to reach from an initial state $x$ to a goal set $\mathcal{B}(q,\epsilon)$, i.e., within an $\epsilon$ distance of $q$. For the training, this is attempted for any $(x,q) \in \mathbb{X} \times \mathbb{Q}$ in an empty environment of given dimensions. The training process collects transitions $(x_t, u_t, \mathcal{C}(x_t,q), x_{t+1}, q)$, which are stored in a replay buffer. The cost function $\mathcal{C}: \mathbb{X} \times \mathbb{Q} \rightarrow \{0,-1\}$ has a value of $\mathcal{C}(x_t,q) = 0$ iff  $x_t \in \mathcal{B}(q,\epsilon)$, and $-1$ otherwise. During each iteration, mini-batches of transitions are sampled from the buffer. A Soft Actor-Critic (SAC) \cite{Haarnoja2018SoftAO} algorithm is employed, which minimizes the total cost $\mathbb{E}_{x_0, q_G \sim \mathbb{X} \times \mathbb{Q}} [\sum_{t=0}^T \mathcal{C}(x_t,q_G)]$. Concurrently, Hindsight Experience Replay (HER) \cite{Andrychowicz2017HindsightER} relabels some transitions with alternative goals to provide additional training signals from past experiences.

\subsection{Building a Roadmap with Gaps offline}
\label{subsec: roadmap}

The proposed approach then builds a ``Roadmap with Gaps"  $(\mathcal{V,E})$ at the target environment, i.e., a graphical representation where nodes $\mathcal{V}$ correspond to configurations $q_i$ of the vehicle. Edges $(q_i,q_j) \in \mathcal{E}$ exist between vertices as long as the application (of maximum duration $T_\text{max}$) of the available controller $\pi(\mathbb{M}^{-1}(q_i), q_j)$ at a zero velocity state $\mathbb{M}^{-1}(q_i)$ of the initial configuration $q_i$ brings the system within a hyperball $\mathcal{B}(q_j,\epsilon)$ of the configuration $q_j$.

The roadmap construction procedure (Fig~\ref{fig:main-figure}) samples first a set of configuration \textit{milestones} $\{q_i\}_{i=1}^{\vert N \vert}$, which form the vertices  $\mathcal{V}$ in the roadmap. In this work, the milestones are chosen over a grid in $\mathbb{Q}$. They are collision-checked and verified to be in $\mathbb{Q}_{\mathrm{f}}$. Then, for the generation of edges $\mathcal{E}$, the procedure selects a random configuration $q_i \in \mathcal{V}$ and obtains two sets: $A$ and $D$. $A$ is the set of all vertices $q \in \mathcal{V},$ whose hyperballs $\mathcal{B}(q,\epsilon)$ can be reached from $x_i = \mathbb{M}^{-1}(q_i)$ given the controller $\pi$. Similarly, $D$ is the set of vertices $q \in \mathcal{V}$, where the state $x = \mathbb{M}^{-1}(q)$ can reach the hyperball $\mathcal{B}(q_i,\epsilon)$ given the controller $\pi$. Then, edges from $q_i$ to vertices in $A$ and edges from vertices in $D$ to $q_i$ are added to the set $\mathcal{E}$. 

The application of the controller for the edge construction, especially a learned one for a non-linear dynamical system, implies that for all edges $(q_i,q_j)$ in the roadmap, there is no guarantee that the resulting configuration $q_j$ can be achieved exactly from $q_i$. Instead, the corresponding edge only guarantees that the system will be within a hyperball $\mathcal{B}(q_G,\epsilon)$ if it is initialized at $\mathbb{M}^{-1}(q_i)$.
\begin{wrapfigure}{r}{0.35\linewidth}
     \centering
     \vspace{-.15in}
     \includegraphics[width=\linewidth]{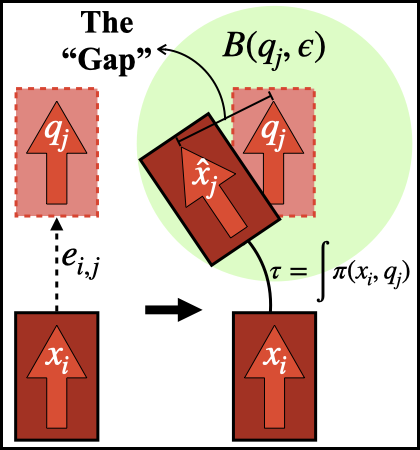}
     \vspace{-.35in}
     \label{fig:enter-label}
\end{wrapfigure}
Consequently, the edges of the roadmap introduce {\bf ``gaps"} as highlighted by the figure. This issue is exacerbated if the controller is used to follow a sequence of edges on the roadmap, e.g., $(q_i,q_j)$ and $(q_j,q_k)$. Since the controller's application over the first edge can only bring the system in the vicinity of $q_j$,  the application of the controller corresponding to the second edge at the resulting configuration may not even bring the system within an $\epsilon$-hyperball of $q_k$. Consequently, paths on the ``Roadmap with Gaps" do not respect the dynamic constraints of the vehicle and cannot be used to directly solve kinodynamic planning problems. They can still provide, however, useful guidance for a kinodynamic planner.

\begin{figure*}[ht!]
    \centering
    \begin{subfigure}{.325\textwidth}
    \includegraphics[width=\textwidth]{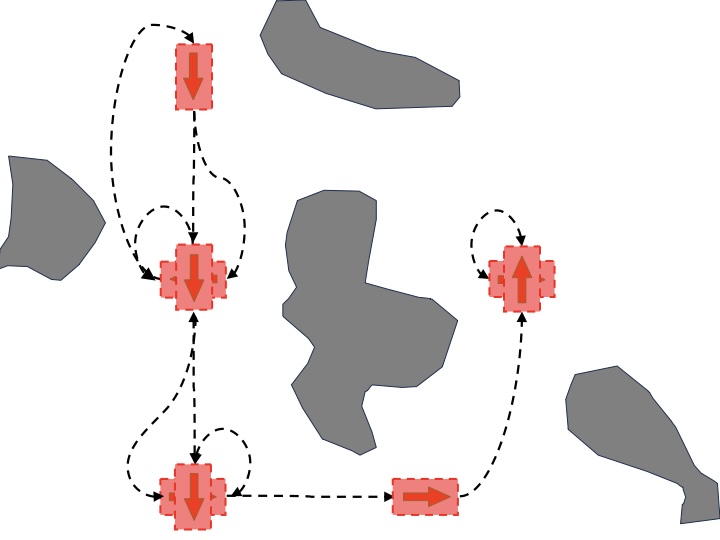}
    \vspace{-.3in}
    \caption{}
    \label{fig:main-a}
    \end{subfigure}
    \begin{subfigure}{.325\textwidth}
    \includegraphics[width=\textwidth]{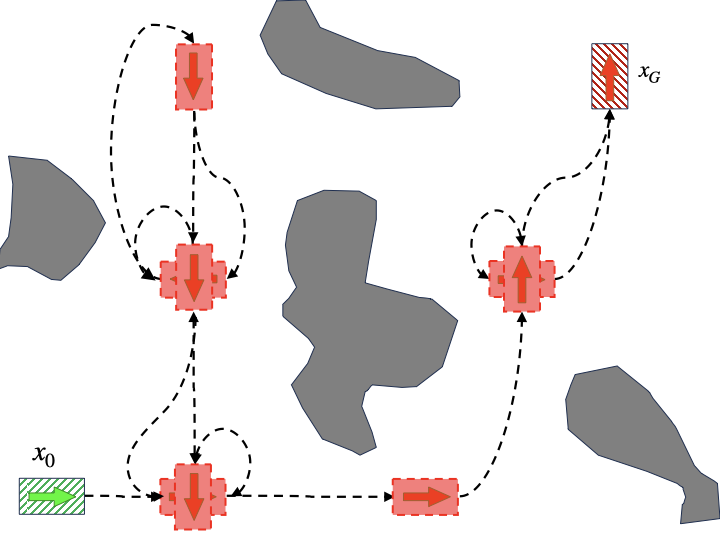}
    \vspace{-.3in}
    \caption{}
    \label{fig:main-b}
    \end{subfigure}
    \begin{subfigure}{.325\textwidth}
     \includegraphics[width=\textwidth]{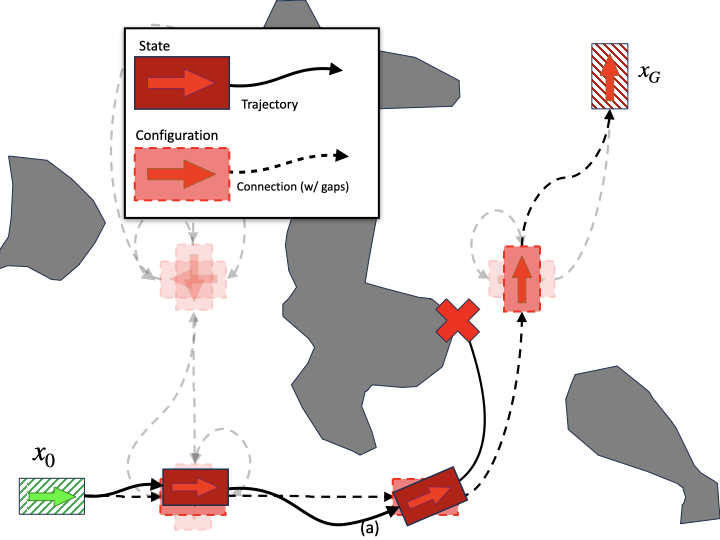}
    \vspace{-.3in}
    \caption{}
    \label{fig:main-c}
    \end{subfigure} 
    \vspace{-.1in}
    \caption{\small (a) A {\it Roadmap with Gaps} consists of vertices (configurations, represented as dotted boxes). The roadmap's directional edges (dotted lines) correspond to where the controller was successfully executed from the source to the target configuration, given some tolerance. (b) For a new planning query $(x_0, x_G)$, the start $q_0$ and goal $q_G$ are added to the roadmap. (c) Due to gaps, naively following the roadmap's shortest path to the goal may not be feasible.}
    \vspace{-.2in}
    \label{fig:main-figure}
\end{figure*}

\subsection{\textbf{\tt RoGuE}: Guiding expansion via a roadmap with gaps}
\label{subsec:rogue}


Algo.~\ref{alg:rogue-tree} describes the \textbf{Ro}admap-\textbf{Gu}ided \textbf{E}xpansion (\textbf{\tt RoGuE})-Tree method, which uses a ``Roadmap with Gaps" to guide the online expansion of an AO Tree Sampling-based Motion Planner given a new query $(x_0,x_G)$.

It first adds vertices $q_0 = \mathbb{M}(x_0)$ and $q_G = \mathbb{M}(x_G)$ to $\mathcal{G}$. 
Then, it adds edges from $q_0$ to all vertices accessible from it, i.e., the set of all vertices $q_i \in \mathcal{V}$ such that applying the controller from $x_0$ for maximum duration $T_\text{max}$ returns a collision-free trajectory $\tau$ with $\mathbb{M}(\tau.\text{end}()) \in \mathcal{B}(q_i,\epsilon)$. 
Similarly, it adds edges to $q_G$ from all vertices that can access the goal set $X_G$.

A \textit{wavefront function} $\mathcal{W}: \mathcal{V} \mapsto \mathbb{R}^+$ is computed over the vertices of the roadmap. Initially, the wavefront value for all vertices is set to infinity. Then, the wavefront value of the goal ($\mathcal{W}(q_G)$) is set to zero, and a backward search is performed. Given the wavefront, every vertex in $\mathcal{V}$ now has the notion of a \textit{successor} associated with it, defined to be its out-neighbor with minimum $\mathcal{W}$ value, i.e., ${\tt Successor}(v) = \argmin_{\mathcal{V}} \{\mathcal{W}(v') \ \vert \ (v,v') \in \mathcal{E}\}$. For a node with an infinite cost-to-goal (because the roadmap does not provide it with a path to the goal), its successor is undefined.

\begin{algorithm}
        \SetAlgoLined
        Add $q_0$, $q_G$  to roadmap $(\mathcal{V,E})$ as start and goal;\\
        $\mathcal{W} \leftarrow \texttt{GET-WAVEFRONT}((\mathcal{V,E}))$; \\
        $\mathrm{T} \leftarrow \{x_0\}$; \\
        \While{termination condition is not met}
        {
        $x_\text{sel} \leftarrow \texttt{SELECT-NODE}(\mathrm{T})$; \\ 
        $u \leftarrow \texttt{RoGuE}(x_\text{sel}, \mathcal{V,W}, \pi)$; \\
        $x_\text{new} \leftarrow \texttt{PROPAGATE}(x_\text{sel},u)$; \\
        \If{$(x_\text{sel} \rightarrow x_\text{new}) \in \mathbb{X}_{\mathrm{f}}$}
        {
        \texttt{EXTEND-TREE}($\mathrm{T}, x_\text{sel} \rightarrow x_\text{new}$);
        }
        }
        \caption{\small {\tt RoGuE-Tree} ($\mathbb{X},\mathbb{U},x_0,x_G,\pi, (\mathcal{V,E}))$}
        \label{alg:rogue-tree}
\end{algorithm}

{\tt RoGuE} (Algo.~\ref{alg:greedy-expand}, Fig~\ref{fig:myopic_rogue}) is an expansion procedure that uses the roadmap wavefront information $\mathcal{W}$ to provide an informed local goal to $\pi$. The first time a tree node $x_\text{sel}$ is selected for expansion, {\tt RoGuE} identifies the closest roadmap node $q_\text{near} \in \mathcal{V}$ according to the distance function $d(\cdot,\cdot$) (Line 2). It then queries the {\tt Successor}($q_\text{near}$) given the wavefront $\mathcal{W}$. If ($q_\text{near}$) is defined, it is provided as the local goal $q_\text{lg}$ to the controller $\pi$ (Lines 3-5). If the successor is not defined, a random local goal in $\mathbb{Q}$ is provided (Line 6-7).

\begin{algorithm}
\SetAlgoLined
\If{first time $x_\text{sel}$ is expanded}
{
$q_\text{near} \leftarrow {\tt ClosestRoadmapNode}(x_\text{sel}, \mathcal{V})$; \\
$q_\text{lg} \leftarrow {\tt Successor}(q_\text{near}, \mathcal{W});$ \\
\If{$q_\text{lg}$ is defined} {
    $u \leftarrow \pi(x_\text{sel},q_\text{lg});$
}
\Else
{
    $u \leftarrow \pi(x_\text{sel},\mathbb{Q}.\text{sample}());$
}
}
\Else
{
$u \leftarrow \mathbb{U}.\text{random-sample}();$
}
\textbf{return } $u$;
\caption{\tt{RoGuE}($x_\text{sel},\mathcal{V,W}, \pi$)}
\label{alg:greedy-expand}
\end{algorithm}


\begin{figure}[t]
    \centering
    \includegraphics[width = .8\linewidth]{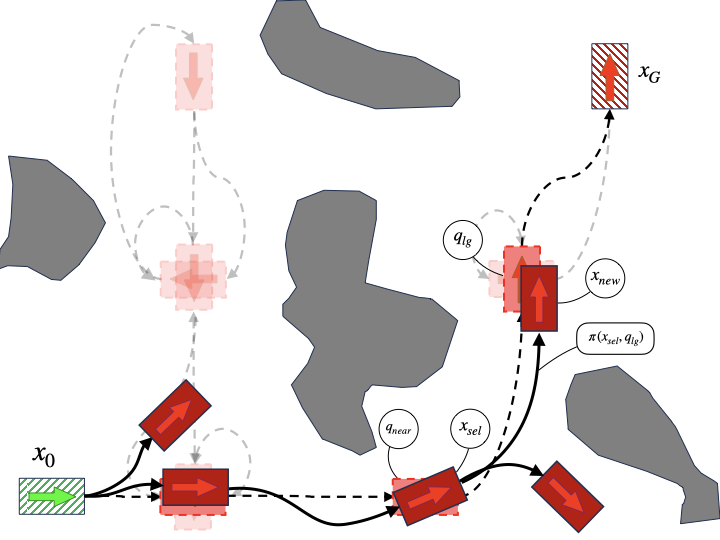}
    \vspace{-.1in}
    \caption{\small Roadmap-Guided Expansion ({\tt RoGuE}): Given the roadmap with gaps in an environment (transparent) and the current planning tree (opaque), {\tt RoGuE} selects informed local goals for the controller $\pi$. The closest roadmap node $q_\text{near}$ to the selected tree node $x_\text{sel}$ is identified. Its successor is passed as the local goal $q_\text{lg}$ to the controller. This expansion adds a new tree node $x_\text{new}$.}
    \label{fig:myopic_rogue}
    \vspace{-.25in} 
\end{figure}

Following the informed operation of the DIRT planner \cite{LB-DIRT}, when the roadmap cost-to-go value of a node is lower than that of its parent's, then the child node is immediately reselected for expansion so that the tree can continue to make progress along a promising path of the roadmap with gaps.  

All (start, goal) pairs in the accompanying experiments below are included as milestones during the roadmap construction to reduce connection time during online planning. Alternatively, the connection of start and goal configurations to the roadmap lends itself to multi-threaded implementations. Or, similarly to the ``roadmap cost-to-go" value definition in Line 2 of Alg. 3, the closest roadmap nodes can be used as surrogates for the corresponding start and goal states. 

{\bf Maintaining Asymptotic Optimality:} As long as all tree nodes have a non-zero probability of being selected for expansion, and the probability of an expansion being successful from a node along the lowest-cost trajectory is non-zero, the sampling-based tree planner remains AO.  The node selection process of {\tt RoGuE} adopts that of the DIRT planner \cite{LB-DIRT}, which, while informed, maintains a positive probability for all tree nodes. Similarly, while {\tt RoGuE} employs an initial set of informed expansions from each node (given the roadmap information), subsequent expansions apply random controls from the set $\mathbb{U}$ (Algo~\ref{alg:greedy-expand}, Line 8-9). In this way, the approach retains the AO properties of DIRT \cite{LB-DIRT}. 

In practice, when the informed processes of {\tt RoGuE} guide the selection and expansion of nodes along high quality solutions, the method achieves an improved convergence rate relative to alternatives that do not have access to the roadmap information as the experimental evaluation below shows. Showing this improvement in convergence rate is an objective of future research efforts for this line of work. 
    \section{Experimental Evaluation}
\label{sec:experiments}

The {\bf robot systems} considered in the evaluation are: (i) an analytically simulated second-order differential-drive, (ii) an analytically simulated car-like vehicle (where $\texttt{dim}(\mathbb{X}) = 5, \texttt{dim}(\mathbb{U}) = 2$), (iii) a MuSHR car \cite{srinivasa2019mushr} physically simulated using MuJoCo \cite{todorov2012mujoco} ($\texttt{dim}(\mathbb{X}) = 27, \texttt{dim}(\mathbb{U}) = 2$), and (iv) a Skydio X2 Autonomous Drone ($\texttt{dim}(\mathbb{X}) = 13, \texttt{dim}(\mathbb{U}) = 4$). For the drone, the distance function $d(\cdot,\cdot)$ operates over the $(x,y,z)$ of its center of mass. For all systems, the parameter $\epsilon$ is set to the same value $0.5$. All planning experiments are implemented using the {\tt ML4KP} software library \cite{ML4KP} and executed on a cluster with Intel(R) Xeon(R) Gold 5220 CPU @ 2.20GHz and 512 GB of RAM.

\begin{figure}[ht!]
    \centering
    \begin{subfigure}{\linewidth}
    \includegraphics[width=0.49\linewidth]{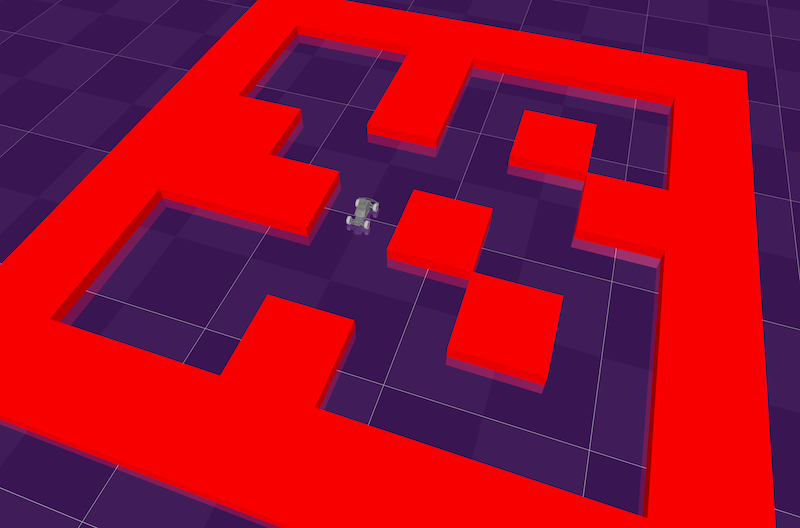}
    \includegraphics[width=0.49\linewidth]{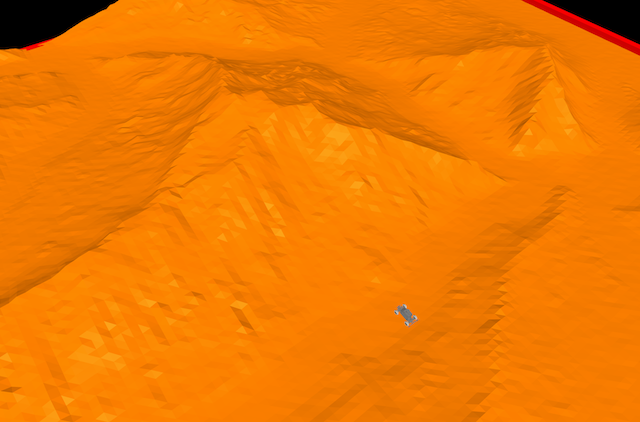}
    \end{subfigure} \\
    \begin{subfigure}{\linewidth}
    \includegraphics[width=0.49\linewidth]{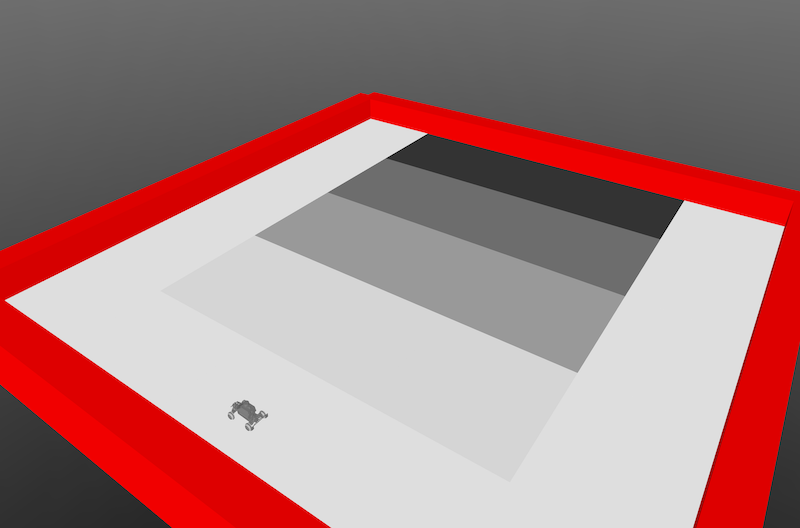}
    \includegraphics[width=0.49\linewidth]{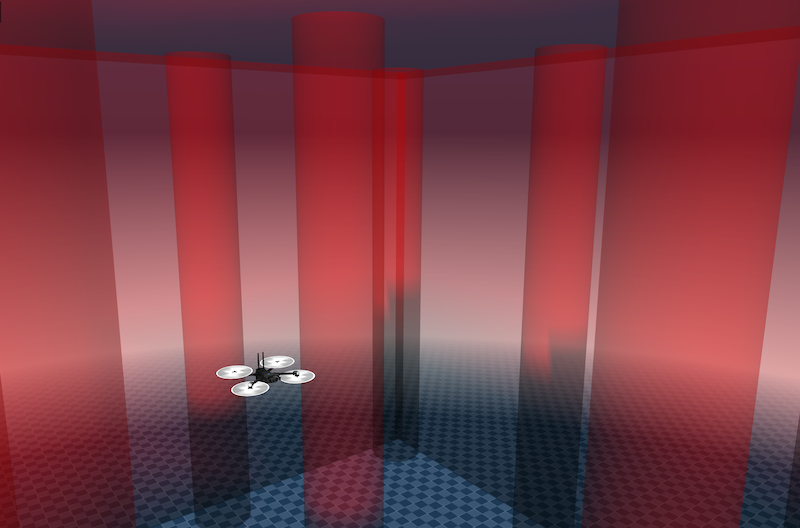}
    \end{subfigure} \\
    \vspace{-.05in}
    \caption{\small Physically simulated benchmarks using MuJoCo. Top: (L-R) {\tt Maze}, {\tt Terrain}. Bottom: (L-R) {\tt Friction}, {\tt Quadrotor.}}
    \vspace{-.2in}
    \label{fig:mujoco-benchmarks}
\end{figure}


Across experiments, the following {\bf metrics} are measured for every planner: (1) Average normalized cost of solutions found over time, and (2) Ratio of experiments for which solutions were found over time. For all planning problems, the path cost is defined as the solution plan's duration. To better reflect the performance of different methods and account for the difficulty of different planning problems, path costs discovered by a method on each planning problem are normalized by dividing by the best path cost ever found for a problem across all planners. 

Two {\bf comparison expansion functions} are considered in the evaluation: (a) \textbf{{\tt Random}} uses a blossom expansion of random controls in $\mathbb{U}$, and (b) \textbf{{\tt RLG}} samples Random Local Goals as input to $\pi$ and outputs the controls returned. In terms of {\bf comparison motion planners}, both an uninformed Rapidly-exploring Randomized Tree (RRT) \cite{lavalle2001randomized} and the informed, AO DIRT planners are considered. For the \textbf{\tt Random} and \textbf{\tt RLG} expansion strategies of DIRT, a blossom of 5 controls is implemented. 

Additional AO planners were considered for experimentation but it was difficult to provide useful results for them. The \textit{Bundle of Edges (BoE)} \cite{shome2021asymptotically} approach failed to find solutions while using a similarly-sized roadmap. The \textit{discontinuity-bounded A*} (dbA*) \cite{honig2022db} relies on motion primitives, which are not available, however, for the robots in the experiments of this paper as they exhibit second-order dynamics. It is not trivial for them to acquire effective motion primitives. 

\begin{figure*}[ht!]
\centering
\begin{subfigure}[b]{\textwidth}
         \includegraphics[width=.235\textwidth]{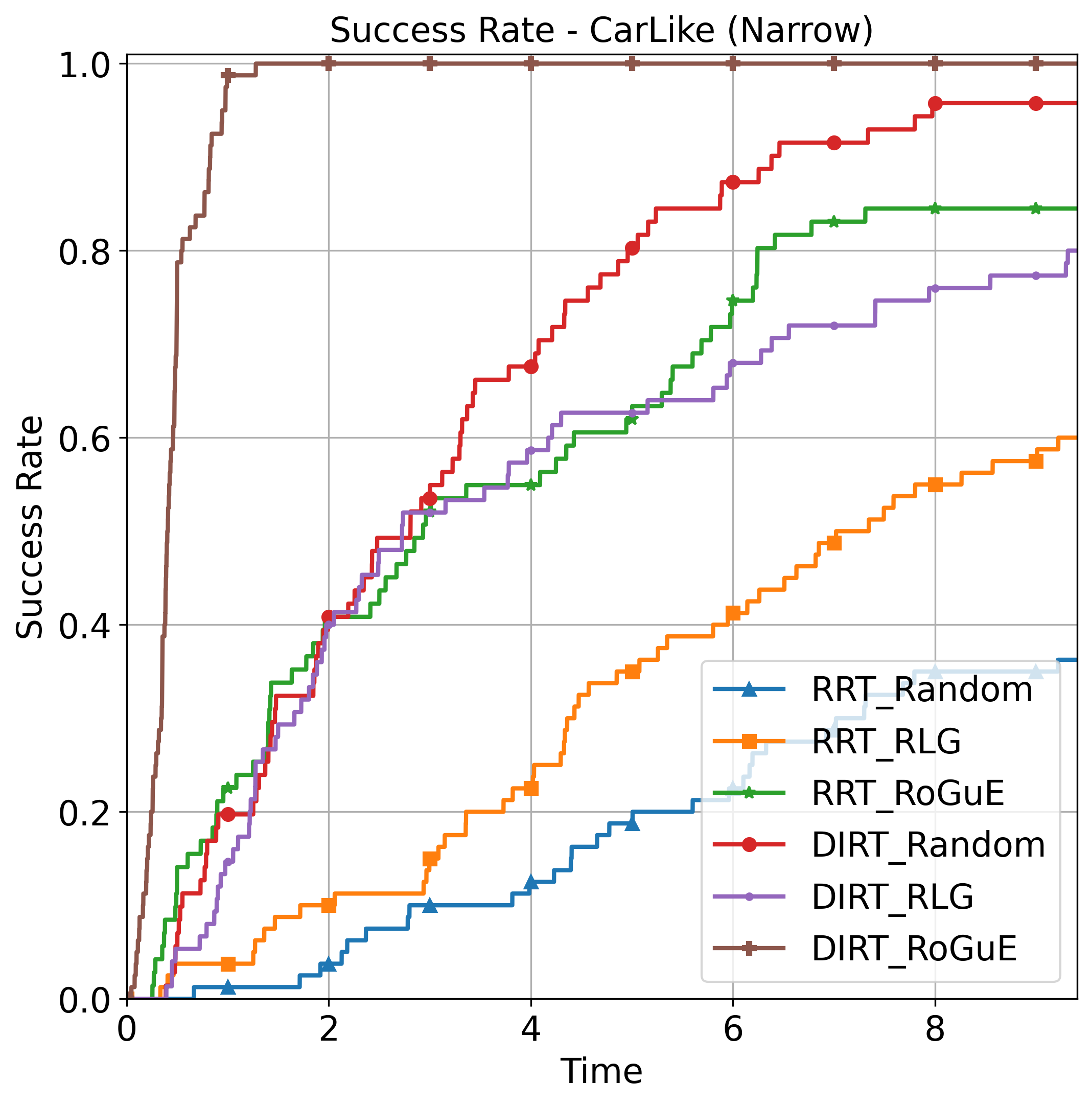}
         \includegraphics[width=.235\textwidth]{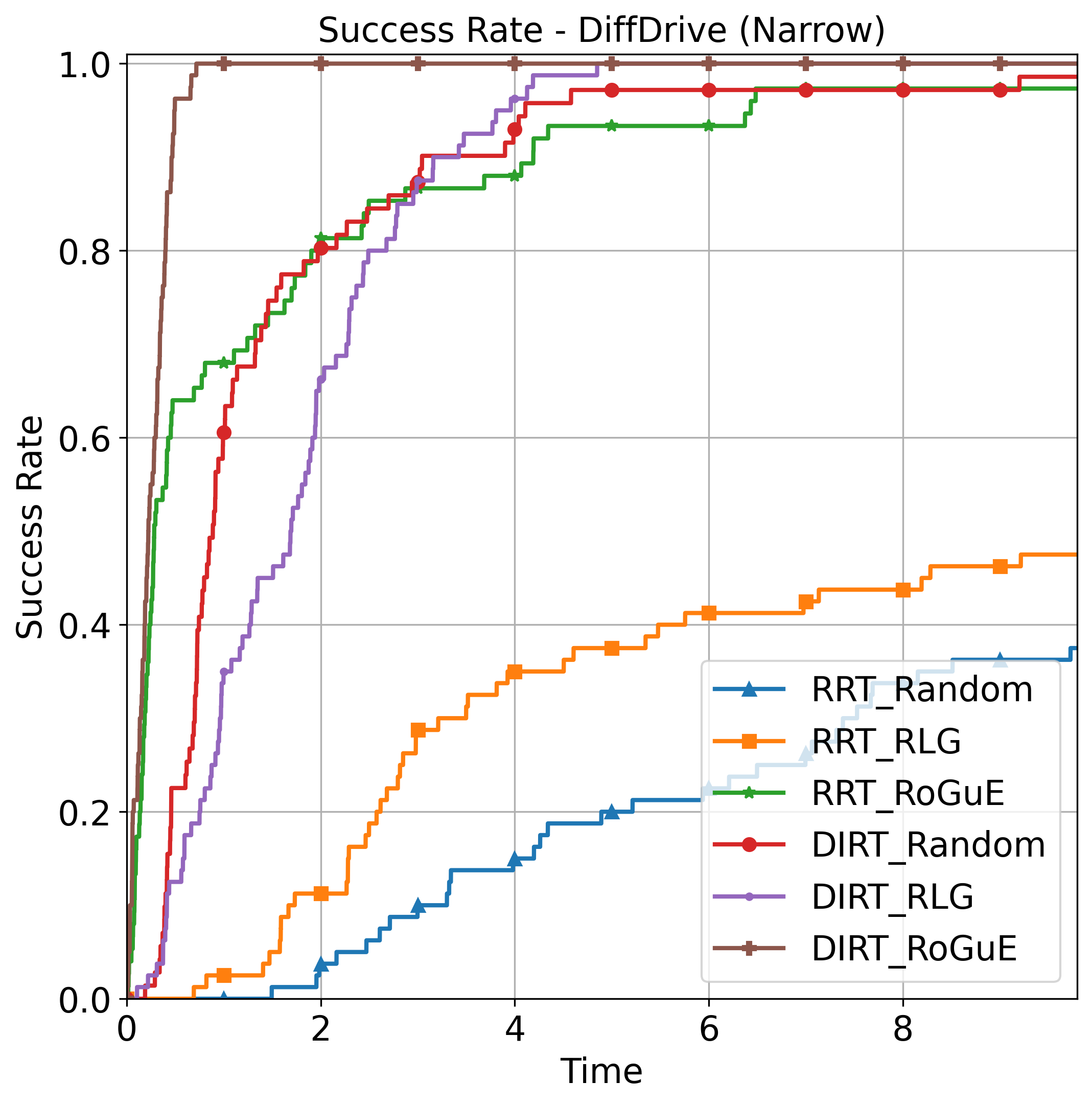}
         \rulesep
         \includegraphics[width=.235\textwidth]{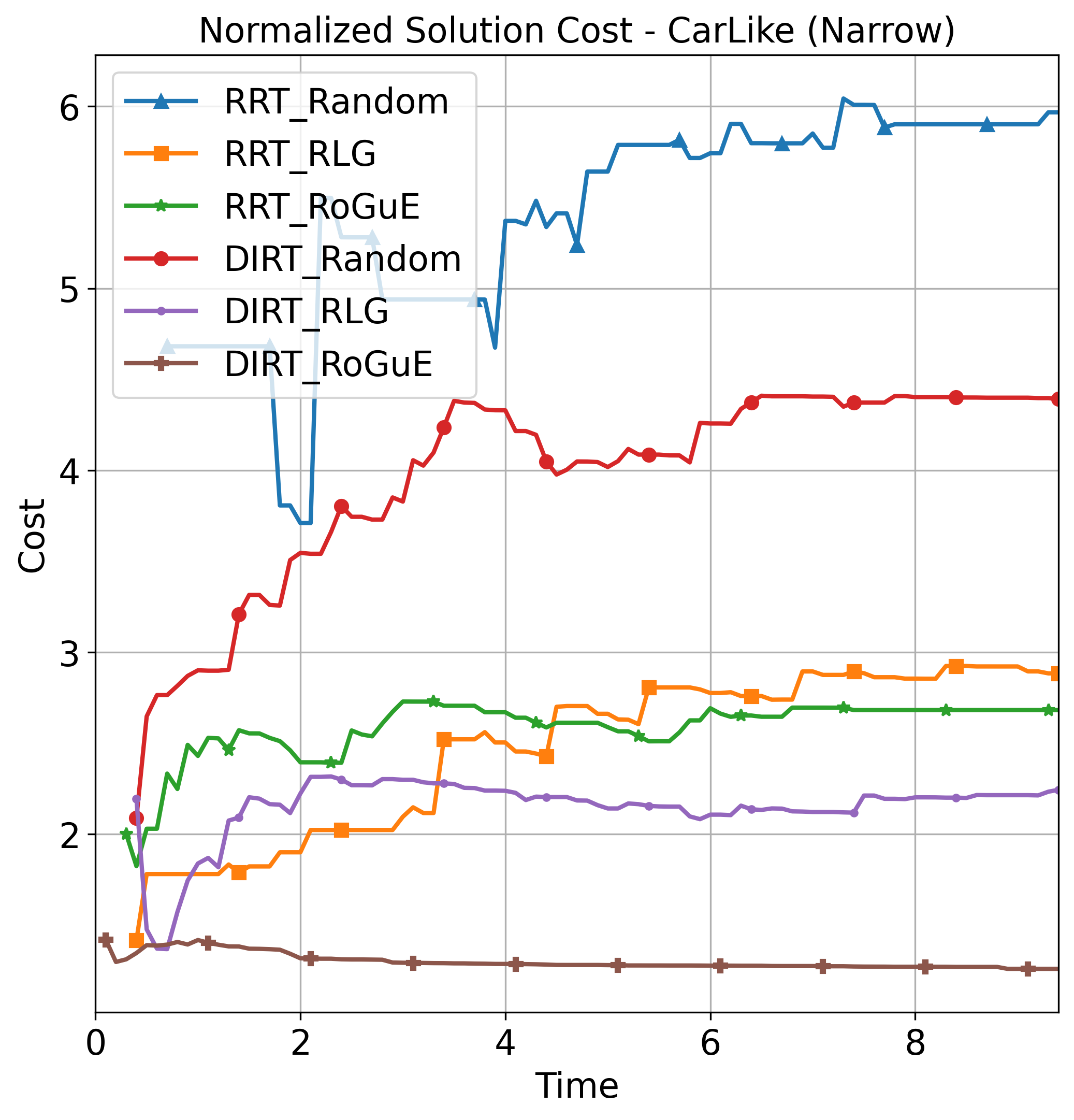}
         \includegraphics[width=.235\textwidth]{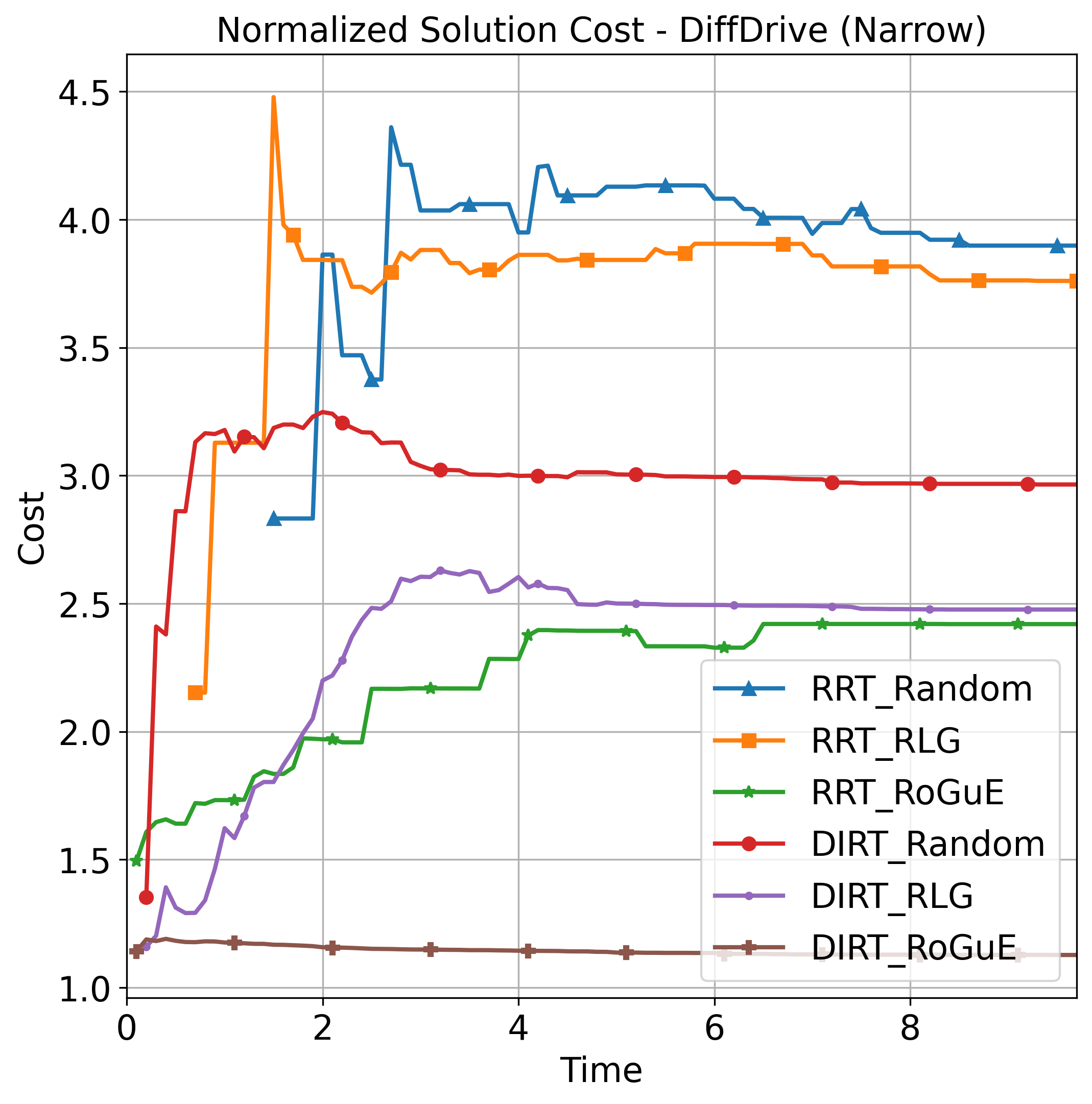}
         \caption{\small Planning results on the \textbf{\tt Narrow} benchmark.}
         \label{fig:narrow-results}
\end{subfigure}\\
\begin{subfigure}[b]{\textwidth}
         \includegraphics[width=.235\textwidth]{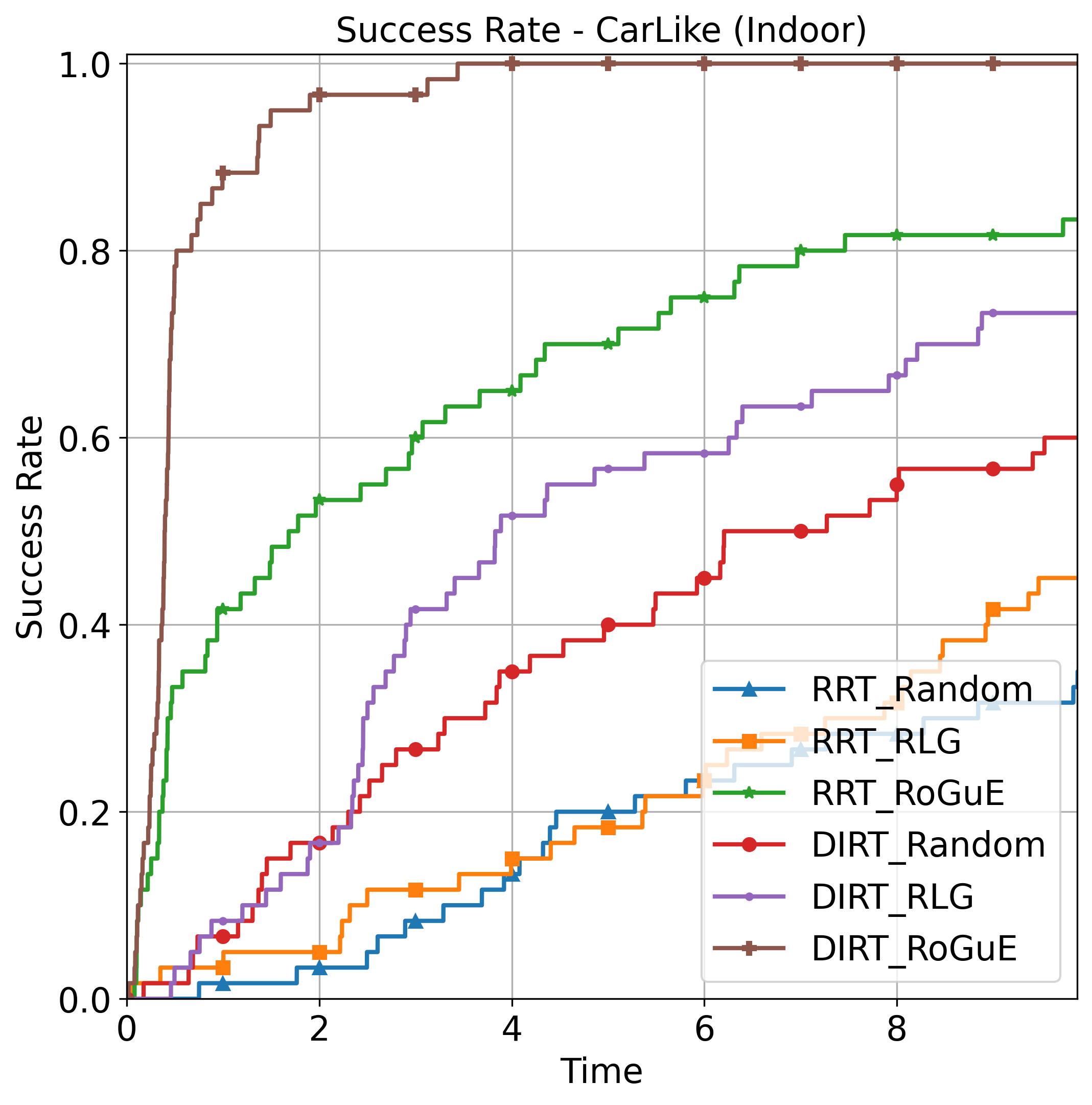}
         \includegraphics[width=.235\textwidth]{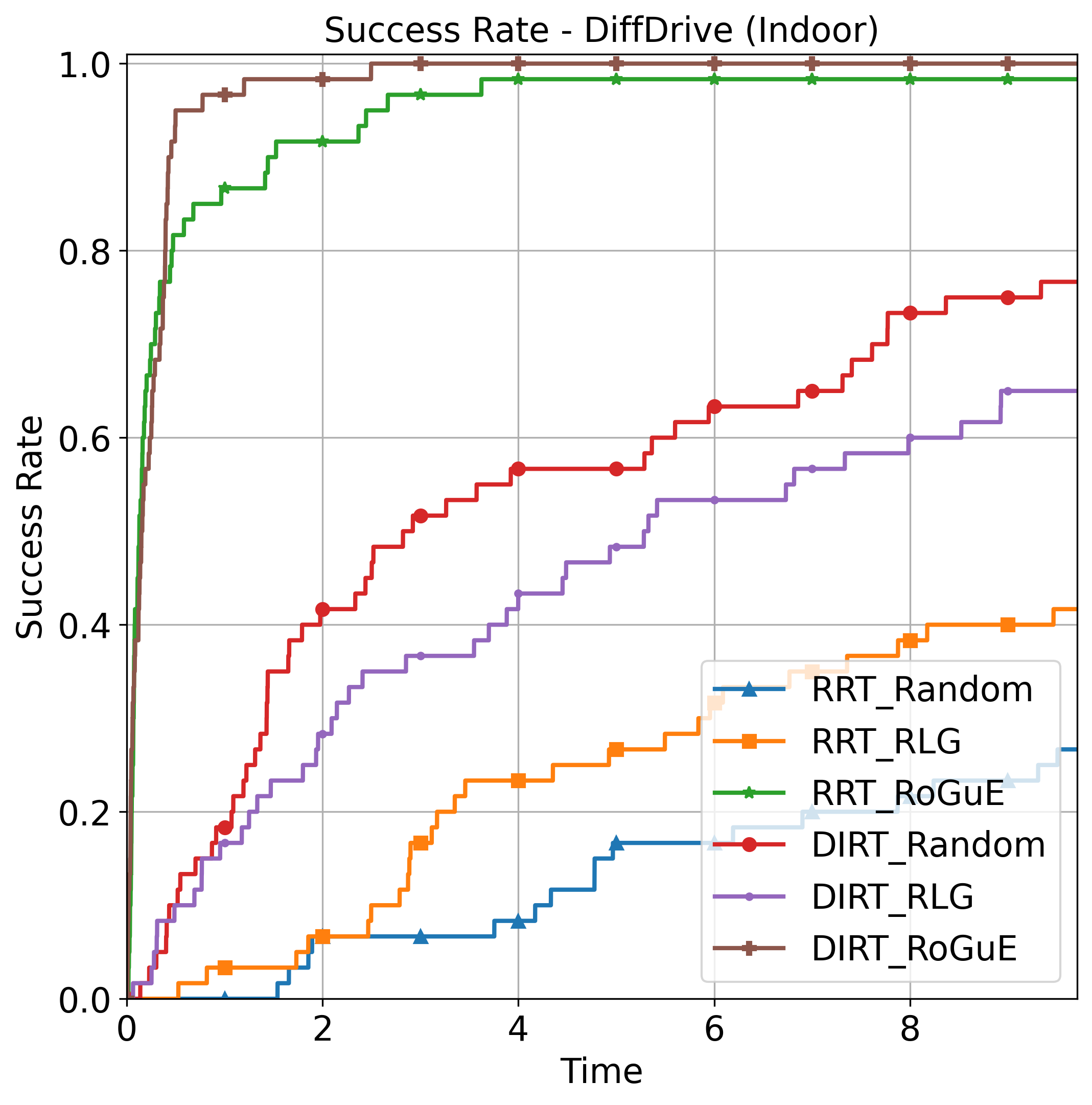}
         \rulesep
         \includegraphics[width=.235\textwidth]{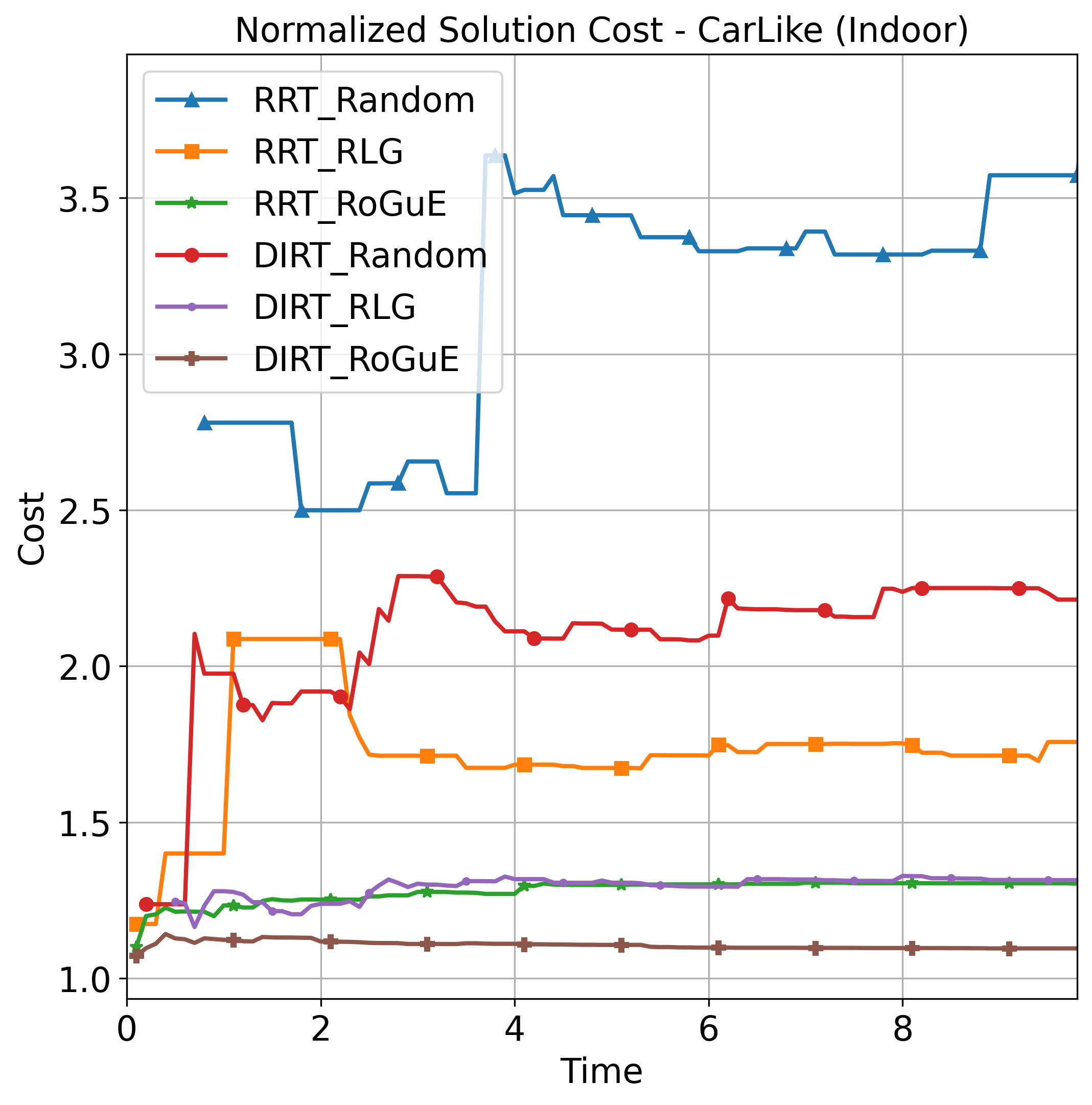}
         \includegraphics[width=.235\textwidth]{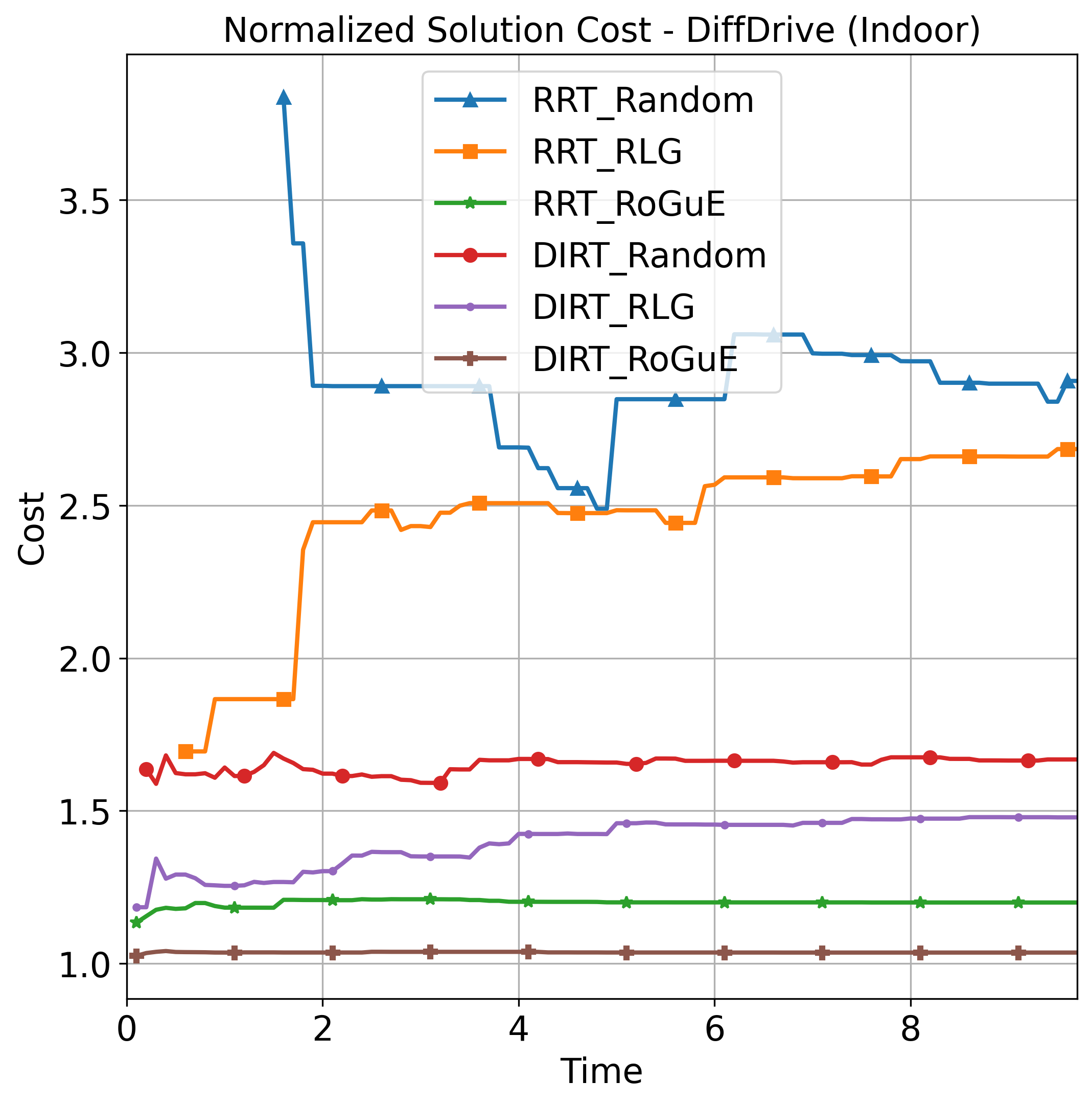}
         \caption{\small Planning results on the \textbf{\tt Indoors} benchmark.}
         \label{fig:indoors-results}
\end{subfigure}\\
\begin{subfigure}[b]{\textwidth}
         \includegraphics[width=.235\textwidth]{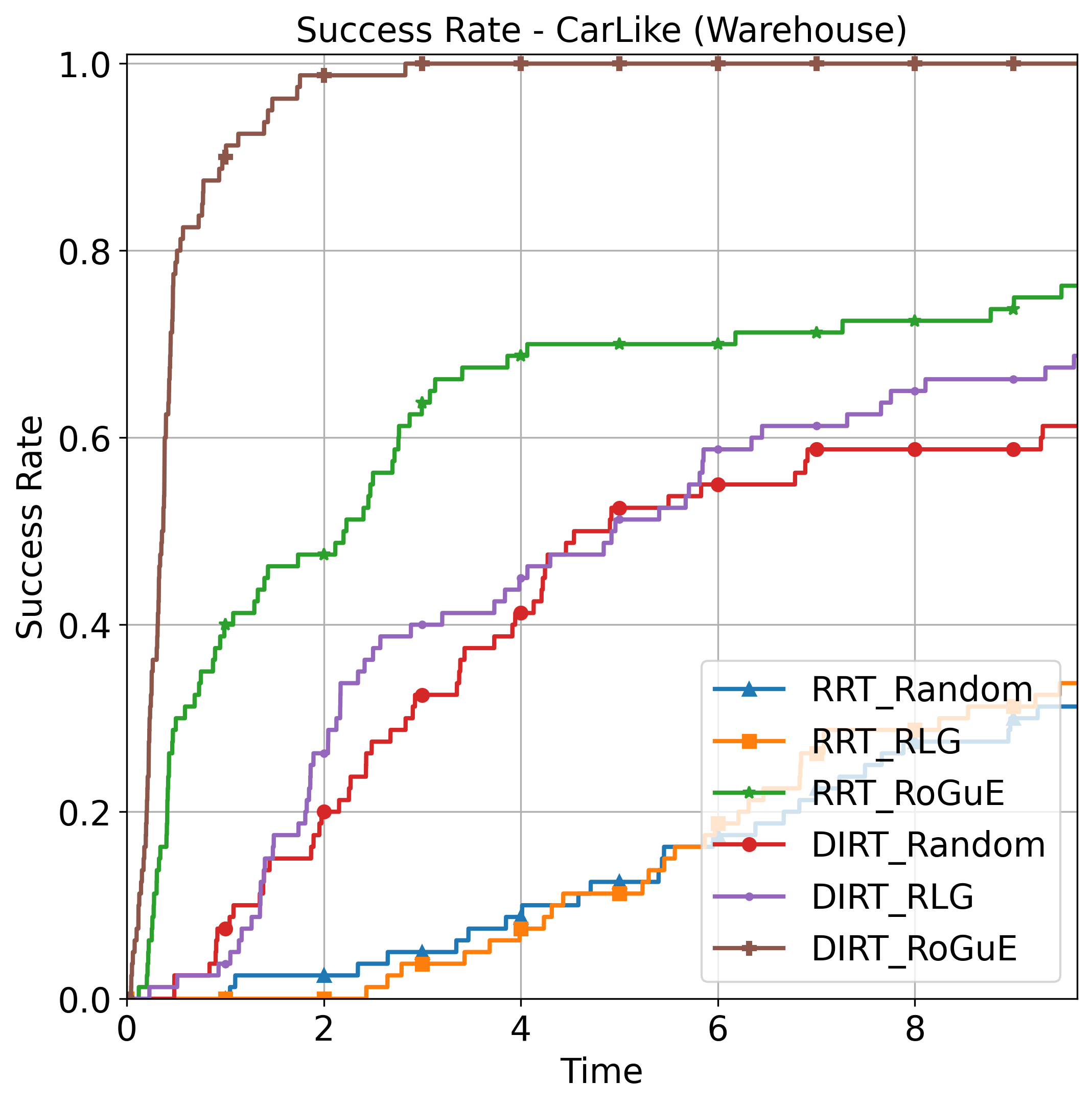}
         \includegraphics[width=.235\textwidth]{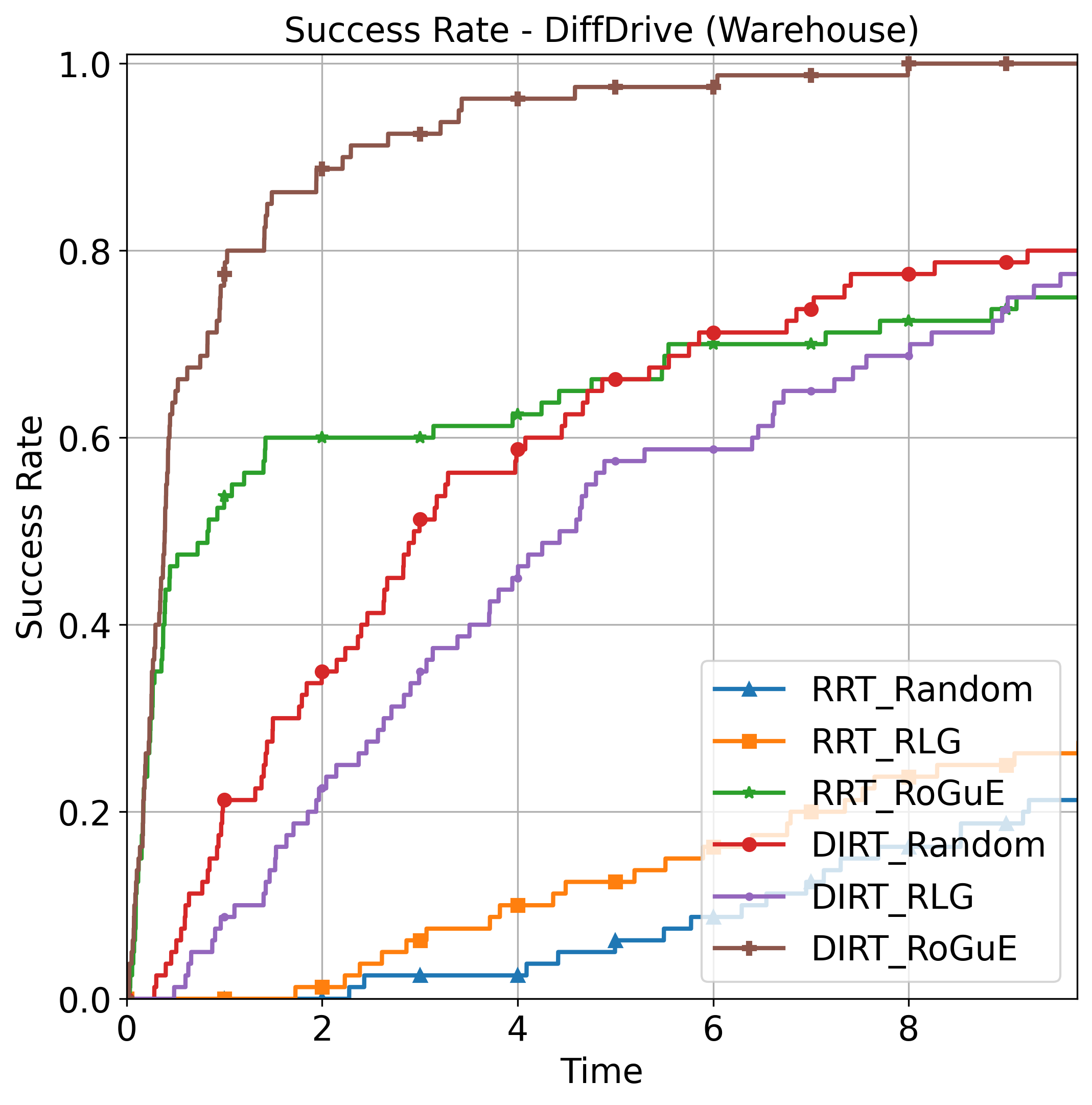}
         \rulesep
         \includegraphics[width=.235\textwidth]{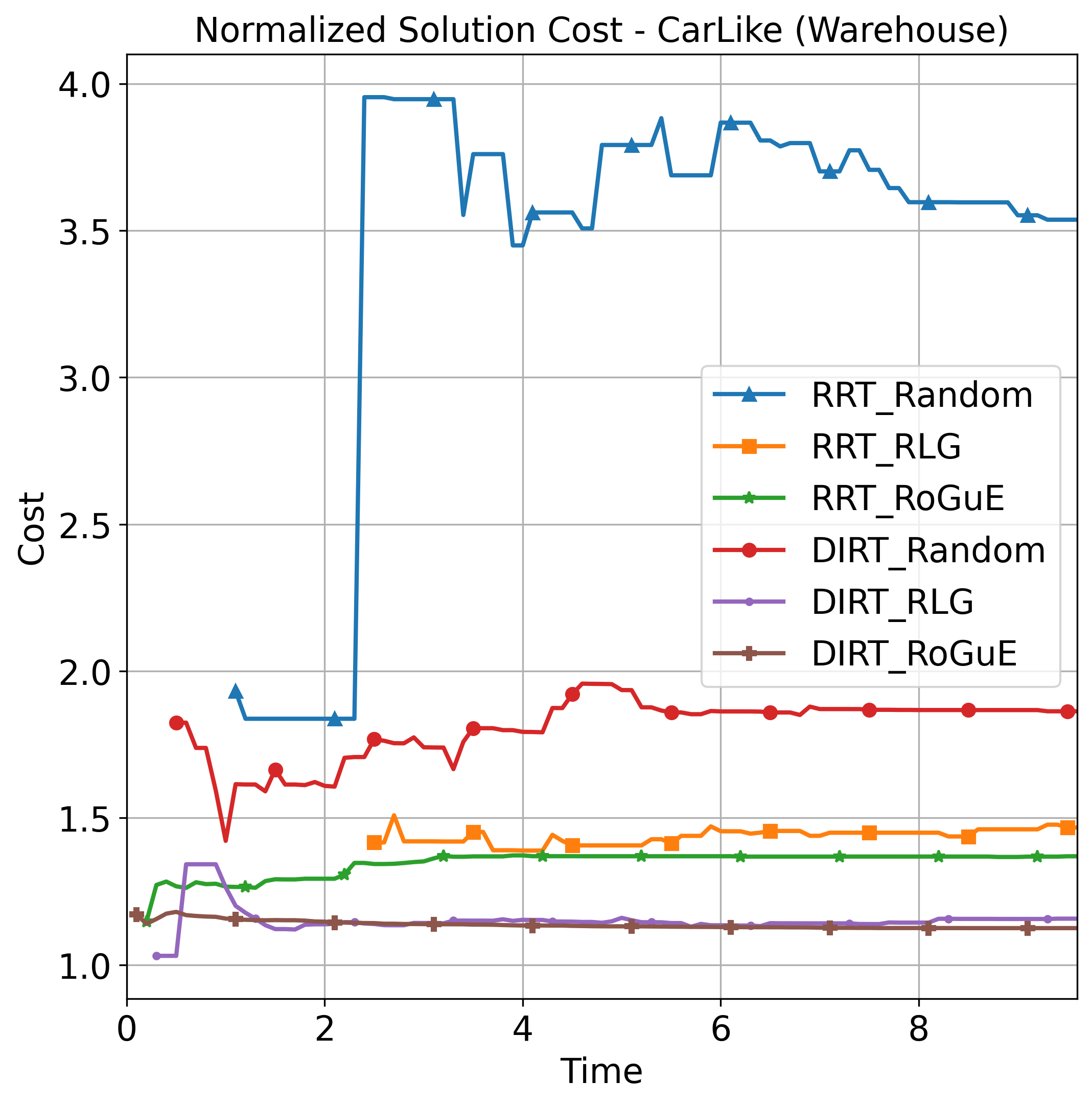}
         \includegraphics[width=.235\textwidth]{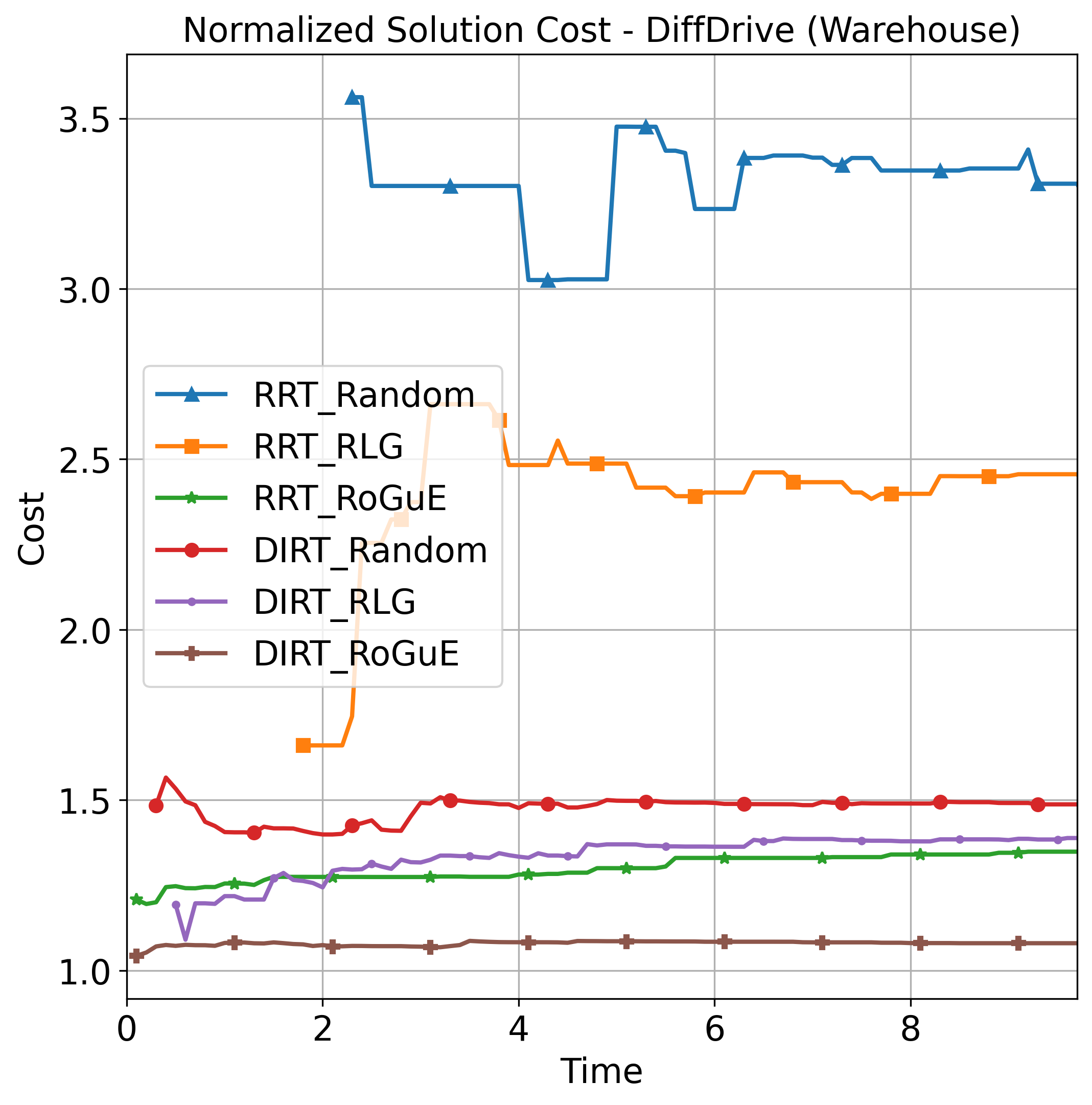}
         \caption{\small Planning results on the \textbf{\tt Warehouse} benchmark.}
         \label{fig:warehouse-results}
\end{subfigure}\\
\vspace{-.05in}
\caption{\small Left and middle-left: higher success rate earlier is better. Right and middle-right: lower path cost is better. As more solutions are discovered, additional solutions for harder problems are discovered, so in some cases, the average cost may increase over time. Each problem instance is run 10 times to account for different random seeds.}
\label{fig:analytical-results}
\vspace{-.2in}
\end{figure*}
 
\subsection{Results on analytically simulated benchmarks} 

In these experiments, the performance of the expansion functions is measured on three sets of planning benchmarks for the analytically simulated vehicular systems: (a) 8 problems in an environment with \textbf{\tt Narrow} passages, (b) 6 problems in the \textbf{\tt Indoor} environment from Arena-bench \cite{kastner2022arena}, and (c) 8 problems in the \textbf{\tt Warehouse} environment from Bench-MR \cite{heiden2021bench}. Figure \ref{fig:analytical-results} provides the numerical results.

Across all benchmarks, for the differential drive and car-like dynamics, \textbf{\tt DIRT-RoGuE} finds the lowest cost solutions overall. It also is the only expansion strategy that returns solutions in all trials. Among the RRT expansion strategies, \textbf{{\tt RRT-RoGuE}} achieves the highest success rate across all benchmarks and the lowest cost solutions overall. \textbf{{\tt RRT-RoGuE}} also achieves comparable performance (both in terms of success rate and solution quality) to the informed \textbf{\tt DIRT-Random} an \textbf{\tt DIRT-RLG} planners. Both \textbf{\tt DIRT-RLG} and \textbf{\tt DIRT-Random} fail to find solutions in all trials on the {\tt Indoor} and {\tt Warehouse} benchmarks and are slow to return solutions relative to  \textbf{\tt DIRT-RoGuE}. 

In the {\tt Narrow} benchmark, the shortest duration trajectories must traverse all the narrow passages. So although {\tt DIRT-Random} and {\tt DIRT-RLG} find solutions in all trials for the second-order differential drive, {\tt DIRT-RoGuE} returns significantly lower-cost solutions much earlier.

\textbf{Comparison to Kinematic Planning and Path Following:} A na\"ive alternative to the proposed solution is to use the configurations along a path on the roadmap as consecutive local goals for a path following controller. For the car-like system, a pose-reaching controller~\cite{corke2011robotics} was employed to drive the robot to a given pose $\left(\begin{array}{c}v\\ \omega \end{array}\right) =\left(\begin{array}{c}k_\rho \rho\\ k_\alpha \alpha + k_\beta \beta \end{array}\right)$, where each $k$ parameter is a gain term, $\rho$ and  $\alpha$ are the distance and bearing to the local goal respectively, and $\beta$ is the angle difference between $\alpha$ and the current angle. The controller is tested on paths retrieved from the Roadmap with Gaps for the benchmarks of Fig~\ref{fig:analytical-results}. Only 2 such executions, however, returned collision-free trajectories. This is due to (a) the environments containing multiple narrow passages and (b) the paths returned by the roadmap still contain ``gaps'' that the controller cannot easily negotiate. This motivates the proposed solution and alternatives that are similarly obstacle-aware and which reason about the robot's dynamics.


\begin{figure*}[ht!]
\centering
\begin{subfigure}[b]{\textwidth}
         \includegraphics[width=.235\textwidth]{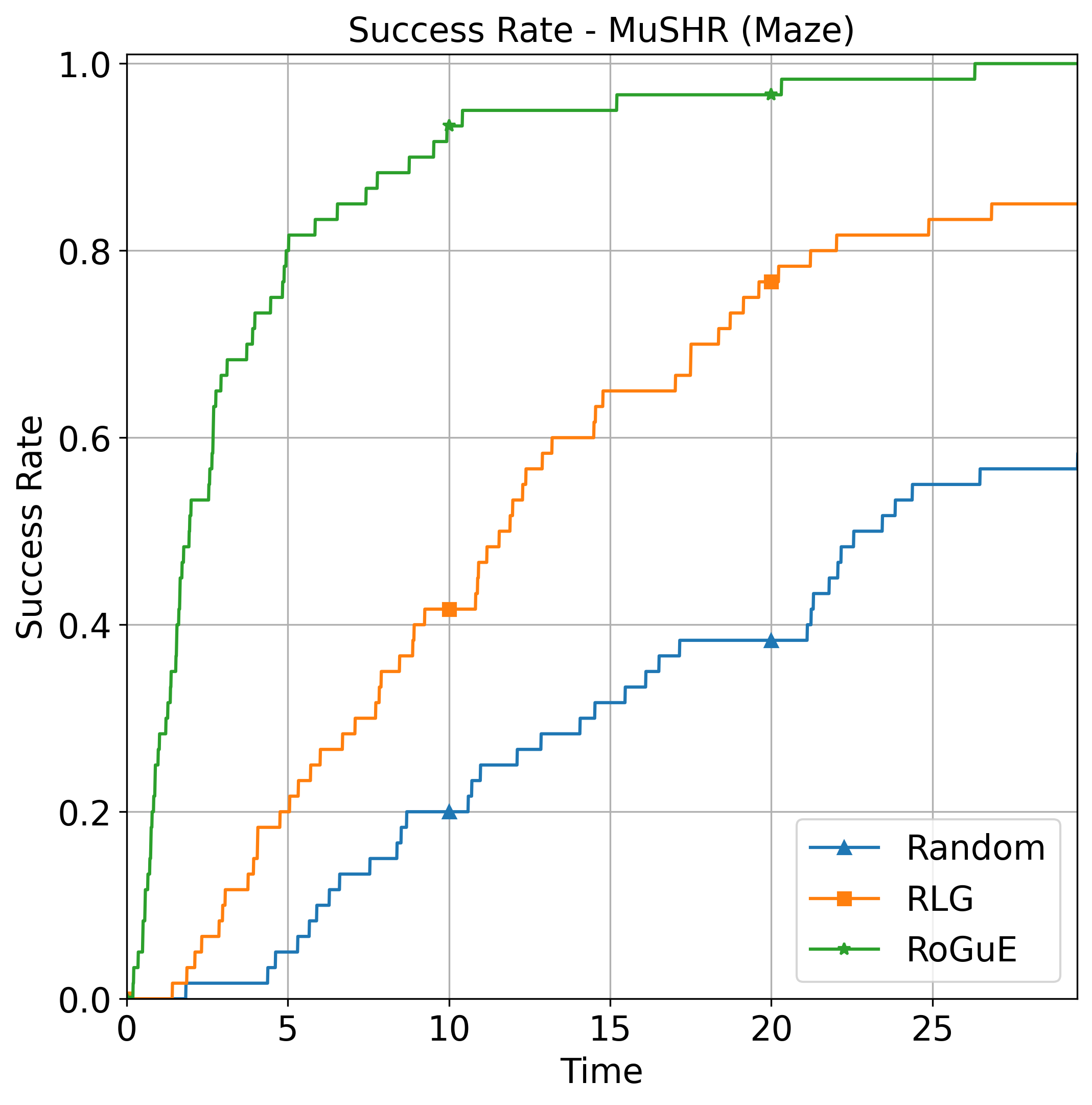}
         \hspace{.01\textwidth}
         \includegraphics[width=.235\textwidth]{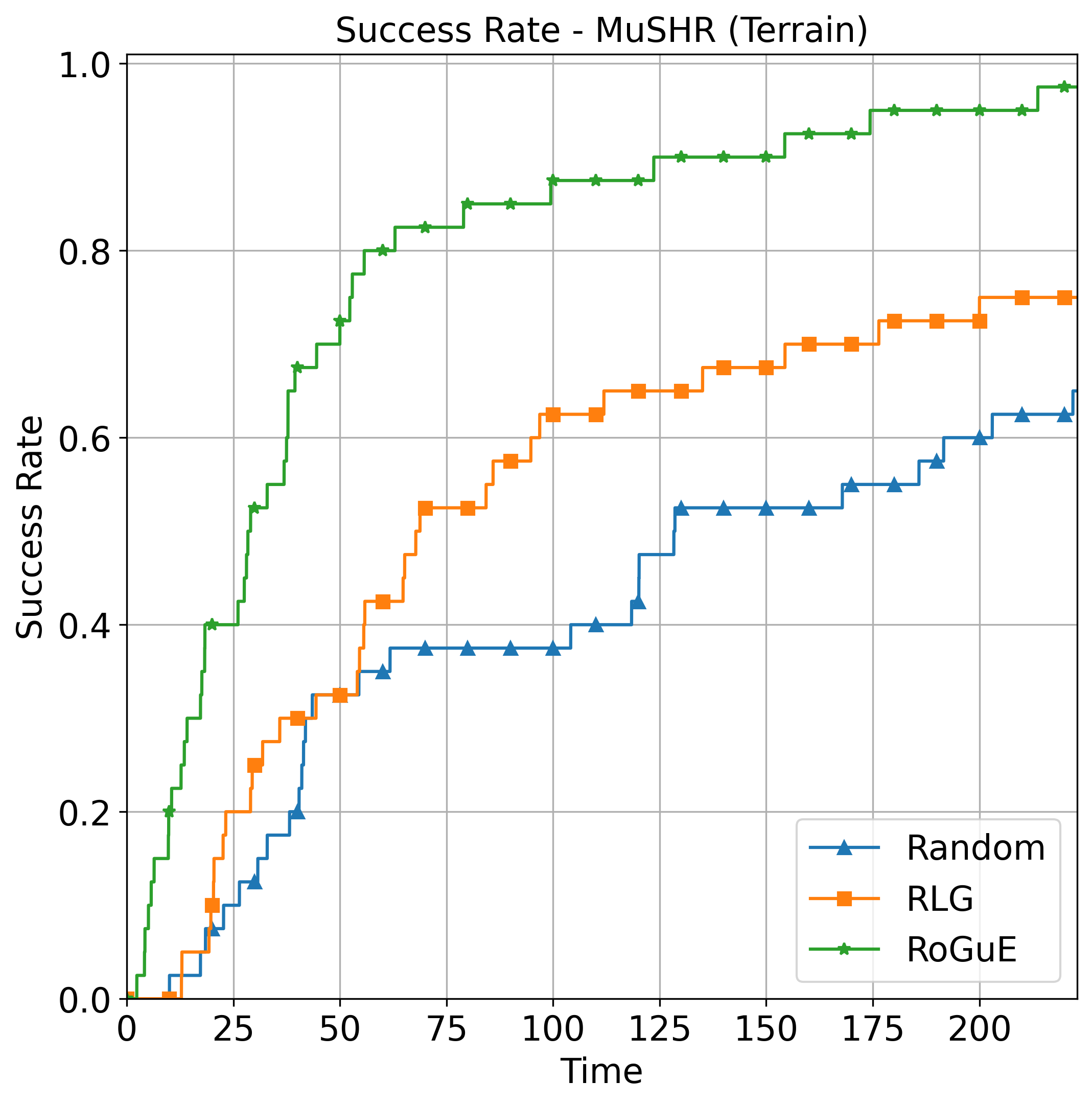}
         \hspace{.01\textwidth}
         \includegraphics[width=.235\textwidth]{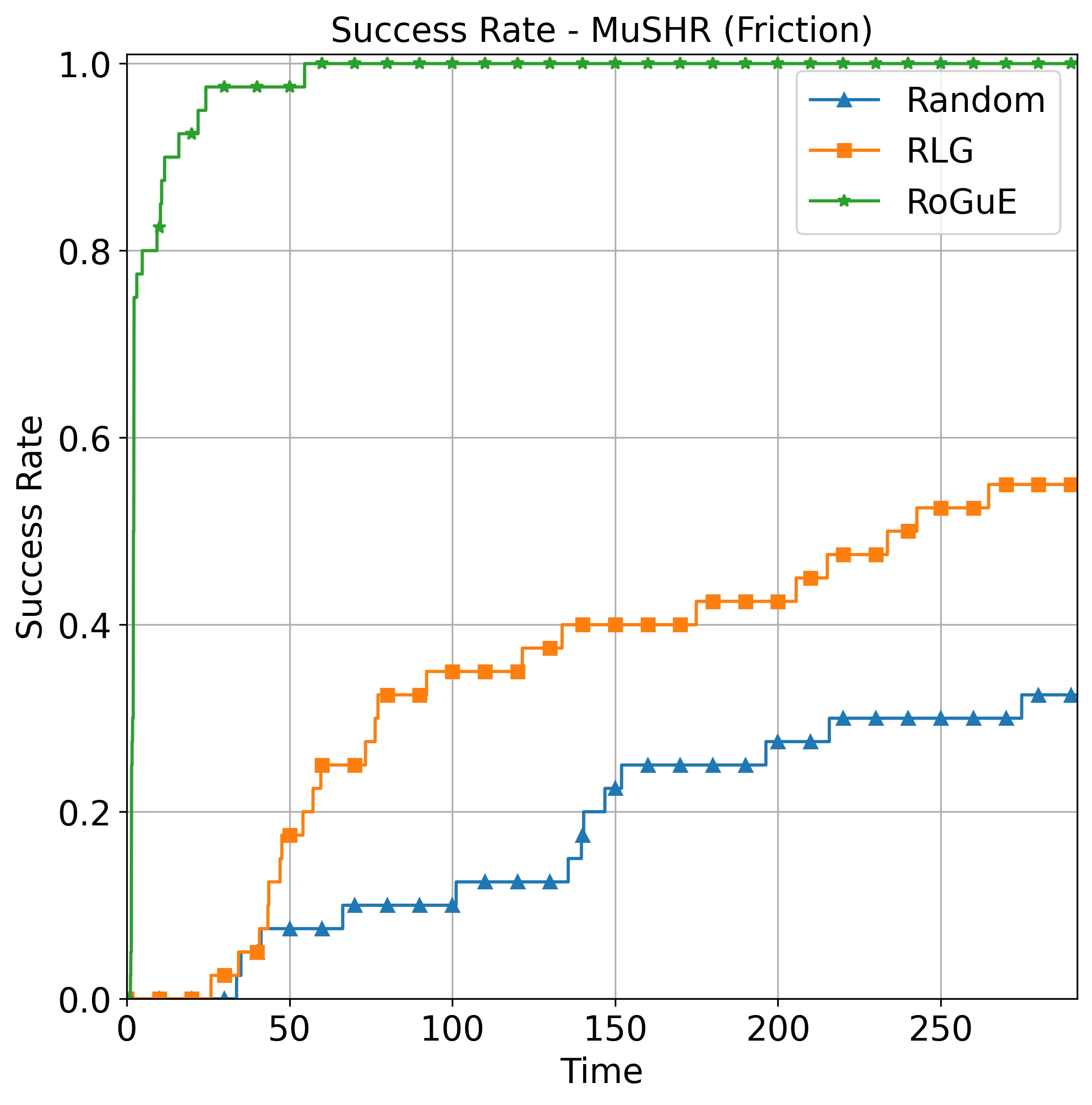}
         \hspace{.01\textwidth}
         \includegraphics[width=.235\textwidth]{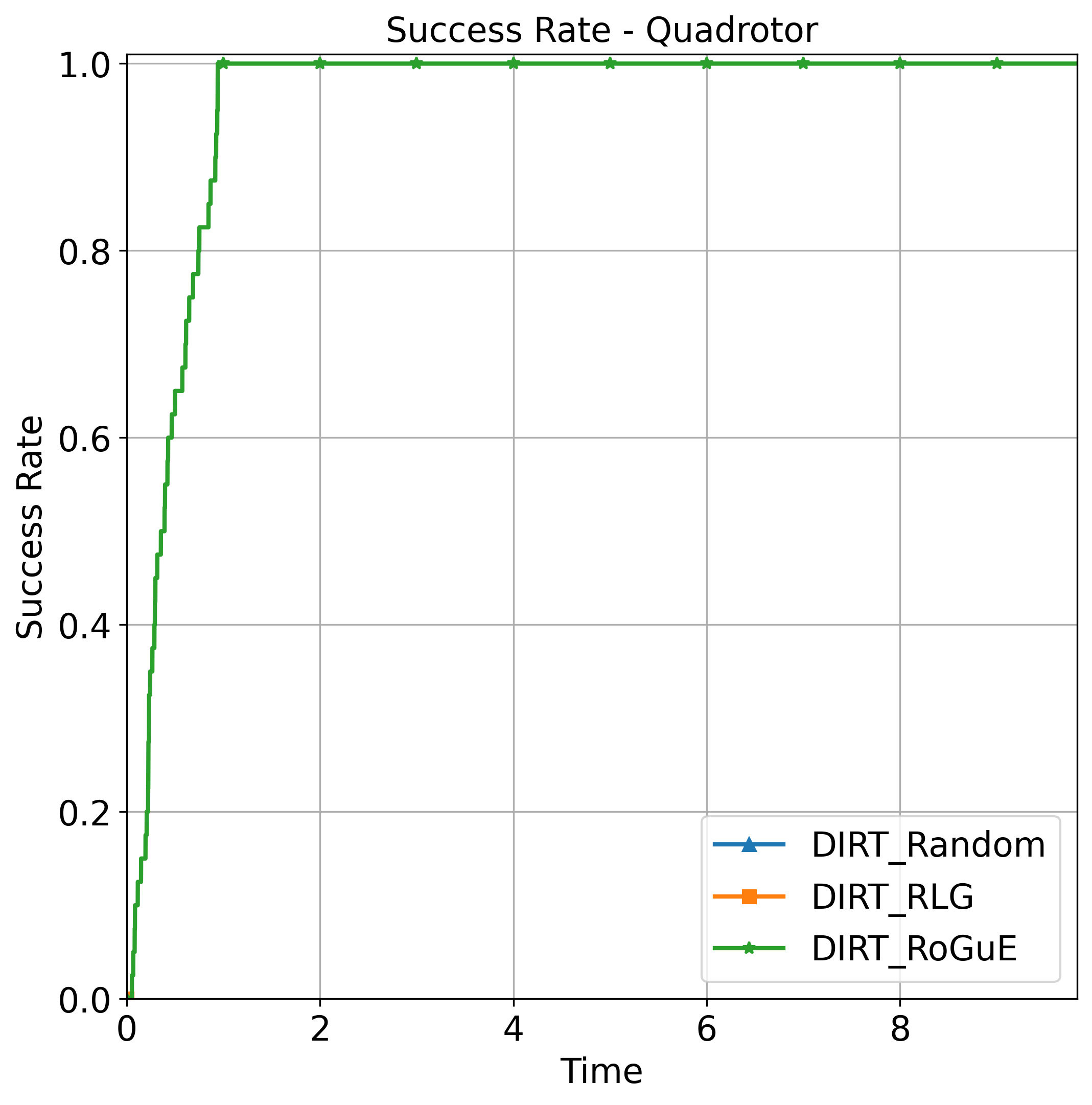}
         \put(-70pt,85pt){\smaller \textcolor{red}{$\mathtt{Random}$ and $\mathtt{RLG}$ do}}
         \put(-68pt,75pt){\smaller \textcolor{red}{not return solutions}}
         \caption{\small Success rates of the planners on each physically simulated benchmark. Higher success rate earlier is better.}
         \label{fig:mujoco-success}
\end{subfigure}\\
\vspace{0.05in}
\begin{subfigure}[b]{\textwidth}
         \includegraphics[width=.235\textwidth]{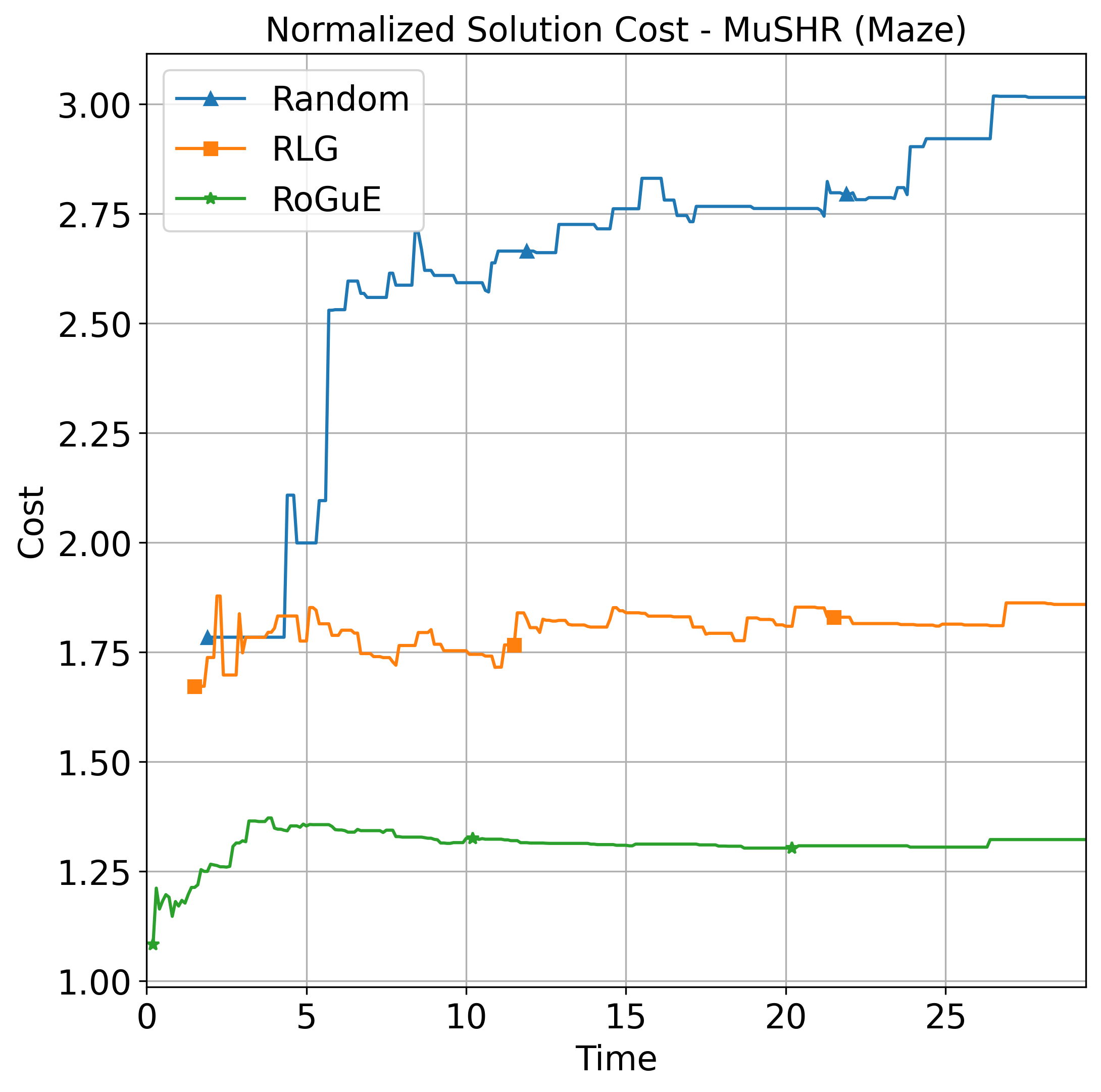}
         \hspace{.01\textwidth}
         \includegraphics[width=.235\textwidth]{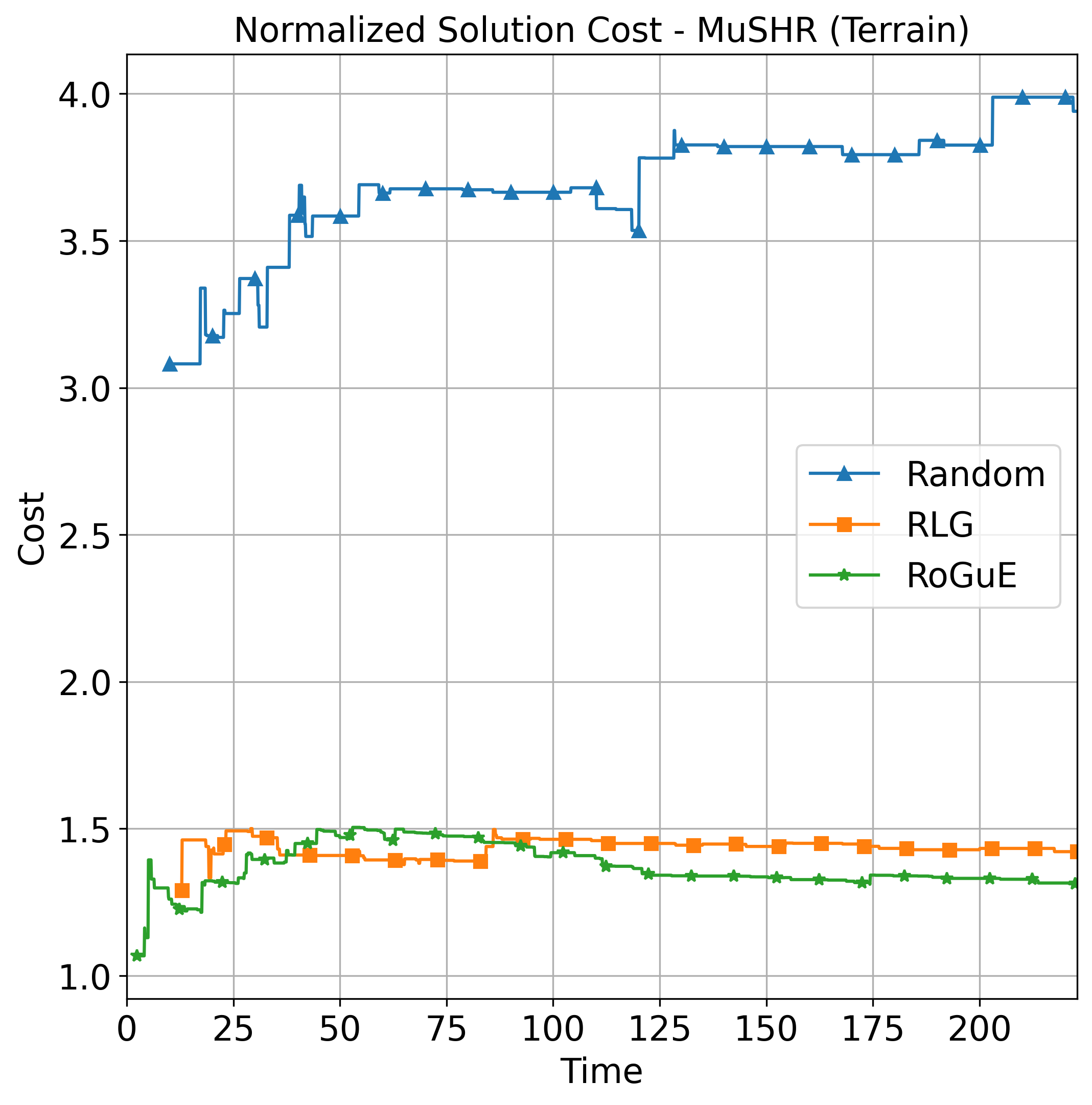}
         \hspace{.01\textwidth}
         \includegraphics[width=.235\textwidth]{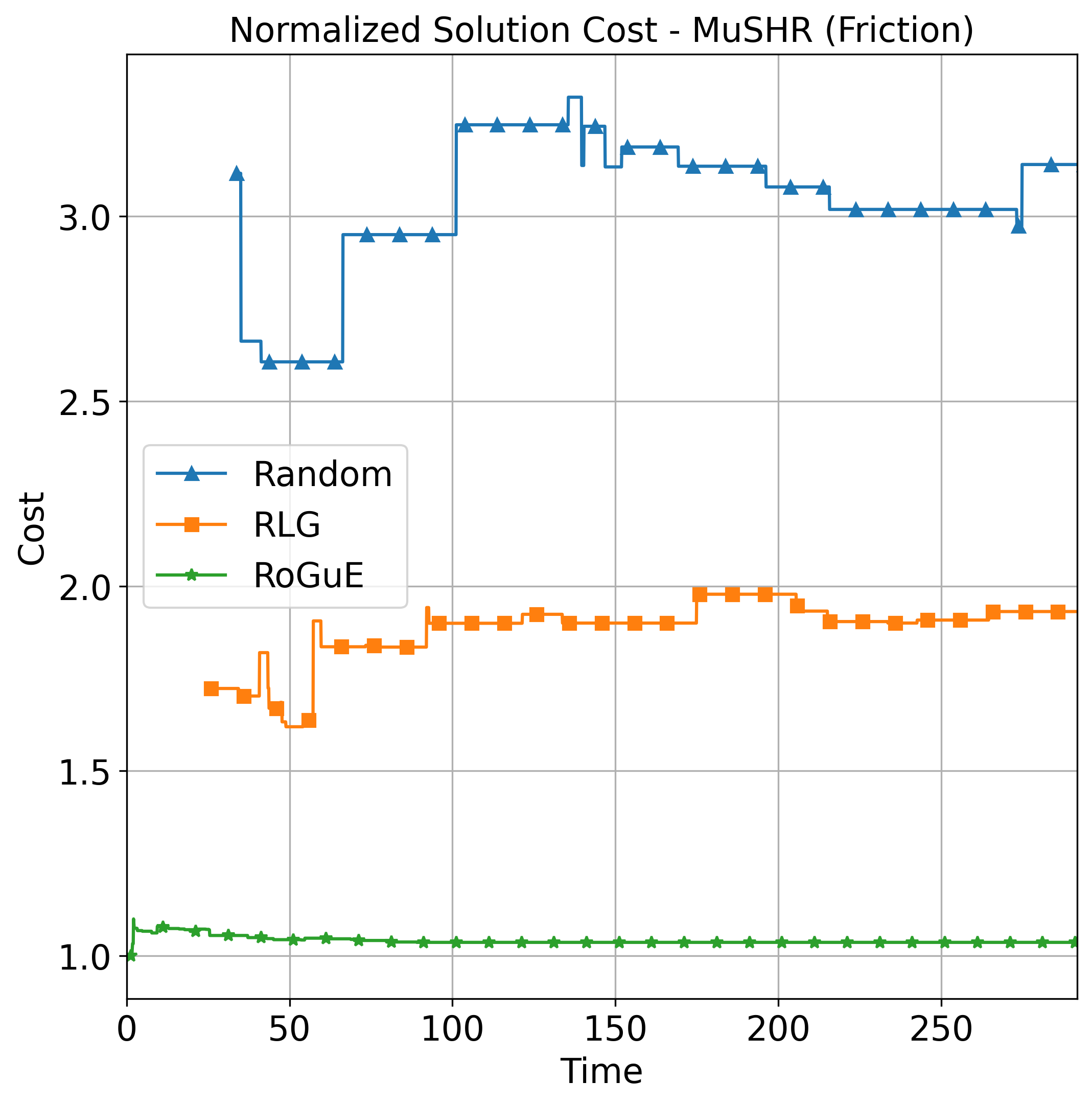}
         \hspace{.01\textwidth}
         \includegraphics[width=.235\textwidth]{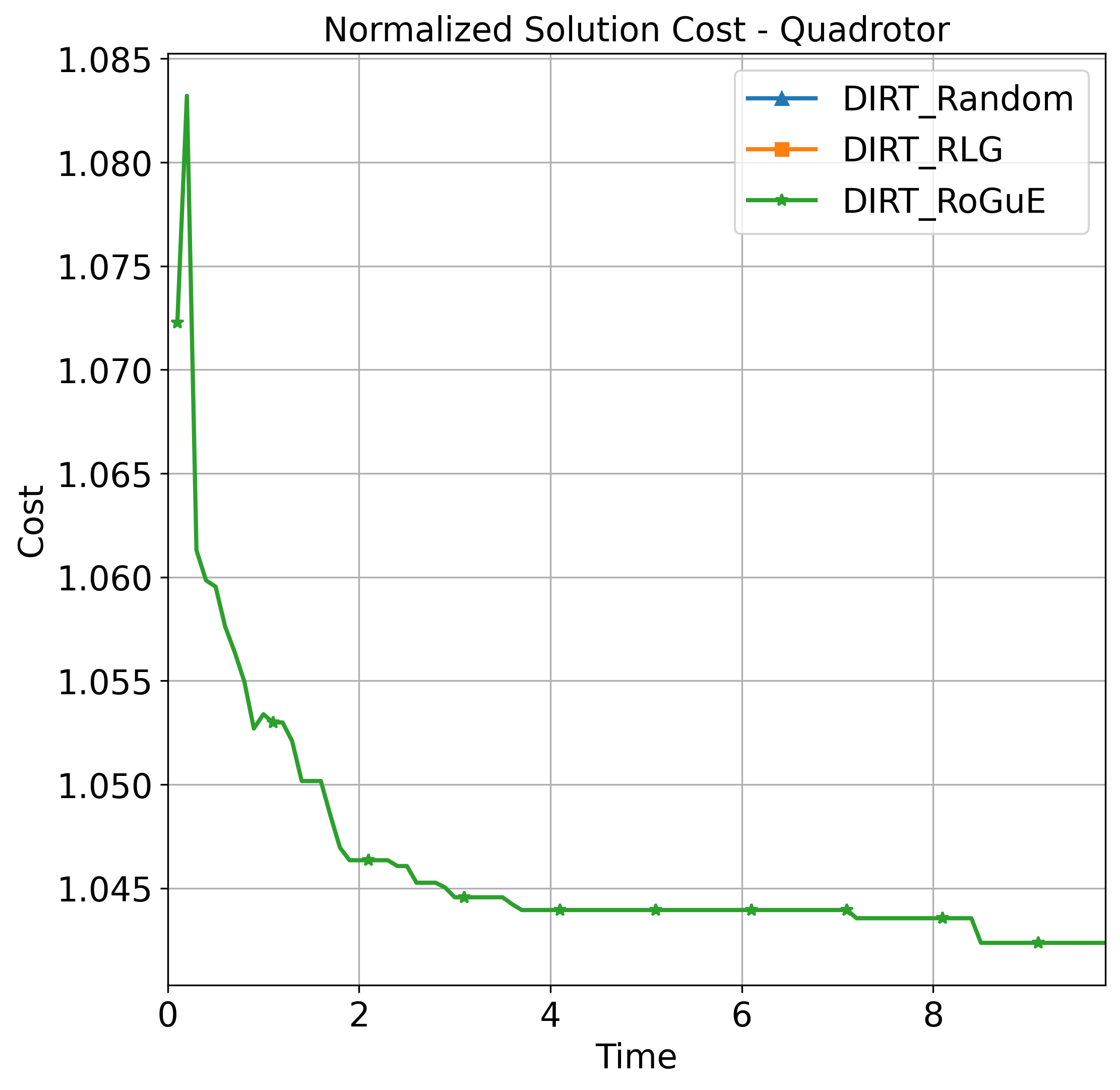}
         \put(-70pt,78pt){\smaller \textcolor{red}{$\mathtt{Random}$ and $\mathtt{RLG}$ do}}
         \put(-68pt,68pt){\smaller \textcolor{red}{not return solutions}}
         \caption{\small Solution quality of the planners on the physically simulated benchmarks. Lower values are better. As more solutions are discovered, harder instances are resolved, so in some cases, the average cost may increase over time.}
         \label{fig:mujoco-cost}
\end{subfigure}\\

\caption{\small Evaluation on the physically simulated systems via MuJoCo. Each instance is executed 10 times with different random seeds. For the {\tt Quadrotor}, the {\tt Random} and {\tt RLG} strategies consistently fail to return solutions and hence their results are not displayed.}
\label{fig:mushr-results}
\vspace{-.1in}
\end{figure*}

\subsection{Results on physically simulated benchmarks} 

The performance of the expansion functions is measured on the following benchmarks for MuSHR: (a) navigating a {\tt Maze} from {\tt D4RL} \cite{fu2020d4rl}, (b) an environment with uneven {\tt Terrain} features \cite{stachowicz2023fastrlap}, (c) and an environment with different friction values ({\tt Friction}). Note that the controller is trained in an empty environment with flat terrain and friction. The roadmap captures the traversability of different parts of the environment using the controller. In the {\tt Quadrotor} benchmark, the X2 drone must navigate an indoor environment with pillars and air pressure effects. 

Fig.~\ref{fig:mujoco-benchmarks} visualize the different environments in MuJoCo, and Fig.~\ref{fig:mushr-results} provides the experimental results.  Only the DIRT motion planner is reported in these experiments as the RRT-based solutions cannot find a solution within the allotted times. Since each call to the MuJoCo engine is expensive, all expansion strategies use a blossom $k=1$.

{\tt RoGuE} finds the lowest cost solution across all benchmarks. Both \textbf{\tt Random} and \textbf{\tt RLG} consistently fail to find solutions across trials. Using the learned controller in \textbf{\tt RLG} does lead to improved solution quality relative to \textbf{\tt Random}. In both the {\tt Maze} and {\tt Friction} benchmarks, {\tt RoGuE} leverages the Roadmap with Gaps to find solutions across trials quickly. {\tt RoGuE} also returned the most solutions in the {\tt Terrain} benchmark given the same planning budget. In the {\tt Quadrotor} benchmark, {\tt RoGuE} is the only expansion method that discovers any solutions due to the tight placement of obstacles and the speed of the X2 drone.

\noindent \textbf{Comparison to Deep RL solutions: } Table~\ref{tab:drl} evaluates two purely DRL-based strategies trained on the benchmarks of Fig~\ref{fig:mujoco-benchmarks} in terms of success rate {\tt Succ}, as well as offline cost ({\tt Offl}, in terms of $\#$ of calls made to the MuJoCo engine). The offline cost of the DRL methods is reported when their best performance is observed, and the success rate does not improve after training for longer. The online costs {\tt Onl} for the sampling-based planners are also reported. The online cost for the RL solutions is minimal. {\tt SAC+HER} trains a goal-conditioned controller $u = \pi(x,q)$ directly in the planning environment. The approach achieves a low success rate relative to the proposed kinodynamic planning solution. The low success on  {\tt Quadrotor} can be attributed to the difficulty in jointly learning the dynamics and obstacle avoidance. Due to simpler dynamics, the performance is better on the {\tt Maze} and {\tt Friction} benchmarks. {\tt H-SAC+HER} follows a hierarchical approach similar to {\tt RoGuE} by training a policy to predict local goals for the controller to reach at every step, i.e., $q_\text{lg} = \phi(x)$. This slightly improves the success rate relative to {\tt SAC+HER} on {\tt Quadrotor}. The success rate is comparable to {\tt SAC+HER} in the {\tt Maze} environment, while it is lower on {\tt Friction}. This indicates the difficulty of DRL strategies in learning an informed local goal procedure, which the Roadmap with Gaps captures via offline computation. 

\begin{table*}[ht!]
\vspace{-.1in}
  \centering
\begin{tabular}{|c|ccc|cc|cc|cc|ccc|}
\hline
          & \multicolumn{3}{c|}{\tt RoGuE}                                           & \multicolumn{2}{c|}{\tt SAC+HER}          & \multicolumn{2}{c|}{{ \tt H-SAC+HER}}   & \multicolumn{2}{c|}{{ \tt Random}}   & \multicolumn{3}{c|}{{ \tt RLG}}   \\ \hline
\textbf{Benchmark}  & \multicolumn{1}{c|}{\tt Offl}      & \multicolumn{1}{c|}{\tt Onl}    & {\tt Succ}               & \multicolumn{1}{c|}{\tt Offl}    & {\tt Succ} & \multicolumn{1}{c|}{\tt Offl} & {\tt Succ} & \multicolumn{1}{c|}{\tt Onl} & {\tt Succ} & \multicolumn{1}{c|}{\tt Offl} & \multicolumn{1}{c|}{\tt Onl}   & {\tt Succ} \\ \hline
{\tt Friction}      & \multicolumn{1}{c|}{2.5M }        & \multicolumn{1}{c|}{58.15}      &  \textbf{100\%}       & \multicolumn{1}{c|}{0.81M}       &     {35\%}    & \multicolumn{1}{c|}{1.48M}       &    {13\%}    & \multicolumn{1}{c|}{295.225}       &     {37.5\%}     &   \multicolumn{1}{c|}{1M} & \multicolumn{1}{c|}{609.225}  &    {60\%}   \\ \hline
{\tt Maze}          & \multicolumn{1}{c|}{2.05M}        & \multicolumn{1}{c|}  {241.62}     &  \textbf{100\%}       & \multicolumn{1}{c|}{0.54M}           &     {34\%}    & \multicolumn{1}{c|}{1.74M}       &     {31\%}  & \multicolumn{1}{c|}{623.57}       &     {58\%}     &  \multicolumn{1}{c|}{1M} & \multicolumn{1}{c|}{693.73}   &    {87\%}    \\ \hline
{\tt Quadrotor}     & \multicolumn{1}{c|}{1M}        & \multicolumn{1}{c|} {736.3}    &  \textbf{100\%}       & \multicolumn{1}{c|}{1.3M}           &    {5\%}     & \multicolumn{1}{c|}{1.01M}       &    {13\%}   & \multicolumn{1}{c|}{14k}       &     {0\%}      &  \multicolumn{1}{c|}{100k}  & \multicolumn{1}{c|}{13.7k} &    {0\%}    \\ \hline
\end{tabular}
  \vspace{-.05in}
  \caption{\small Comparing DRL approaches against SBMPs in terms of computation costs ($\#$ of calls made to the physics engine) and success rate on the physically-simulated benchmarks.}
\label{tab:drl}
\vspace{-.25in}
\end{table*}


\subsection{Ablation studies}  

\begin{figure}[ht!]
    \centering
    \begin{subfigure}[b]{0.49\textwidth}
    \includegraphics[width=0.49\linewidth]{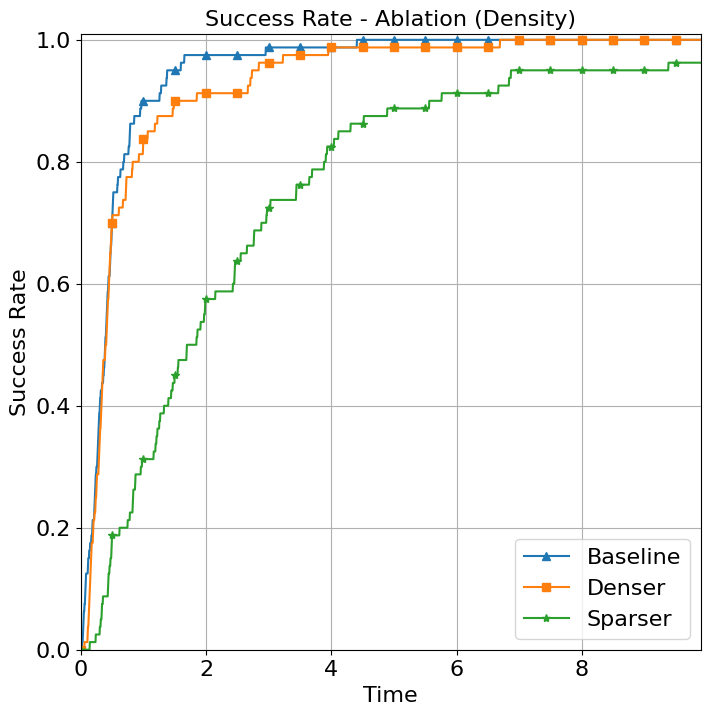}
    \includegraphics[width=0.49\textwidth]{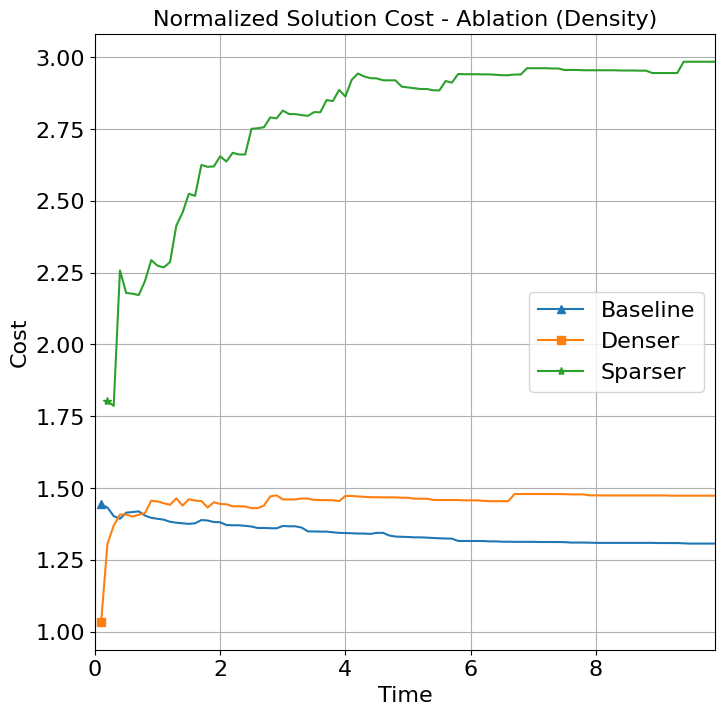}
    \end{subfigure}
    \vspace{-.1in}
    \caption{\small Ablation experiments for various roadmap sizes.}
    \vspace{-.1in}
    \label{fig:ablation-size}
\end{figure}

Fig.~\ref{fig:ablation-size} considers 3 roadmaps of different sizes by varying the number of configurations $N$ during roadmap construction: \textbf{\tt Baseline} ($\vert \mathcal{V} \vert = 1088, \vert \mathcal{E} \vert = 10911$), \textbf{\tt Denser} ($\vert \mathcal{V} \vert = 1360, \vert \mathcal{E} \vert = 156936$), and \textbf{\tt Sparser} ($\vert \mathcal{V} \vert = 688, \vert \mathcal{E} \vert = 4718$). Planners using the \textbf{\tt Baseline} and \textbf{\tt Denser} roadmaps find competitive solutions quickly, while the \textbf{\tt Sparser} roadmap cannot do so, motivating the use of large roadmaps for \textbf{\tt RoGuE}.

\begin{figure}[ht!]
    \centering
    \begin{subfigure}[b]{0.49\textwidth}
    \includegraphics[width=0.49\textwidth]{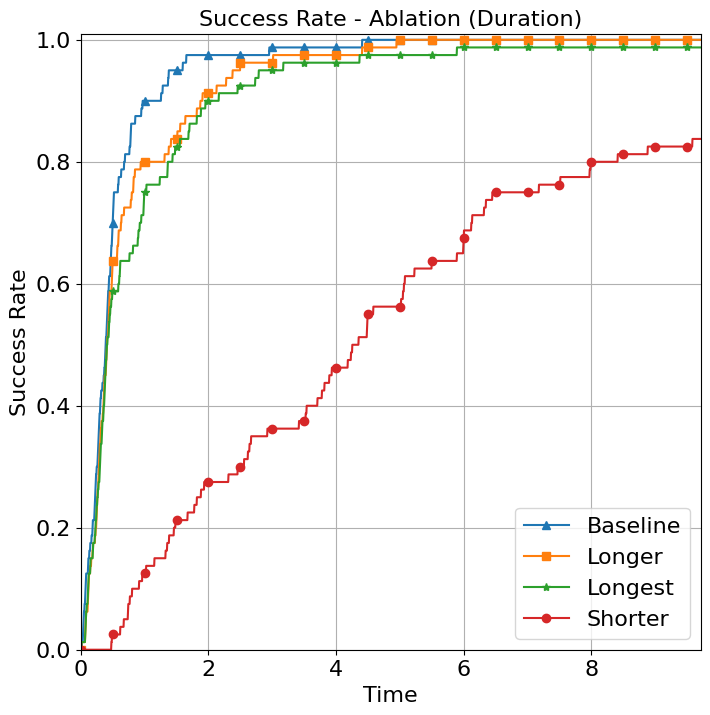}
    \includegraphics[width=0.49\textwidth]{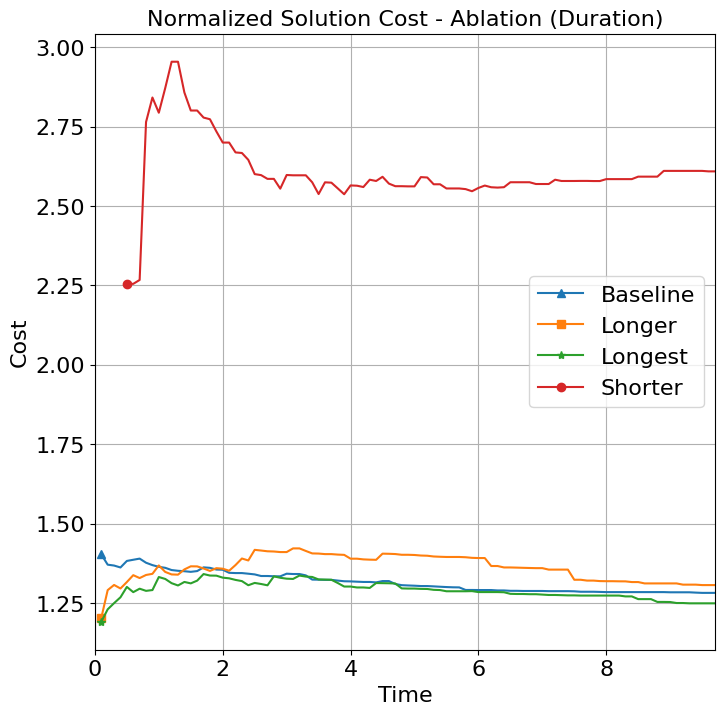}
    \end{subfigure}
    \vspace{-.1in}
     \caption{\small Ablation experiments varying the max. edge duration.}
     \vspace{-.2in}
    \label{fig:ablation-duration}
\end{figure}

Fig.~\ref{fig:ablation-duration} studies the effect of the maximum allowed edge cost $T_\text{max}$ on the online planner. Four different values of $T_\text{max}$ are considered: \textbf{{\tt Baseline}} (10s), \textbf{{\tt Shorter}} (5s), \textbf{{\tt Longer}} (20s), and \textbf{{\tt Longest}} (30s). The planner using the \textbf{\tt Shorter} roadmap underperforms, while the planners using the other values find solutions quickly. As $T_\text{max}$ increases, the time taken by the planner to find solutions to all problems increases slightly without significantly affecting the cost of the returned solution. This suggests that while longer edges may help find high-quality solutions, they may also suffer due to the gaps in the roadmap.

    \section{Conclusion}
\label{sec:conclusion}

This paper proposes a strategy that can benefit from learned controllers to improve the efficiency of kinodynamic planning for robots with significant dynamics. It utilizes a controller trained offline in an empty environment. The target environment is represented via a ``Roadmap with Gaps'' over local regions and applications of the controller between them. Given a wavefront over the roadmap for a specific goal, a tree sampling-based motion planner generates informed subgoals and uses the controller to reach them. When the controller cannot reach a subgoal, the planner resorts to random exploration. Evaluation shows the significant improvement in  planning efficiency.

For higher-dimensional systems, the memory requirements of the roadmap can be improved by considering sparse representations \cite{dobson2014sparse}. Furthermore, learned reachability estimators can assist in efficient roadmap construction and online queries. This work assumes an accurate model of the environment and the robot, which complicates deployment on real systems. This motivates integrating the proposed motion planner with system identification, state estimation and feedback control to track the planned trajectory.


}{
    
    \section{Preliminaries}
\label{sec:prelims}

Consider a system with state space $\mathbb{X}$ and control space $\mathbb{U}$. $\mathbb{X}$ is divided into  collision-free ($\mathbb{X}_f$) and obstacle ($\mathbb{X}_{\mathrm{o}}$) subsets. The dynamics $\dot{x} = f(x,u)$ (where $x \in \mathbb{X}_f, u \in \mathbb{U}$) govern the robot's motions.  The process $f$ can be an analytical ordinary differential equation (ODE) or modeled via a physics engine, e.g., MuJoCo \cite{todorov2012mujoco}.  A function $\mathbb{M}: \mathbb{X} \rightarrow \mathbb{Q}$ maps a state $x \in \mathbb{X}$ to its corresponding \textit{configuration space} point $q \in \mathbb{Q}$ ($q = \mathbb{M}(x)$). The inverse mapping $\mathbb{M}^{-1}(q_i)$ returns a state $x_i \in \mathbb{X}$ by setting the dynamics term to some nominal value (e.g., by setting all velocity terms to 0). A distance function $d(\cdot,\cdot)$ is defined over $\mathbb{Q}$, which corresponds to a weighted Euclidean distance metric in SE(2).  


A \textit{plan} $p_T$ is a sequence of piecewise-constant controls of duration $T$ that induce a trajectory $\tau \in \mathcal{T}$, where $\tau: [0,T] \mapsto \mathbb{X}_f$. Given a start state $x_0 \in \mathbb{X}_f$ and a goal set $X_G \subset \mathbb{X}_f$, a feasible motion planning problem admits a solution trajectory of the form $\tau(0) = x_0, \tau(T) \in X_G$.   The goal set is defined as $X_G = \{x \in \mathbb{X}_f \ \vert \ d(\mathbb{M}(x), q_G) < \epsilon \}$, or equivalently, $\mathcal{B}(q_G, \epsilon)$ where $\epsilon$ is a tolerance parameter.
A heuristic $h: \mathbb{X} \rightarrow \mathbb{R}^+$ estimates the \textit{cost-to-go} of an input state $x$ to the goal region $X_G$.
Each solution trajectory has a cost given by $\texttt{cost}: \mathcal{T} \rightarrow \mathbb{R}^+$. An optimal motion planning problem aims to minimize the cost of the solution trajectory.

{\bf Sampling-Based Planning Framework:} The framework uses the forward propagation model $f$ to explore $\mathbb{X}_f$ by incrementally constructing a tree via sampling in $\mathbb{U}$. Algo.~\ref{alg:tree-sbmp} outlines the high-level operation of a sampling-based motion planner (Tree-SBMP) that builds a tree of states rooted at the initial state $x_0$ until it reaches $X_G$. 

\vspace{-.15in}
\begin{algorithm}
        \SetAlgoLined
        $\mathrm{T} \leftarrow \{x_0\}$; \\
        \While{termination condition is not met}
        {
        $x_\text{sel} \leftarrow \texttt{SELECT-NODE}(\mathrm{T})$; \\ 
        $u \leftarrow \texttt{EXPAND}(x_\text{sel})$; \\
        $x_\text{new} \leftarrow \texttt{PROPAGATE}(x_\text{sel},u)$; \\
        \If{$(x_\text{sel} \rightarrow x_\text{new}) \in \mathbb{X}_f$}
        {
        \texttt{EXTEND-TREE}($\mathrm{T}, x_\text{sel} \rightarrow x_\text{new}$);
        }
        }
        \caption{\small Tree-SBMP($\mathbb{X},\mathbb{U},x_0,X_G$)}
        \label{alg:tree-sbmp}
\end{algorithm}
\vspace{-.2in}


Each iteration of Tree-SBMP selects an existing tree node/state $x_\text{sel}$ to expand (Line 3). Then, it generates a control sequence $u$ and propagates it from $x_\text{sel}$ to obtain a new state $x_\text{new}$ (Lines 4-5). The resulting edge is added to the tree if not in collision (Lines 6-7). Upon termination (e.g., a time threshold), if the tree has states in $X_G$, the best-found solution according to $cost$ is returned.  By varying how the key operations in Algo. ~\ref{alg:tree-sbmp} are implemented, different variations can be obtained. The specific variant this work adopts is the informed and AO ``Dominance-Informed Region Tree'' (DIRT) \cite{LB-DIRT}. It uses an admissible state heuristic function $h$ to select nodes in an informed manner. If $x_\text{new}$ improves upon $x_\text{sel}$ given $h$, then it is selected as the next $x_\text{sel}$. DIRT also propagates a ``blossom'' of $k$ controls at every iteration and prioritizes the propagation of the edge that brings the robot closer to the goal given the heuristic.

\section{Proposed Method}
\label{sec:proposed}

The proposed method has 2 offline stages: (i) training a controller $\pi(x,q)$ in an obstacle-free environment, and (ii) building a ``Roadmap with Gaps" $(\mathcal{V,E})$ in the target environment given the controller. Online, given a motion planning query $(x_0,x_G)$ in the target environment, a sampling-based kinodynamic planner expands a tree of feasible trajectories similar to Algo. ~\ref{alg:tree-sbmp} above. Line 4 of Algo. ~\ref{alg:tree-sbmp} is adapted so as to use guidance from a wavefront function computed over the ``Roadmap with Gaps".
The method computes first a local goal $q_\text{lg}$ that looks promising given the wavefront. It then propagates a control $u = \pi(x_\text{sel},q_\text{lg})$ by using the trained controller $\pi$. The controller is applied at the selected node $x_\text{sel} \in \mathbb{X}$ and generates a control towards reaching the local goal $q_\text{lg}$ without considering obstacles. 


\subsection{Training a Controller offline}

A goal-conditioned controller $u = \pi(x,q)$ is first trained via Reinforcement Learning to reach from an initial state $x$ to a goal set $\mathcal{B}(q,\epsilon)$, i.e., within an $\epsilon$ distance of $q$. For the training, this is attempted for any $(x,q) \in \mathbb{X} \times \mathbb{Q}$ in an empty environment of given dimensions. The training process collects transitions $(x_t, u_t, \mathcal{C}(x_t,q), x_{t+1}, q)$, which are stored in a replay buffer. The cost function $\mathcal{C}: \mathbb{X} \times \mathbb{Q} \rightarrow \{0,-1\}$ has a value of $\mathcal{C}(x_t,q) = 0$ iff  $x_t \in \mathcal{B}(q,\epsilon)$, and $-1$ otherwise. During each iteration, mini-batches of transitions are sampled from the buffer. A Soft Actor-Critic (SAC) \cite{Haarnoja2018SoftAO} algorithm is employed, which minimizes the total cost $\mathbb{E}_{x_0, q_G \sim \mathbb{X} \times \mathbb{Q}} [\sum_{t=0}^T \mathcal{C}(x_t,q_G)]$. Concurrently, Hindsight Experience Replay (HER) \cite{Andrychowicz2017HindsightER} relabels some transitions with alternative goals to provide additional training signals from past experiences.


\subsection{Building a Roadmap with Gaps offline}
\label{subsec: roadmap}

The proposed approach then builds a ``Roadmap with Gaps"  $(\mathcal{V,E})$ at the target environment, i.e., a graphical representation where nodes $\mathcal{V}$ correspond to configurations $q_i$ of the vehicle. Edges $(q_i,q_j) \in \mathcal{E}$ exist between vertices as long as the application (of maximum duration $T_\text{max}$) of the available controller $\pi(\mathbb{M}^{-1}(q_i), q_j)$ at a zero velocity state $\mathbb{M}^{-1}(q_i)$ of the initial configuration $q_i$ brings the system within a hyperball $\mathcal{B}(q_j,\epsilon)$ of the configuration $q_j$.

The roadmap construction procedure (Fig~\ref{fig:main-a}) samples first configurations $\{q_i\}_{i=1}^{\vert N \vert}$, which form the roadmap vertices  $\mathcal{V}$. Here, the vertices are sampled over a grid in $\mathbb{Q}$. They are collision-checked and verified to be in $\mathbb{Q}_{\mathrm{f}}$. Then, for the edges $\mathcal{E}$, the procedure selects a random $q_i \in \mathcal{V}$ and obtains two sets: $A(q_i)$ and $D(q_i)$. $A(q_i)$ is the set of vertices $q \in \mathcal{V},$ whose hyperballs $\mathcal{B}(q,\epsilon)$ can be reached from $x_i = \mathbb{M}^{-1}(q_i)$ given $\pi$. Similarly, $D(q_i)$ is the set of vertices $q \in \mathcal{V}$, where the state $x = \mathbb{M}^{-1}(q)$ can reach the hyperball $\mathcal{B}(q_i,\epsilon)$ given $\pi$. Then, edges from $q_i$ to vertices in $A(q_i)$ and edges from vertices in $D(q_i)$ to $q_i$ are added to the set $\mathcal{E}$. 

\begin{wrapfigure}{r}{0.44\columnwidth}
     \centering
     \vspace{-.15in}
     \includegraphics[width=0.43\columnwidth]{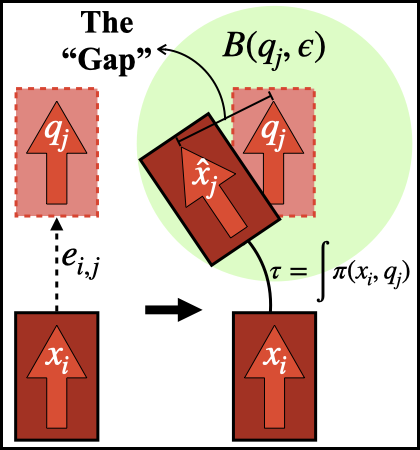}
     \vspace{-0.3in}
     \label{fig:enter-label}
\end{wrapfigure}

The application of the controller for the edge construction, especially a learned one for a non-linear dynamical system, implies that for all edges $(q_i,q_j)$ in the roadmap, there is no guarantee that the resulting configuration $q_j$ can be achieved exactly from $q_i$. Instead, the corresponding edge only guarantees that the system will be within a hyperball $\mathcal{B}(q_G,\epsilon)$ if it is initialized at $\mathbb{M}^{-1}(q_i)$. Consequently, the edges of the roadmap introduce {\bf ``gaps"} as highlighted by the figure. This issue is exacerbated if the controller is used to follow a sequence of edges on the roadmap, e.g., $(q_i,q_j)$ and $(q_j,q_k)$. Since the controller's application over the first edge can only bring the system in the vicinity of $q_j$,  the application of the controller corresponding to the second edge at the resulting configuration may not even bring the system within an $\epsilon$-hyperball of $q_k$. Consequently, paths on the ``Roadmap with Gaps" do not respect the dynamic constraints of the vehicle and cannot be used to directly solve kinodynamic planning problems. They can still provide, however, useful guidance for a kinodynamic planner.  

\begin{figure}[t]
    \centering
    \begin{subfigure}{.34\linewidth}
    \includegraphics[width=\textwidth]{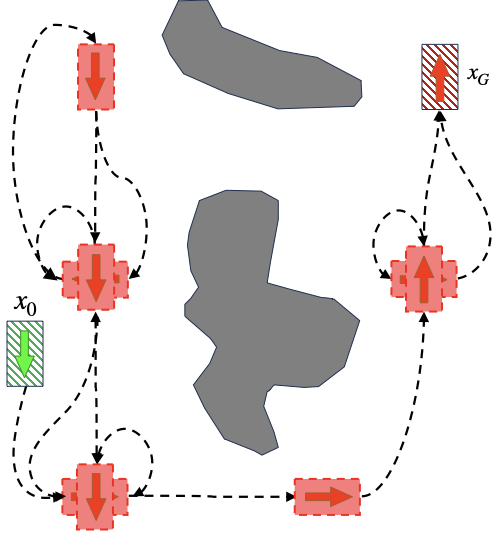}
    \vspace{-.25in}
    \caption{}
    \label{fig:main-a}
    \end{subfigure}
    \begin{subfigure}{.64\linewidth}
    \includegraphics[width=\textwidth]{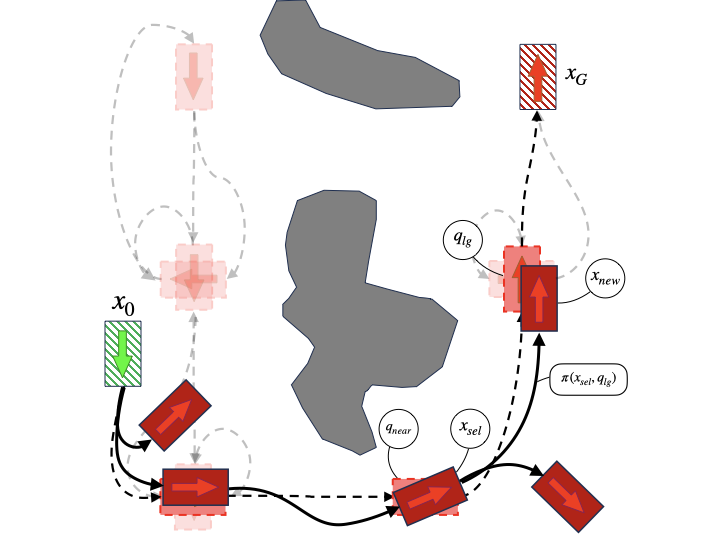}
    \vspace{-.25in}
    \caption{}
    \label{fig:main-b}
    \end{subfigure}
    \vspace{-.15in}
    \caption{\small (a) A ``Roadmap with Gaps" consists of vertices (configurations shown as dotted boxes). The roadmap's directional edges (dotted lines) correspond to where the controller executed from the source successfully reached the vicinity of the target. For a new query $(x_0, x_G)$, the start $q_0$ and goal $q_G$ are added to the roadmap (transparent). (b)  During the expansion of the planning tree (opaque), {\tt RoGuE} selects informed local goals $q_\text{lg}$ for the controller $\pi$. The closest roadmap node $q_\text{near}$ to the selected tree node $x_\text{sel}$ is identified. Its roadmap successor is passed as the local goal $q_\text{lg}$ to the controller. This expansion adds a new tree node $x_\text{new}$.}
    \vspace{-.2in}
    \label{fig:main-figure}
\end{figure}

\subsection{\textbf{\tt RoGuE}: Guiding expansion via a roadmap with gaps}
\label{subsec:rogue}

Algo.~\ref{alg:rogue-tree} describes the \textbf{Ro}admap-\textbf{Gu}ided \textbf{E}xpansion (\textbf{\tt RoGuE})-Tree method, which uses the ``Roadmap with Gaps" to expand online an AO Tree Sampling-based Motion Planner, given a new query $(x_0,x_G)$. It first adds vertices $q_0 = \mathbb{M}(x_0)$ and $q_G = \mathbb{M}(x_G)$ to the roadmap and connects them appropriately, i.e., it adds edges from $q_0$ to the set $A(q_0)$ and edges from the set $D(q_G)$ to $q_G$.


Then, it computes a \textit{wavefront} $\mathcal{W}: \mathcal{V} \mapsto \mathbb{R}^+$ over the roadmap. Initially, all vertices have an infinite  value but the goal, where $\mathcal{W}(q_G) = 0$. Then, a backward search computes the value of every node according to $\mathcal{W}(v) = min_{\mathcal{V}} \{ c(v,v') + \mathcal{W}(v') \ \vert \  \in \mathcal{E}\}$, where $c(v,v')$ is the cost of edge $(v,v')$, i.e., the duration required to reach the vicinity of $v'$ from $v$ given $\pi$. This also allows to compute the ${\tt Successor}(v)$ of a node, which is the out-neighbor that defines the wavefront value of $v$. For a node with an infinite value at the end of the process, its successor is undefined.

\vspace{-.15in}
\begin{algorithm}
        \SetAlgoLined
        Add $q_0$, $q_G$  to roadmap $(\mathcal{V,E})$ as start and goal;\\
        $\mathcal{W} \leftarrow \texttt{GET-WAVEFRONT}((\mathcal{V,E}))$; \\
        $\mathrm{T} \leftarrow \{x_0\}$; \\
        \While{termination condition is not met}
        {
        $x_\text{sel} \leftarrow \texttt{SELECT-NODE}(\mathrm{T})$; \\ 
        $u \leftarrow \texttt{RoGuE}(x_\text{sel}, \mathcal{V,W}, \pi)$; \\
        $x_\text{new} \leftarrow \texttt{PROPAGATE}(x_\text{sel},u)$; \\
        \If{$(x_\text{sel} \rightarrow x_\text{new}) \in \mathbb{X}_{\mathrm{f}}$}
        {
        \texttt{EXTEND-TREE}($\mathrm{T}, x_\text{sel} \rightarrow x_\text{new}$);
        }
        }
        \caption{\small {\tt RoGuE-Tree} ($\mathbb{X},\mathbb{U},x_0,x_G,\pi, (\mathcal{V,E}))$}
        \label{alg:rogue-tree}
\end{algorithm}
\vspace{-.2in}

Then, a tree $T$ data structure is expanded by selecting a tree node $x_{sel}$, propagating a control $u$ from $x_{sel}$ and generating a new tree node $x_{new}$. For the selection, and following the DIRT planner \cite{LB-DIRT}, when the cost-to-go value of the roadmap node closer to $x_{new}$ is lower than that of its parent's, then $x_{new}$ is selected for expansion at the consecutive step so the tree can make progress along a promising path. 

The {\tt RoGuE} function selects the control $u$ and is detailed in Algo.~\ref{alg:greedy-expand} and Fig~\ref{fig:main-b}). It uses the wavefront $\mathcal{W}$ to provide an informed local goal. The first time a tree node $x_\text{sel}$ is selected, {\tt RoGuE} identifies the closest roadmap node $q_\text{near} \in \mathcal{V}$ according to a distance function $d(\cdot,\cdot$) (Line 2). It then queries the {\tt Successor}($q_\text{near}$). If defined, it is used as the local goal $q_\text{lg}$ for the controller $\pi$ (Lines 3-5). If the successor is undefined, a random local goal in $\mathbb{Q}$ is chosen (Line 6-7). 


\vspace{-.15in}
\begin{algorithm}
\SetAlgoLined
\If{first time $x_\text{sel}$ is expanded}
{
$q_\text{near} \leftarrow {\tt ClosestRoadmapNode}(x_\text{sel}, \mathcal{V})$; \\
$q_\text{lg} \leftarrow {\tt Successor}(q_\text{near}, \mathcal{W});$ \\
\If{$q_\text{lg}$ is defined} {
    $u \leftarrow \pi(x_\text{sel},q_\text{lg});$
}
\Else
{
    $u \leftarrow \pi(x_\text{sel},\mathbb{Q}.\text{sample}());$
}
}
\Else
{
$u \leftarrow \mathbb{U}.\text{random-sample}();$
}
\textbf{return } $u$;
\caption{\tt{RoGuE}($x_\text{sel},\mathcal{V,W}, \pi$)}
\label{alg:greedy-expand}
\end{algorithm}
\vspace{-.2in}

{\bf Maintaining Asymptotic Optimality:} When tree nodes have a non-zero probability of being selected, and the probability of expansion along the lowest-cost trajectory is non-zero, the planner is AO.  {\tt RoGuE} adopts the node selection process of DIRT \cite{LB-DIRT}, which, while informed, maintains a positive probability for all tree nodes. Similarly, while {\tt RoGuE} initially applies informed expansions from each node, subsequent expansions apply random controls from the set $\mathbb{U}$ (Algo~\ref{alg:greedy-expand}, Line 8-9) to retain AO properties \cite{LB-DIRT}. 


    \section{Experimental Evaluation}
\label{sec:experiments}

The {\bf robot systems} considered in the evaluation are: (i) an analytically simulated second-order differential-drive, (ii) an analytically simulated car-like vehicle (where $\texttt{dim}(\mathbb{X}) = 5, \texttt{dim}(\mathbb{U}) = 2$), (iii) a MuSHR car \cite{srinivasa2019mushr} physically simulated on MuJoCo \cite{todorov2012mujoco} ($\texttt{dim}(\mathbb{X}) = 27, \texttt{dim}(\mathbb{U}) = 2$), and (iv) a Skydio X2 Autonomous Drone ($\texttt{dim}(\mathbb{X}) = 13, \texttt{dim}(\mathbb{U}) = 4$). For the drone, the distance function $d(\cdot,\cdot)$ considers its center of mass 3D position. For all systems, the parameter $\epsilon$ is set to the same value $0.5$. All planning experiments are implemented using the {\tt ML4KP} software library \cite{ML4KP} and executed on a cluster with Intel(R) Xeon(R) Gold 5220 CPU @ 2.20GHz and 512 GB of RAM.

\begin{figure}[ht!]
    \vspace{-.1in}
    \centering

    \includegraphics[width=0.32\linewidth]{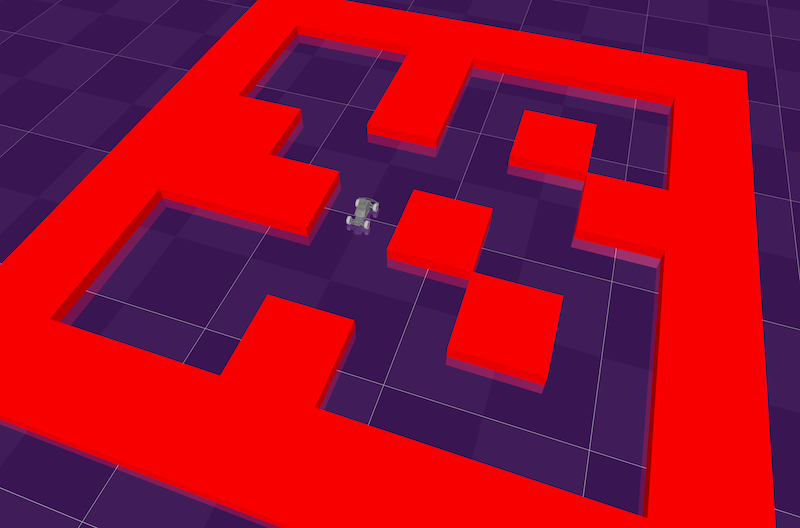}
    \includegraphics[width=0.32\linewidth]{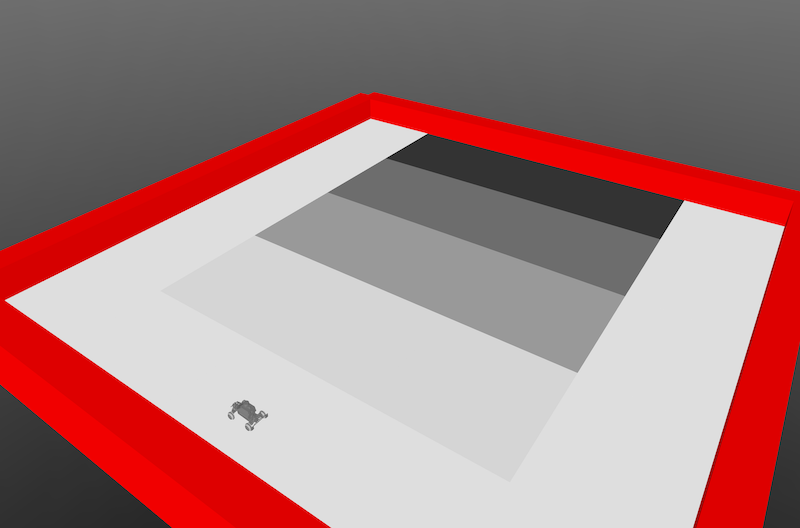}
    \includegraphics[width=0.32\linewidth]{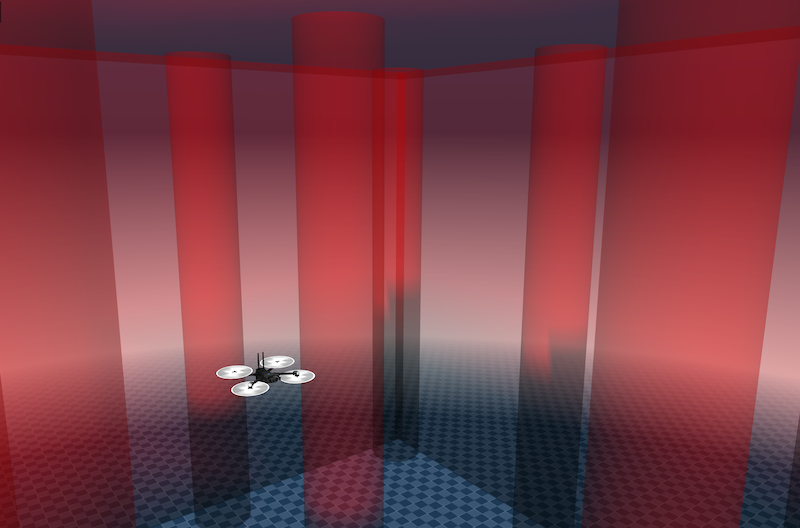}
    \vspace{-.05in}
    \caption{\small Physically simulated benchmarks using MuJoCo. (L-R) {\tt Maze}, {\tt Friction}, {\tt Quadrotor.}}
    \vspace{-.2in}
    \label{fig:mujoco-benchmarks}
\end{figure}


Across experiments, the following {\bf metrics} are measured: (1) Average normalized cost of solutions found over computation time; and (2) Ratio of problems solved over time. For all problems, the path cost is the solution plan's duration. To better reflect the performance of different methods and account for the difficulty of different problems, path costs reported are normalized by dividing by the best path cost ever found for a problem across all planners. 

Two {\bf comparison expansion functions} are considered in the evaluation: (a) \textbf{{\tt Random}} uses a blossom expansion of random controls in $\mathbb{U}$, and (b) \textbf{{\tt RLG}} samples Random Local Goals as input to $\pi$ and outputs the controls returned. In terms of {\bf comparison motion planners}, both an uninformed Rapidly-exploring Randomized Tree (RRT) \cite{lavalle2001randomized} and the informed but AO DIRT are considered. For the \textbf{\tt Random} and \textbf{\tt RLG} expansion strategies of DIRT, a blossom of 5 controls is implemented. 

Additional AO planners were considered for experimentation but it was difficult to achieve useful results with them. The \textit{Bundle of Edges (BoE)} \cite{shome2021asymptotically} approach failed to find solutions while using a similarly-sized roadmap. The \textit{discontinuity-bounded A*} (dbA*) \cite{honig2022db} relies on motion primitives, which are not available, however, for the second-order  robots considered here. 

All (start, goal) pairs in the experiments are included as milestones during the roadmap construction to reduce connection time during online planning. These connections also lend themselves to multi-threading or alternatively the closest roadmap nodes can be used as surrogates for the actual start and goal states. 

\begin{figure*}[ht!]
\centering
     \includegraphics[width=.235\textwidth]{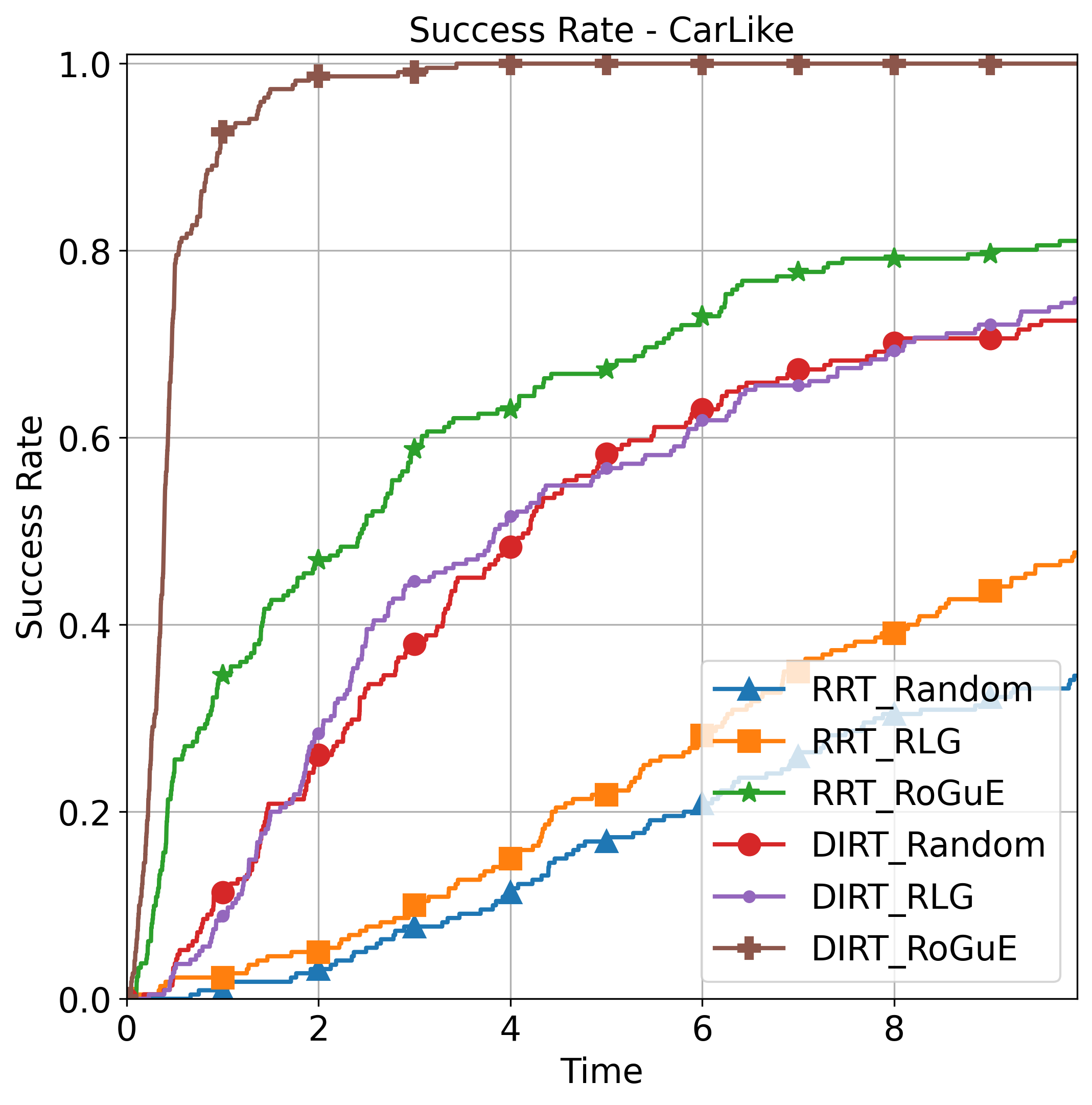}
     \includegraphics[width=.235\textwidth]{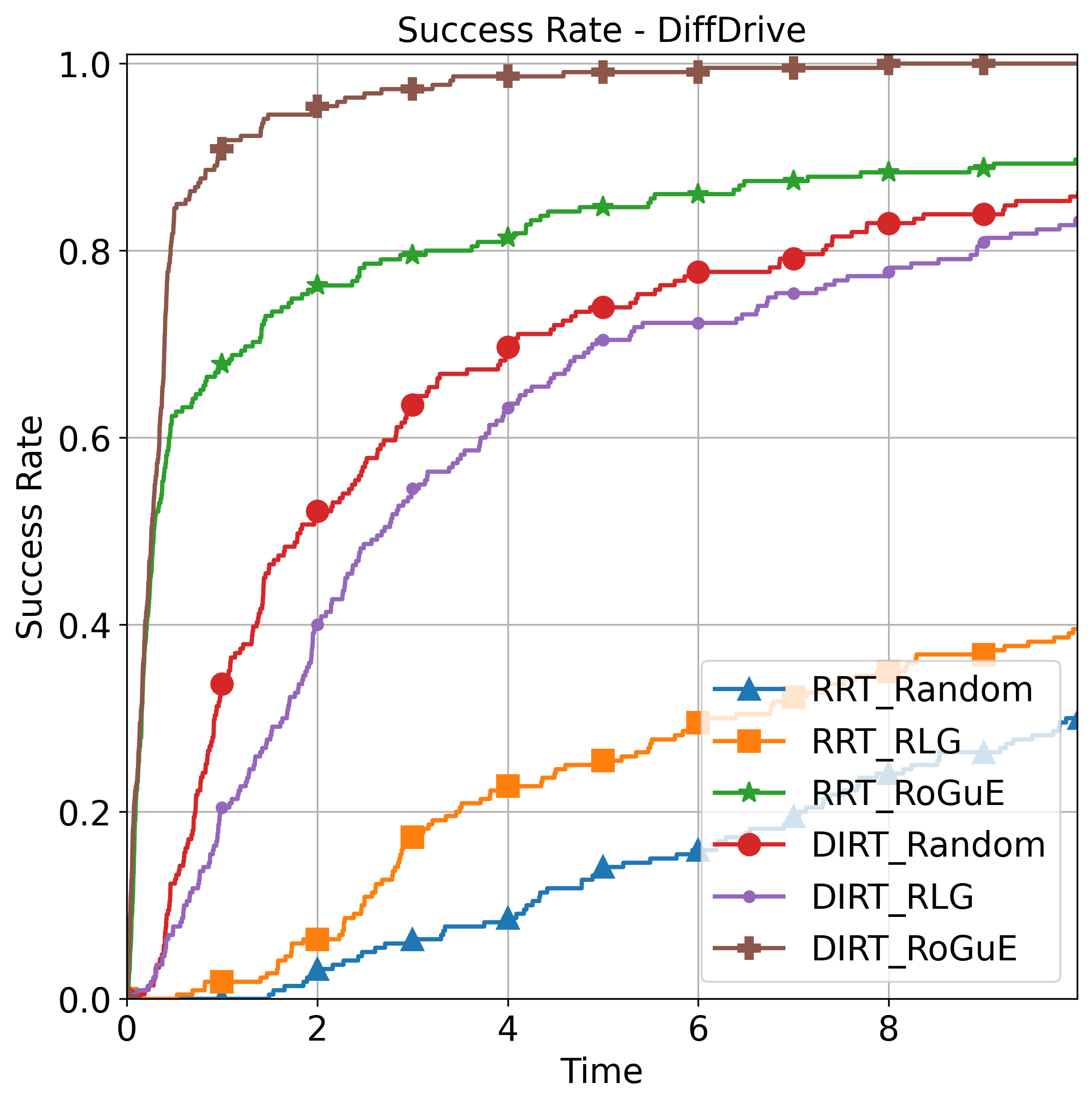}
     \rulesep
     \includegraphics[width=.235\textwidth]{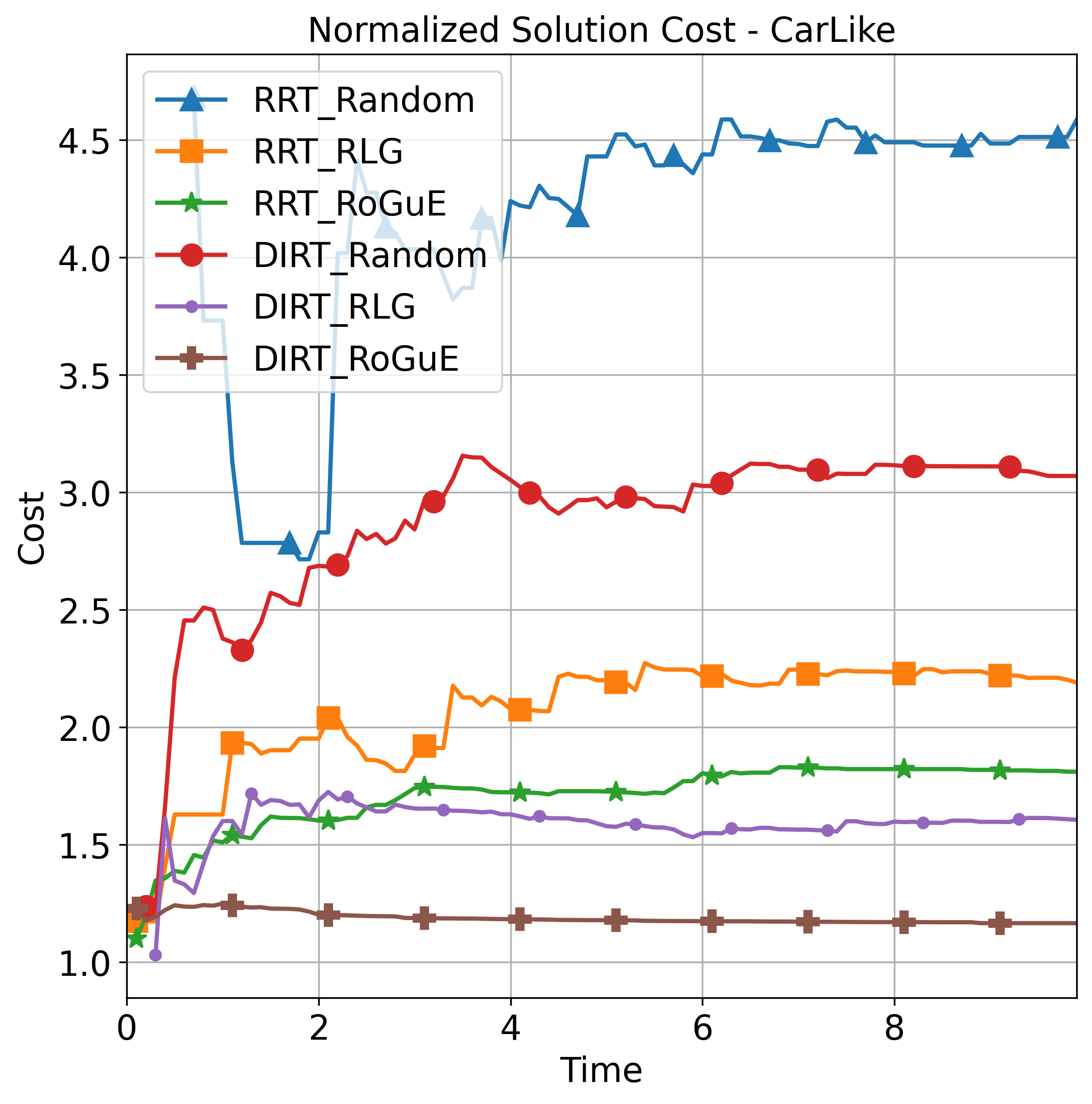}
     \includegraphics[width=.235\textwidth]{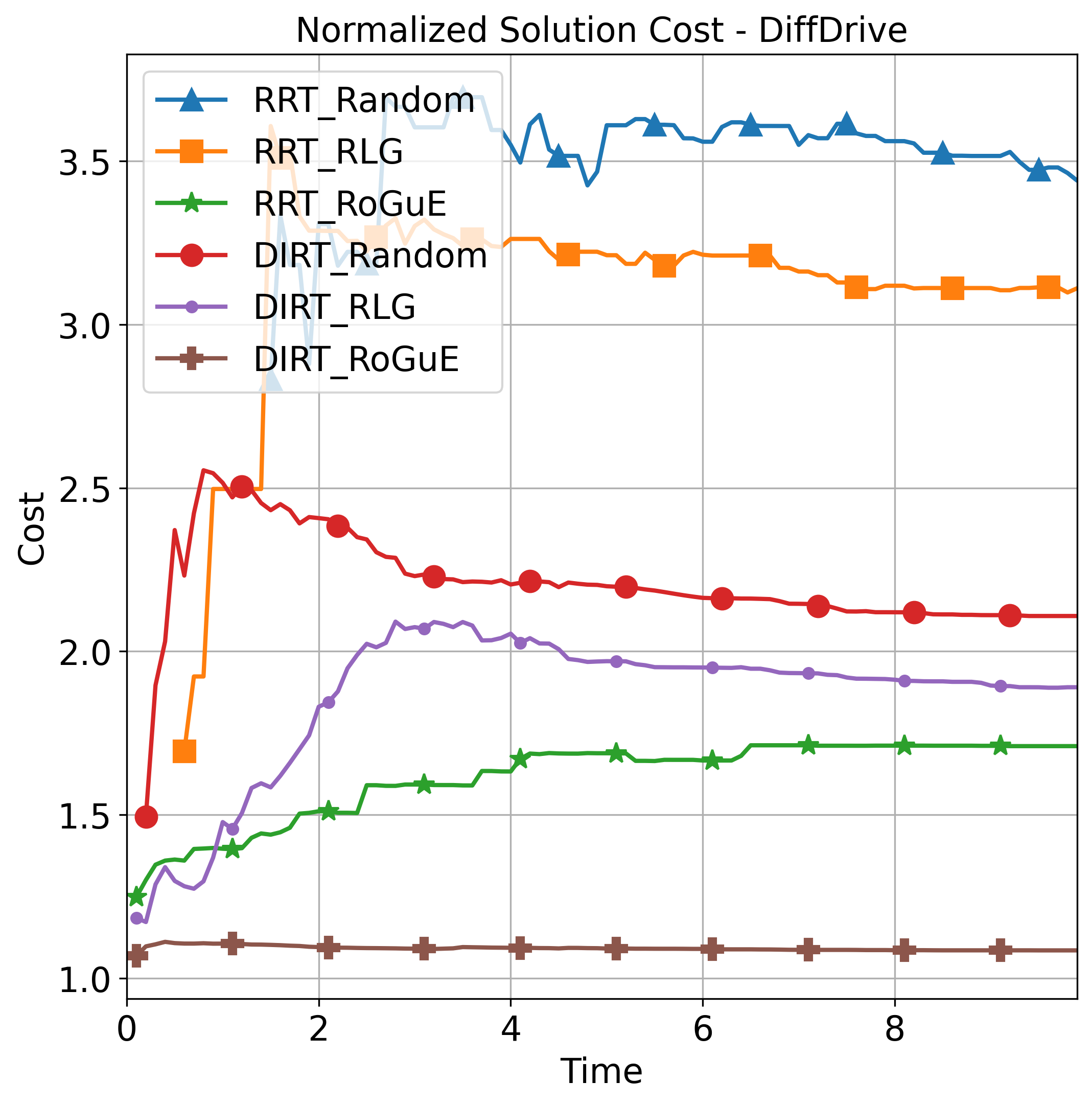}
\caption{\small Analytical benchmarks: Left \& middle-left: higher success rate earlier is better. Middle-right \& Right: lower path cost is better. As more solutions are discovered, the average cost may increase. Each problem instance is run 10 times to account for different seeds.}
\label{fig:analytical-results}
\vspace{-.1in}
\end{figure*}

\begin{figure*}[ht!]
\centering
\begin{subfigure}[b]{\textwidth}
         \includegraphics[width=.235\textwidth]{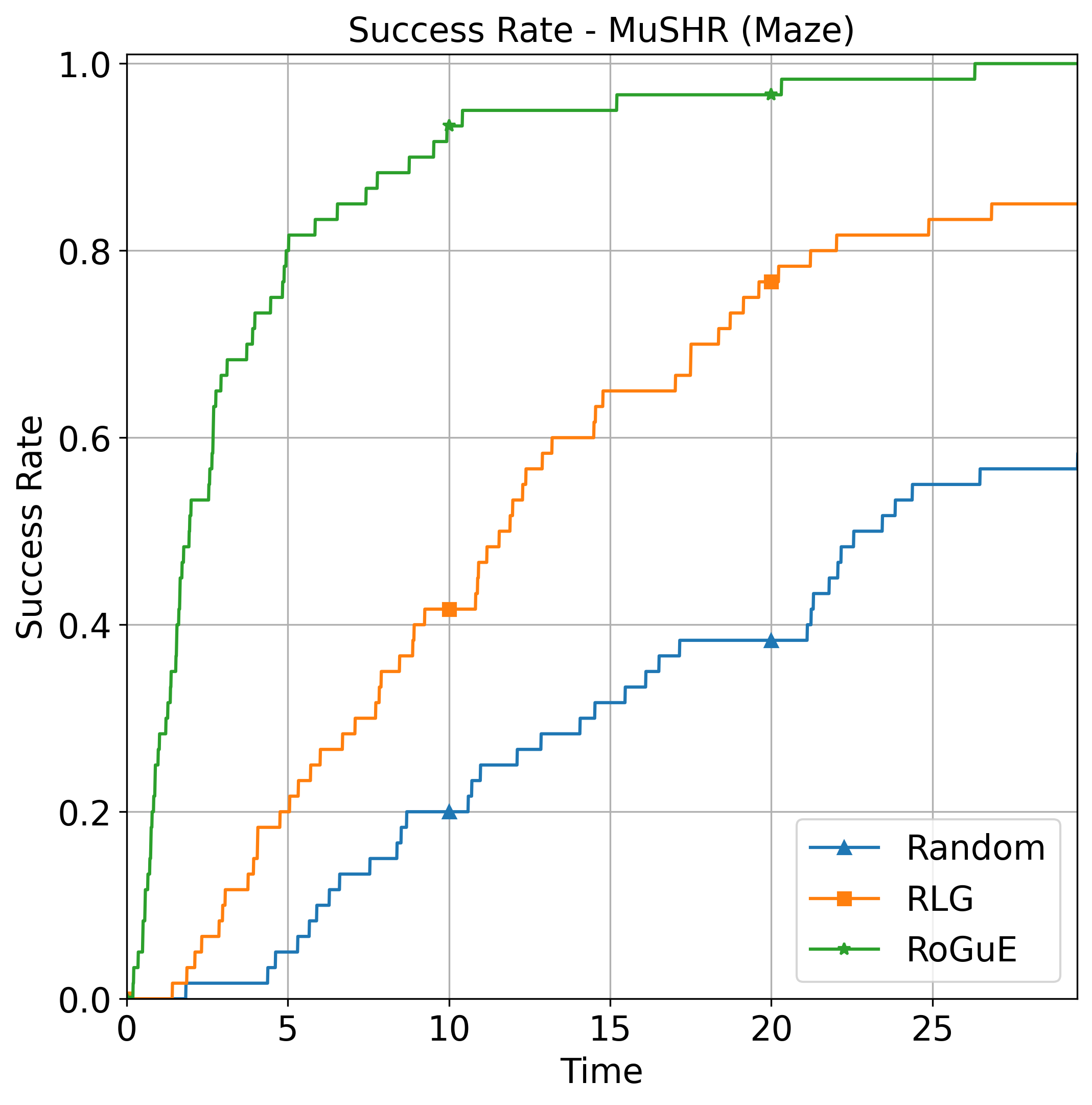}
         \hspace{.01\textwidth}
         \includegraphics[width=.235\textwidth]{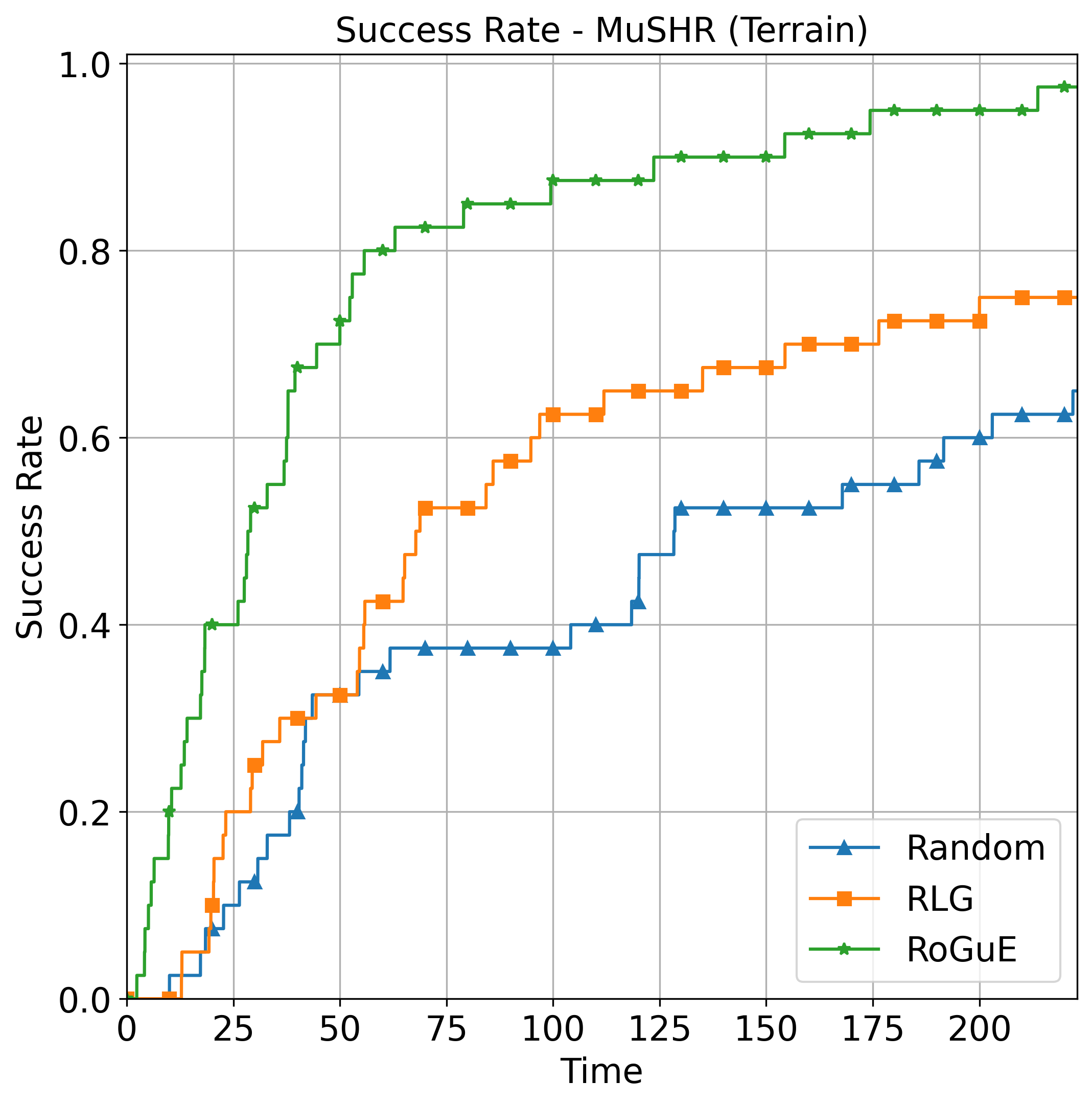}
         \hspace{.01\textwidth}
         \includegraphics[width=.235\textwidth]{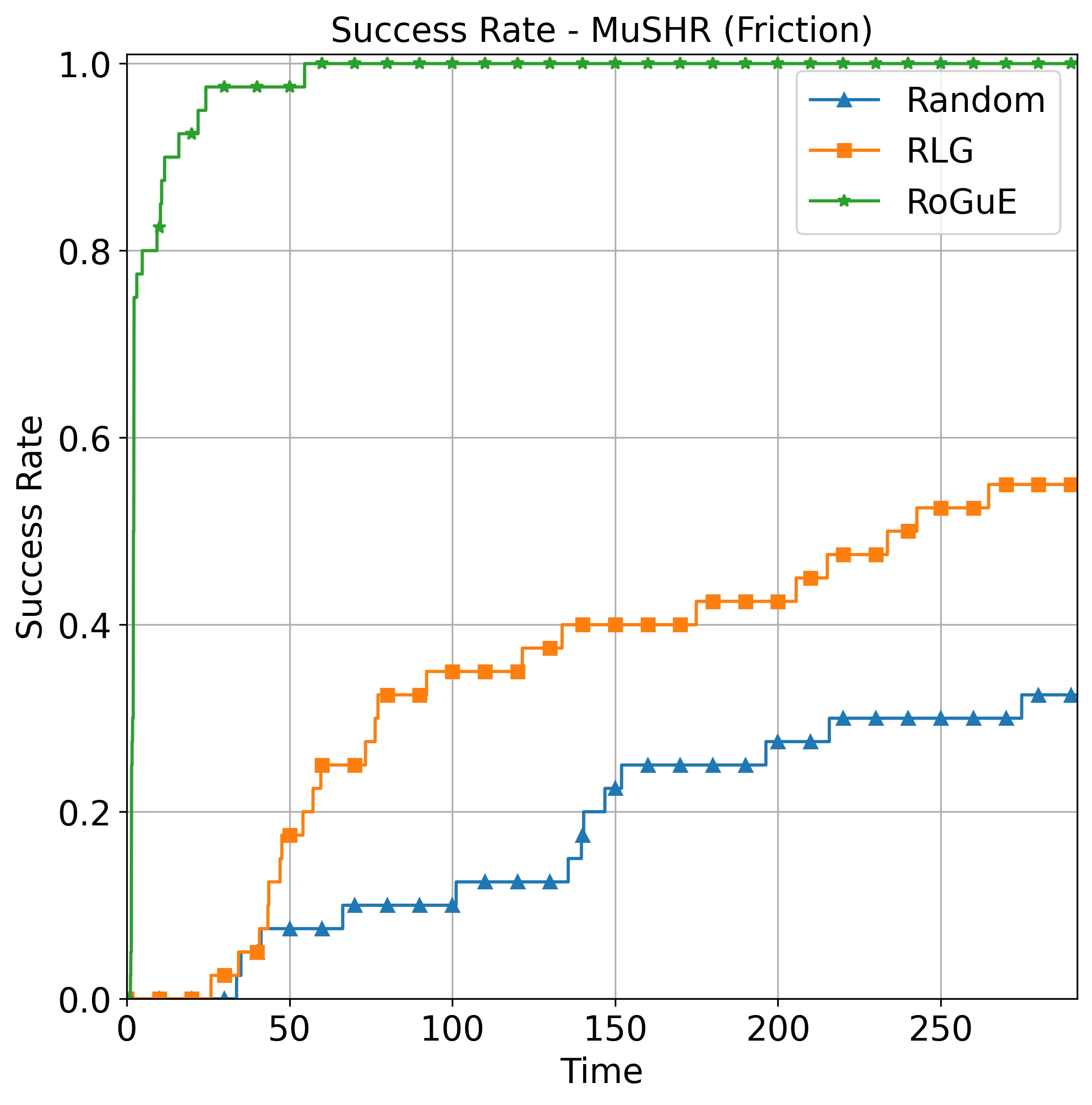}
         \hspace{.01\textwidth}
         \includegraphics[width=.235\textwidth]{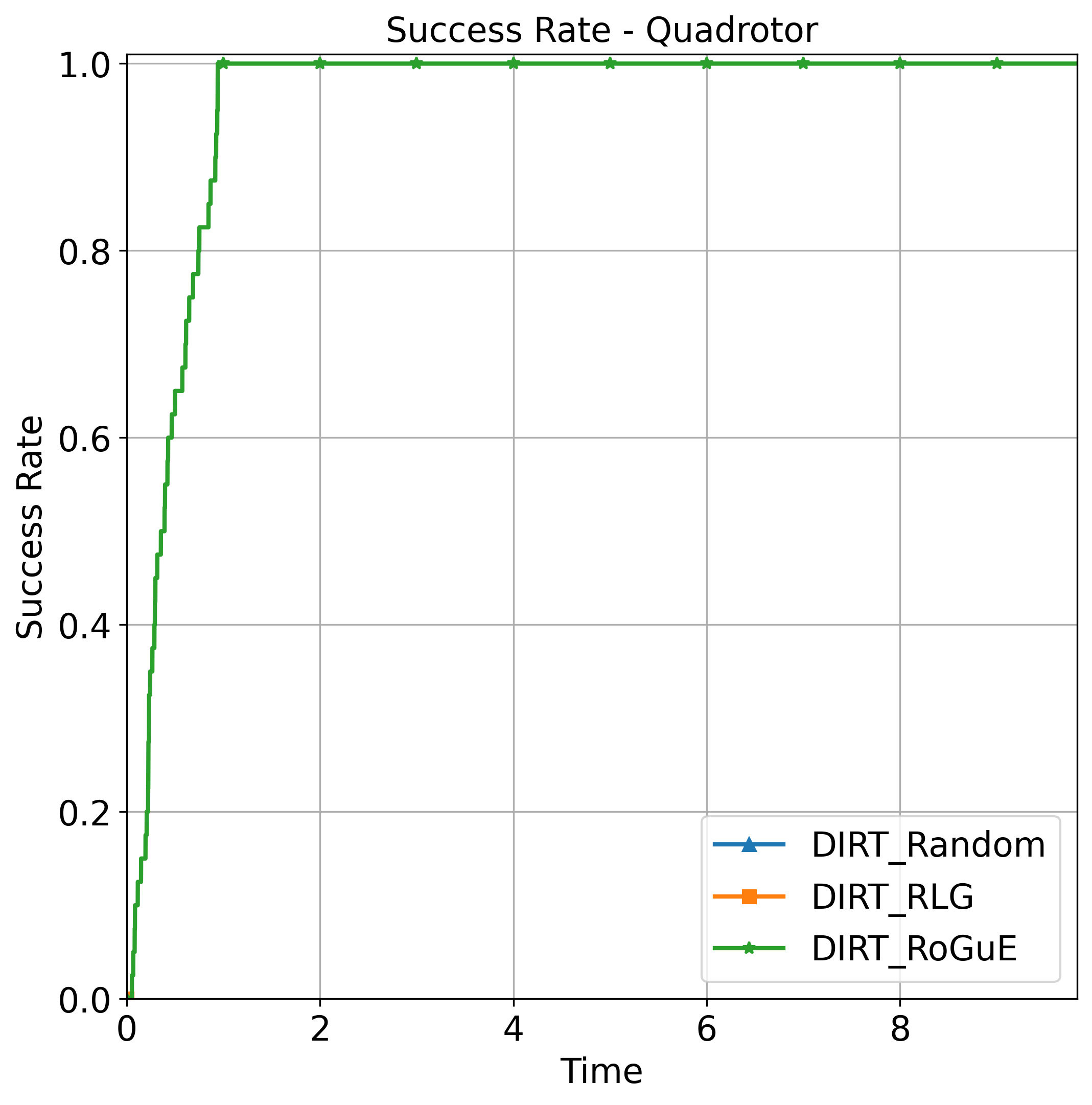}
         \put(-70pt,85pt){\smaller \textcolor{red}{$\mathtt{Random}$ and $\mathtt{RLG}$ do}}
         \put(-68pt,75pt){\smaller \textcolor{red}{not return solutions}}
         \caption{\small Success rates of the planners on each physically simulated benchmark. Higher success rate earlier is better.}
         \label{fig:mujoco-success}
\end{subfigure}\\
\vspace{0.05in}
\begin{subfigure}[b]{\textwidth}
         \includegraphics[width=.235\textwidth]{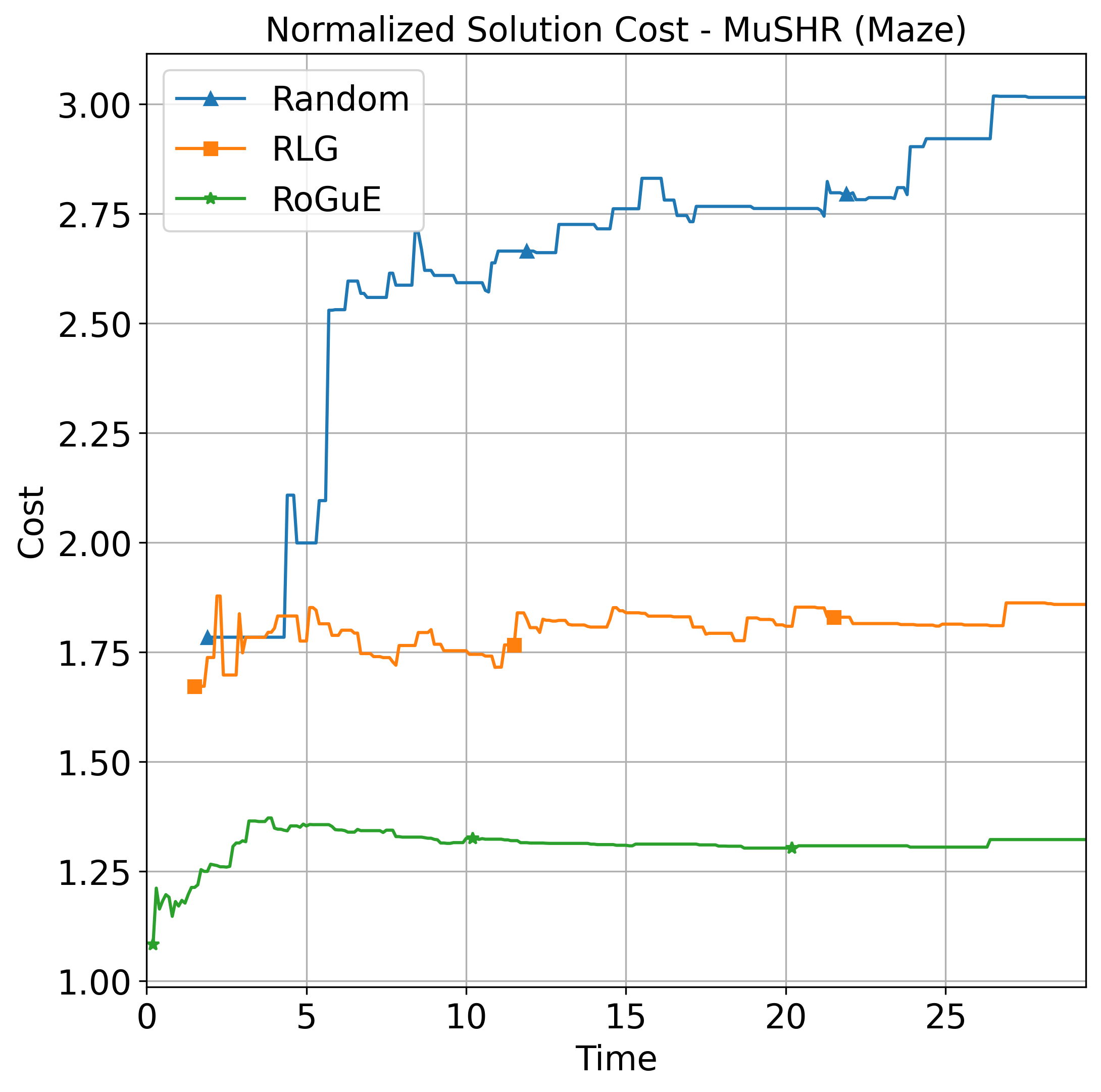}
         \hspace{.01\textwidth}
         \includegraphics[width=.235\textwidth]{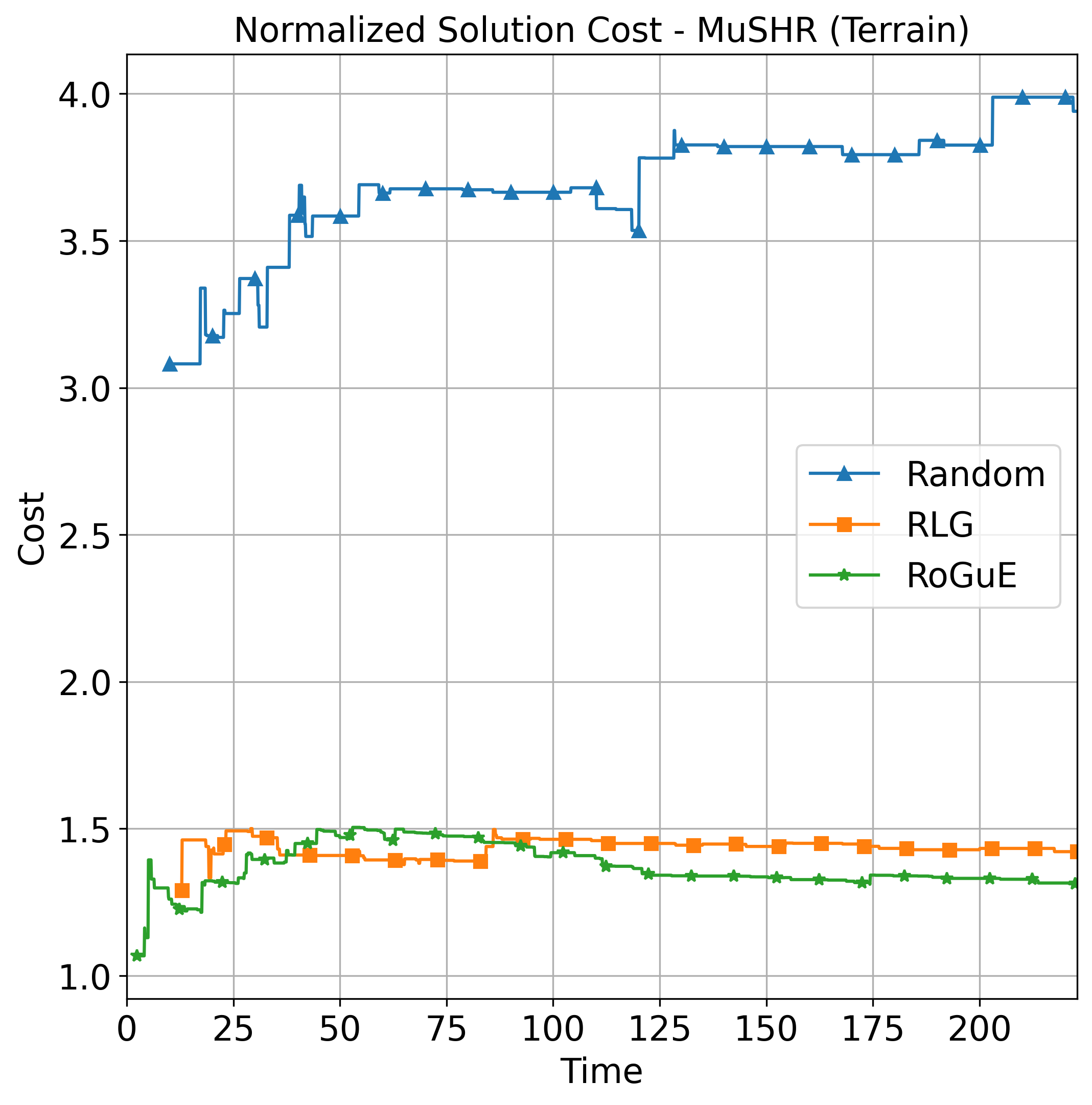}
         \hspace{.01\textwidth}
         \includegraphics[width=.235\textwidth]{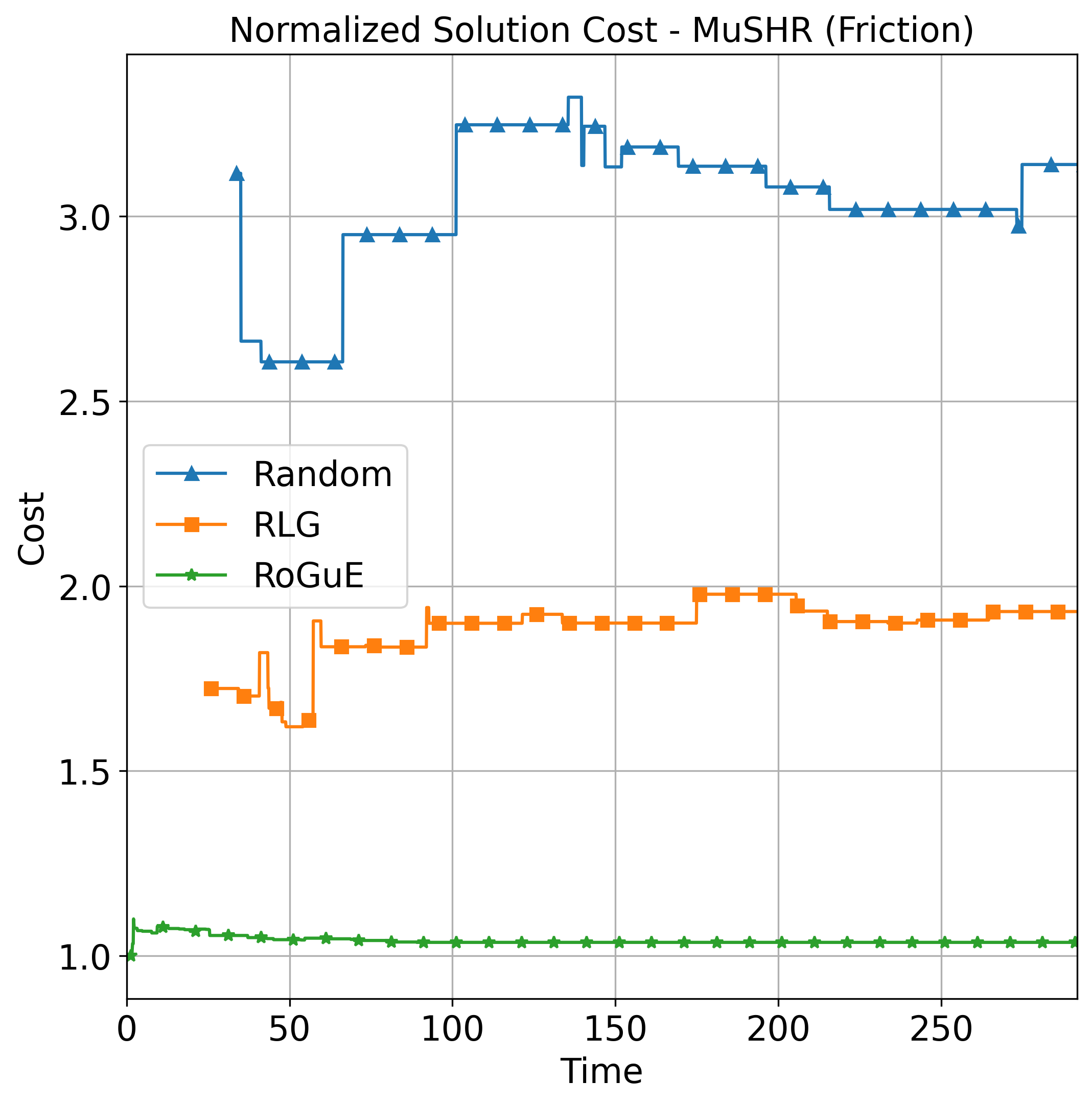}
         \hspace{.01\textwidth}
         \includegraphics[width=.235\textwidth]{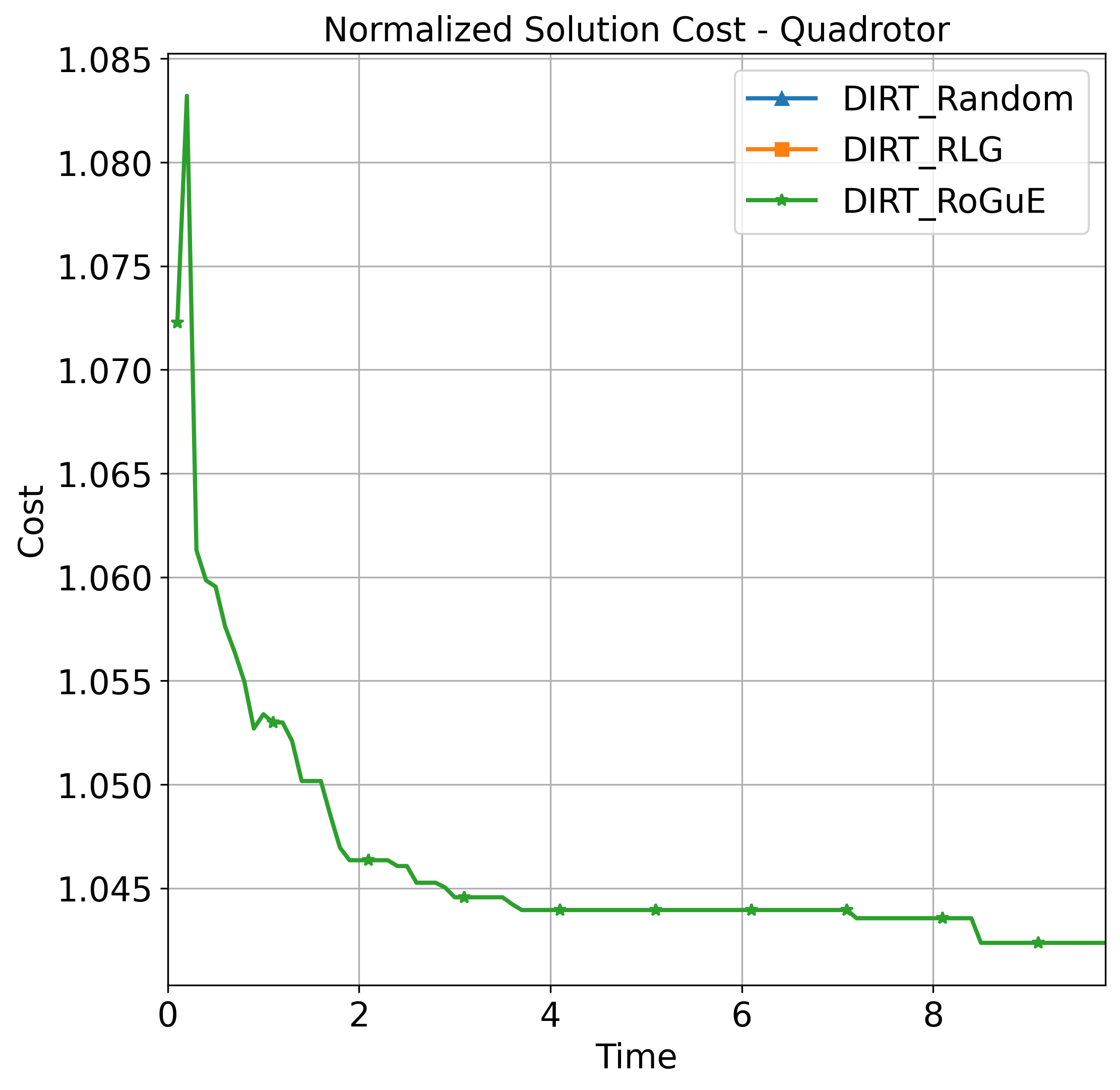}
         \put(-70pt,78pt){\smaller \textcolor{red}{$\mathtt{Random}$ and $\mathtt{RLG}$ do}}
         \put(-68pt,68pt){\smaller \textcolor{red}{not return solutions}}
         \caption{\small Solution quality. Lower values are better. As more solutions are discovered, the average cost may increase over time.}
         \label{fig:mujoco-cost}
\end{subfigure}\\

\caption{\small Physically simulated systems on MuJoC: Each instance is executed 10 times with different random seeds. For the {\tt Quadrotor}, the {\tt Random} and {\tt RLG} strategies consistently fail to return solutions and hence their results are not displayed.}
\label{fig:mushr-results}
\vspace{-.2in}
\end{figure*}
 
{\bf Analytically simulated benchmarks:} The performance of the expansion functions is measured on 3 sets of planning benchmarks for the analytically simulated vehicular systems: (a) 8 problems in an environment with \textbf{\tt Narrow} passages, (b) 6 problems in the \textbf{\tt Indoor} environment from Arena-bench \cite{kastner2022arena}, and (c) 8 problems in the \textbf{\tt Warehouse} environment from Bench-MR \cite{heiden2021bench}. Figure \ref{fig:analytical-results} provides the numerical results. Across all benchmarks, for the differential drive and car-like dynamics, \textbf{\tt DIRT-RoGuE} finds the lowest cost solutions. It is also the only expansion strategy that returns solutions in all trials. Among the RRT expansion strategies, \textbf{{\tt RRT-RoGuE}} achieves the highest success rate and the lowest cost solutions. 


\textbf{Comparison to Path Following:} A na\"ive alternative to the proposed solution is to use the configurations along a path on the roadmap as consecutive local goals for a path following controller. For the car-like system, a pose-reaching controller~\cite{corke2011robotics} was employed to drive the robot to a given pose $\left(\begin{array}{c}v\\ \omega \end{array}\right) =\left(\begin{array}{c}k_\rho \rho\\ k_\alpha \alpha + k_\beta \beta \end{array}\right)$, where each $k$ parameter is a gain term, $\rho$ and  $\alpha$ are the distance and bearing to the local goal respectively, and $\beta$ is the angle difference between $\alpha$ and the current angle. The controller is tested on paths retrieved from the Roadmap with Gaps for the benchmarks of Fig~\ref{fig:analytical-results}. Only 2 such executions, however, returned collision-free trajectories. This is due to (a) the environments containing multiple narrow passages and (b) the paths returned by the roadmap still contain ``gaps'' that the controller cannot easily negotiate. 



{\bf Physically simulated benchmarks:} The performance of the expansion functions is measured on the following benchmarks for MuSHR: (a) a {\tt Maze} from {\tt D4RL} \cite{fu2020d4rl}, (b) uneven {\tt Terrain} environment \cite{stachowicz2023fastrlap}, (c) and a varying friction environment ({\tt Friction}). The controller is trained in an empty environment with flat terrain and uniform friction. The roadmap captures the traversability of different parts of each environment using the controller. For the {\tt Quadrotor} benchmark, the X2 drone must navigate an indoor environment with pillars and air pressure effects. 

\begin{table*}[ht!]
  \centering
\begin{tabular}{|c|ccc|cc|cc|cc|ccc|}
\hline
          & \multicolumn{3}{c|}{\tt RoGuE}                                           & \multicolumn{2}{c|}{\tt SAC+HER}          & \multicolumn{2}{c|}{{ \tt H-SAC+HER}}   & \multicolumn{2}{c|}{{ \tt Random}}   & \multicolumn{3}{c|}{{ \tt RLG}}   \\ \hline
\textbf{Benchmark}  & \multicolumn{1}{c|}{\tt Offl}      & \multicolumn{1}{c|}{\tt Onl}    & {\tt Succ}               & \multicolumn{1}{c|}{\tt Offl}    & {\tt Succ} & \multicolumn{1}{c|}{\tt Offl} & {\tt Succ} & \multicolumn{1}{c|}{\tt Onl} & {\tt Succ} & \multicolumn{1}{c|}{\tt Offl} & \multicolumn{1}{c|}{\tt Onl}   & {\tt Succ} \\ \hline
{\tt Friction}      & \multicolumn{1}{c|}{2.5M }        & \multicolumn{1}{c|}{58.15}      &  \textbf{100\%}       & \multicolumn{1}{c|}{0.81M}       &     {35\%}    & \multicolumn{1}{c|}{1.48M}       &    {13\%}    & \multicolumn{1}{c|}{295.225}       &     {37.5\%}     &   \multicolumn{1}{c|}{1M} & \multicolumn{1}{c|}{609.225}  &    {60\%}   \\ \hline
{\tt Maze}          & \multicolumn{1}{c|}{2.05M}        & \multicolumn{1}{c|}  {241.62}     &  \textbf{100\%}       & \multicolumn{1}{c|}{0.54M}           &     {34\%}    & \multicolumn{1}{c|}{1.74M}       &     {31\%}  & \multicolumn{1}{c|}{623.57}       &     {58\%}     &  \multicolumn{1}{c|}{1M} & \multicolumn{1}{c|}{693.73}   &    {87\%}    \\ \hline
{\tt Quadrotor}     & \multicolumn{1}{c|}{1M}        & \multicolumn{1}{c|} {736.3}    &  \textbf{100\%}       & \multicolumn{1}{c|}{1.3M}           &    {5\%}     & \multicolumn{1}{c|}{1.01M}       &    {13\%}   & \multicolumn{1}{c|}{14k}       &     {0\%}      &  \multicolumn{1}{c|}{100k}  & \multicolumn{1}{c|}{13.7k} &    {0\%}    \\ \hline
\end{tabular}
  \vspace{-.05in}
  \caption{\small Physically-simulated benchmarks: {\tt RoGuE} vs RL in terms of computation cost ($\#$ of calls to physics engine) and success rate.}
\label{tab:RL}
\vspace{-.2in}
\end{table*}

Figs.~\ref{fig:overview} and ~\ref{fig:mujoco-benchmarks} visualize the different environments in MuJoCo, and Fig.~\ref{fig:mushr-results} provides the experimental results.  Only the DIRT motion planner is reported in these experiments as the RRT-based solutions cannot find a solution within the allotted times. Since calls to the MuJoCo engine are expensive, all expansion strategies use a blossom $k=1$.

{\tt RoGuE} finds the lowest cost solution across all benchmarks. \textbf{\tt Random} and \textbf{\tt RLG} consistently fail to find solutions but \textbf{\tt RLG} returns better solutions relative to \textbf{\tt Random}. In the {\tt Maze} and {\tt Friction} benchmarks, {\tt RoGuE} finds solutions across trials quickly. {\tt RoGuE} also returned the most solutions in the {\tt Terrain} benchmark given the same planning budget. In the {\tt Quadrotor} benchmark, {\tt RoGuE} is the only expansion method that discovers any solutions given the tight placement of obstacles and the speed of the X2 drone.

\noindent \textbf{Comparison to RL solutions: } Table~\ref{tab:RL} evaluates two RL-based strategies trained on the benchmarks of Fig~\ref{fig:mushr-results} in terms of success rate {\tt Succ} and offline cost ({\tt Offl} ($\#$ of calls made to the MuJoCo engine). The offline cost is reported when the best performance is achieved and  success rate no longer improves. The online cost {\tt Onl} for sampling-based planners is also reported. The online cost for the RL solutions is minimal. {\tt SAC+HER} trains a goal-conditioned controller $u = \pi(x,q)$ directly in the planning environment. The approach achieves a low success rate, especially for the {\tt Quadrotor} given the joint challenge of complex dynamics and obstacle avoidance. {\tt H-SAC+HER} follows a hierarchical approach similar to {\tt RoGuE} by training a policy to predict local goals for the controller to reach at every step, i.e., $q_\text{lg} = \phi(x)$. This slightly improves success rate relative to {\tt SAC+HER} on {\tt Quadrotor} but lowers it on {\tt Friction}. This shows the difficulty of RL strategies in identifying  informed local goals, which {\tt RoGuE} is able to achieve.

    \section{Conclusion}
\label{sec:conclusion}

This paper proposes a strategy that can benefit from learned controllers to improve the efficiency of kinodynamic planning for robots with significant dynamics. It utilizes a controller trained offline in an empty environment. The target environment is represented via a ``Roadmap with Gaps'' over local regions and applications of the controller between them. Given a wavefront over the roadmap for a specific goal, a tree sampling-based motion planner generates informed subgoals and uses the controller to reach them. When the controller cannot reach a subgoal, the planner resorts to random exploration. Evaluation shows the significant improvement in  planning efficiency.

For higher-dimensional systems, the memory requirements of the roadmap can be improved by considering sparse representations \cite{dobson2014sparse}. Furthermore, learned reachability estimators can assist in efficient roadmap construction and online queries. This work assumes an accurate model of the environment and the robot, which complicates deployment on real systems. This motivates integrating the proposed motion planner with system identification, state estimation and feedback control to track the planned trajectory.


}

\bibliographystyle{format/IEEEtran}
\bibliography{refs.bib}

\begin{thebibliography}{10}
\providecommand{\url}[1]{#1}
\csname url@rmstyle\endcsname
\providecommand{\newblock}{\relax}
\providecommand{\bibinfo}[2]{#2}
\providecommand\BIBentrySTDinterwordspacing{\spaceskip=0pt\relax}
\providecommand\BIBentryALTinterwordstretchfactor{4}
\providecommand\BIBentryALTinterwordspacing{\spaceskip=\fontdimen2\font plus
\BIBentryALTinterwordstretchfactor\fontdimen3\font minus
  \fontdimen4\font\relax}
\providecommand\BIBforeignlanguage[2]{{%
\expandafter\ifx\csname l@#1\endcsname\relax
\typeout{** WARNING: IEEEtran.bst: No hyphenation pattern has been}%
\typeout{** loaded for the language `#1'. Using the pattern for}%
\typeout{** the default language instead.}%
\else
\language=\csname l@#1\endcsname
\fi
#2}}

\bibitem{lavalle2001randomized}
S.~M. LaValle and J.~J. Kuffner~Jr, ``Randomized kinodynamic planning,''
  \emph{IJRR}, vol.~20, no.~5, pp. 378--400, 2001.

\bibitem{li2016asymptotically}
Y.~Li, Z.~Littlefield, and K.~E. Bekris, ``Asymptotically optimal
  sampling-based kinodynamic planning,'' \emph{IJRR}, 2016.

\bibitem{hauser2016asymptotically}
K.~Hauser and Y.~Zhou, ``Asymptotically optimal planning by feasible
  kinodynamic planning in a state--cost space,'' \emph{T-RO}, 2016.

\bibitem{LB-DIRT}
Z.~{Littlefield} and K.~E. {Bekris}, ``Efficient and asymptotically optimal
  kinodynamic motion planning via dominance-informed regions,'' in \emph{IROS},
  2018.

\bibitem{kleinbort2020refined}
M.~Kleinbort, E.~Granados, K.~Solovey, R.~Bonalli, K.~E. Bekris, and
  D.~Halperin, ``Refined analysis of asymptotically-optimal kinodynamic
  planning in the state-cost space,'' in \emph{ICRA}, 2020.

\bibitem{Faust2018PRMRLLR}
A.~Faust, O.~Ram{\'i}rez, M.~Fiser, K.~Oslund, A.~Francis, J.~O. Davidson, and
  L.~Tapia, ``{PRM-RL}: Long-range robotic navigation tasks by combining
  reinforcement learning and sampling-based planning,'' \emph{ICRA}, 2018.

\bibitem{chiang2019rl}
H.-T.~L. Chiang, J.~Hsu, M.~Fiser, L.~Tapia, and A.~Faust, ``{RL-RRT}:
  Kinodynamic motion planning via learning reachability estimators from {RL}
  policies,'' \emph{RA-L}, vol.~4, no.~4, pp. 4298--4305, 2019.

\bibitem{ROB-063}
T.~McMahon, A.~Sivaramakrishnan, E.~Granados, and K.~E. Bekris, ``A survey on
  the integration of machine learning with sampling-based motion planning,''
  \emph{Foundations and Trends in Robotics}, 2022.

\bibitem{Li2018NeuralNA}
Y.~Li, R.~Cui, Z.~Li, and D.~Xu, ``Neural network approximation based
  near-optimal motion planning with kinodynamic constraints using {RRT},''
  \emph{IEEE Transactions on Industrial Electronics}, vol.~65, pp. 8718--8729,
  2018.

\bibitem{kontoudis2019kinodynamic}
G.~P. Kontoudis and K.~G. Vamvoudakis, ``Kinodynamic motion planning with
  continuous-time q-learning: An online, model-free, and safe navigation
  framework,'' \emph{IEEE transactions on neural networks and learning
  systems}, vol.~30, no.~12, pp. 3803--3817, 2019.

\bibitem{johnson2020mpnet}
J.~J. {Johnson}, L.~{Li}, F.~{Liu}, A.~H. {Qureshi}, and M.~C. {Yip},
  ``Dynamically constrained motion planning networks for non-holonomic
  robots,'' in \emph{IROS}, 2020, pp. 6937--6943.

\bibitem{li2021mpc}
L.~Li, Y.~Miao, A.~H. Qureshi, and M.~C. Yip, ``Mpc-mpnet: Model-predictive
  motion planning networks for fast, near-optimal planning under kinodynamic
  constraints,'' \emph{RA-L}, vol.~6, no.~3, 2021.

\bibitem{xu2022benchmarking}
Z.~Xu, B.~Liu, X.~Xiao, A.~Nair, and P.~Stone, ``Benchmarking reinforcement
  learning techniques for autonomous navigation,'' in \emph{ICRA}, 2023.

\bibitem{eysenbach2019search}
B.~Eysenbach, R.~R. Salakhutdinov, and S.~Levine, ``Search on the replay
  buffer: Bridging planning and reinforcement learning,'' \emph{NeurIPS}, 2019.

\bibitem{emmons2020sparse}
S.~Emmons, A.~Jain, M.~Laskin, T.~Kurutach, P.~Abbeel, and D.~Pathak, ``Sparse
  graphical memory for robust planning,'' \emph{NeurIPS}, 2020.

\bibitem{bagaria2019option}
A.~Bagaria and G.~Konidaris, ``Option discovery using deep skill chaining,'' in
  \emph{ICLR}, 2019.

\bibitem{bagaria2021robustly}
A.~Bagaria, J.~K. Senthil, M.~Slivinski, and G.~Konidaris, ``Robustly learning
  composable options in deep reinforcement learning.'' in \emph{IJCAI}, 2021.

\bibitem{bagaria2021skill}
A.~Bagaria, J.~K. Senthil, and G.~Konidaris, ``Skill discovery for exploration
  and planning using deep skill graphs,'' in \emph{International Conference on
  Machine Learning}.\hskip 1em plus 0.5em minus 0.4em\relax PMLR, 2021, pp.
  521--531.

\bibitem{strudel2020learning}
R.~Strudel, R.~Garcia, J.~Carpentier, J.-P. Laumond, I.~Laptev, and C.~Schmid,
  ``Learning obstacle representations for neural motion planning,'' in
  \emph{CoRL}, 2020.

\bibitem{sivaramakrishnan2021improving}
A.~Sivaramakrishnan, E.~Granados, S.~Karten, T.~McMahon, and K.~E. Bekris,
  ``Improving kinodynamic planners for vehicular navigation with learned
  goal-reaching controllers,'' in \emph{IROS}, 2021.

\bibitem{troy2022terrains}
T.~McMahon, A.~Sivaramakrishnan, K.~Kedia, E.~Granados, and K.~E. Bekris,
  ``Terrain-aware learned controllers for sampling-based kinodynamic planning
  over physically simulated terrains,'' in \emph{IROS}, 2022.

\bibitem{le2014guiding}
D.~Le and E.~Plaku, ``Guiding sampling-based tree search for motion planning
  with dynamics via probabilistic roadmap abstractions,'' in \emph{IROS}, 2014.

\bibitem{westbrook2020anytime}
M.~G. Westbrook and W.~Ruml, ``Anytime kinodynamic motion planning using
  region-guided search,'' in \emph{IROS}, 2020.

\bibitem{shome2021asymptotically}
R.~Shome and L.~E. Kavraki, ``Asymptotically optimal kinodynamic planning using
  bundles of edges,'' in \emph{ICRA}, 2021.

\bibitem{todorov2012mujoco}
E.~Todorov, T.~Erez, and Y.~Tassa, ``Mujoco: A physics engine for model-based
  control,'' in \emph{IROS}, 2012.

\bibitem{Haarnoja2018SoftAO}
T.~Haarnoja, A.~Zhou, P.~Abbeel, and S.~Levine, ``Soft actor-critic: Off-policy
  maximum entropy deep reinforcement learning with a stochastic actor,'' in
  \emph{ICML}, 2018.

\bibitem{Andrychowicz2017HindsightER}
M.~Andrychowicz, D.~Crow, A.~Ray, J.~Schneider, R.~H. Fong, P.~Welinder,
  B.~McGrew, J.~Tobin, P.~Abbeel, and W.~Zaremba, ``Hindsight experience
  replay,'' in \emph{NeurIPS}, 2017.

\bibitem{srinivasa2019mushr}
S.~S. Srinivasa, P.~Lancaster, J.~Michalove, M.~Schmittle, C.~Summers,
  M.~Rockett, J.~R. Smith, S.~Chouhury, C.~Mavrogiannis, and F.~Sadeghi,
  ``{MuSHR}: A low-cost, open-source robotic racecar for education and
  research,'' \emph{CoRR}, vol. abs/1908.08031, 2019.

\bibitem{ML4KP}
E.~Granados, A.~Sivaramakrishnan, T.~McMahon, Z.~Littlefield, and K.~E. Bekris,
  ``Machine learning for kinodynamic planning (ml4kp),''
  \url{https://github.com/PRX-Kinodynamic/ML4KP}, 2021--2022.

\bibitem{honig2022db}
W.~H{\"o}nig, J.~Ortiz-Haro, and M.~Toussaint, ``db-a*: Discontinuity-bounded
  search for kinodynamic mobile robot motion planning,'' in \emph{IROS}, 2022.

\bibitem{kastner2022arena}
L.~K{\"a}stner, T.~Bhuiyan, T.~A. Le, E.~Treis, J.~Cox, B.~Meinardus,
  J.~Kmiecik, R.~Carstens, D.~Pichel, B.~Fatloun, \emph{et~al.}, ``Arena-bench:
  A benchmarking suite for obstacle avoidance approaches in highly dynamic
  environments,'' \emph{RA-L}, vol.~7, no.~4, 2022.

\bibitem{heiden2021bench}
E.~Heiden, L.~Palmieri, L.~Bruns, K.~O. Arras, G.~S. Sukhatme, and S.~Koenig,
  ``Bench-mr: A motion planning benchmark for wheeled mobile robots,''
  \emph{RA-L}, vol.~6, no.~3, 2021.

\bibitem{corke2011robotics}
P.~I. Corke, W.~Jachimczyk, and R.~Pillat, \emph{Robotics, vision and control:
  fundamental algorithms in MATLAB}.\hskip 1em plus 0.5em minus 0.4em\relax
  Springer, 2011, vol.~73.

\bibitem{fu2020d4rl}
\BIBentryALTinterwordspacing
J.~Fu, A.~Kumar, O.~Nachum, G.~Tucker, and S.~Levine, ``D4rl: Datasets for deep
  data-driven reinforcement learning,'' 2020. [Online]. Available:
  \url{https://arxiv.org/abs/2004.07219}
\BIBentrySTDinterwordspacing

\bibitem{stachowicz2023fastrlap}
K.~Stachowicz, A.~Bhorkar, D.~Shah, I.~Kostrikov, and S.~Levine, ``{FastRLAP: A
  System for Learning High-Speed Driving via Deep RL and Autonomous
  Practicing},'' 2023.

\bibitem{dobson2014sparse}
A.~Dobson and K.~E. Bekris, ``Sparse roadmap spanners for asymptotically
  near-optimal motion planning,'' \emph{IJRR}, 2014.

\end{thebibliography}

\end{document}